\documentclass[a4paper,fleqn]{cas-sc}
\usepackage{setspace}
\usepackage[authoryear,longnamesfirst]{natbib}

\usepackage{amsthm}
\usepackage{algorithm}
\usepackage{algorithmic}
\usepackage{makecell}
\usepackage{hhline}
\usepackage[export]{adjustbox}
\usepackage{placeins}
\usepackage{float}
\newtheorem{definition}{Definition}

\begin{document}
\let\WriteBookmarks\relax
\def\floatpagepagefraction{1}
\def\textpagefraction{.001}
\shorttitle{Weak-PDE-Net}
\shortauthors{X. Li et~al.}

\title [mode = title]{Weak-PDE-Net: Discovering Open-Form PDEs via Differentiable Symbolic Networks and Weak Formulation}

\author[1]{Xinxin Li}[
                        auid=000,bioid=1,
                        orcid=0009-0008-8962-0084]

\ead{51265500102@stu.ecnu.edu.cn}
\affiliation[1]{organization={School of Mathematical Sciences, East China Normal University},
                city={Shang Hai},
                postcode={200241}, 
                country=China}

\author[2,4]{Xingyu Cui}
\ead{cuixingyu212@163.com}
\author[2]{Jin Qi}
\ead{qi_jin@iapcm.ac.cn}
\author[3,4]{Juan Zhang}
\cormark[1]
\ead{zhang_juan@buaa.edu.cn}

\affiliation[2]{organization={Institute of Applied Physics and Computational Mathematics},
                postcode={100094}, 
                city={Beijing},
                country={China}}

\affiliation[3]{organization={Institute of Artiﬁcial Intelligence, Beihang University},
                postcode={100191}, 
                city={Beijing},
                country={China}}

\affiliation[4]{organization={Shanghai Zhangjiang Institute of Mathematics},
                postcode={201203}, 
                city={Shanghai},
                country={China}}

\author[5]{Da Li}
\affiliation[5]{organization={Academy for Advanced Interdisciplinary Studies, Northeast Normal University},
                postcode={130024}, 
                city={Changchun},
                state={Jilin},
                country={China}}
\ead{dli@nenu.edu.cn}
\author[1,2,4,5]{Junping Yin}
\cormark[1]
\ead{yinjp829829@126.com}
\cortext[cor1]{Corresponding author}

\begin{abstract}
Discovering governing Partial Differential Equations (PDEs) from sparse and noisy data is a challenging issue in data-driven scientific computing. Conventional sparse regression methods often suffer from two major limitations: (i) the instability of numerical differentiation under sparse and noisy data, and (ii) the restricted flexibility of a pre-defined candidate library. We propose Weak-PDE-Net, an end-to-end differentiable framework that can robustly identify open-form PDEs. Weak-PDE-Net consists of two interconnected modules: a forward response learner and a weak-form PDE generator. The learner embeds learnable Gaussian kernels within a lightweight MLP, serving as a surrogate model that adaptively captures system dynamics from sparse observations. Meanwhile, the generator integrates a symbolic network with an integral module to construct weak-form PDEs, avoiding explicit numerical differentiation and improving robustness to noise. To relax the constraints of the pre-defined library, we leverage Differentiable Neural Architecture Search strategy during training to explore the functional space, which enables the efficient discovery of open-form PDEs. The capability of Weak-PDE-Net in multivariable systems discovery is further enhanced by incorporating Galilean Invariance constraints and symmetry equivariance hypotheses to ensure physical consistency. Experiments on several challenging PDE benchmarks demonstrate that Weak-PDE-Net accurately recovers governing equations, even under highly sparse and noisy observations.
\end{abstract}


\maketitle

\section{Introduction}
\label{sec: Introduction}
A central goal of modern science is to discover interpretable mathematical models governing complex natural phenomena from observations. Among various models, partial differential equations (PDEs) are a powerful tool for depicting complex systems in fluid dynamics \citep{anderson1995computational}, pharmacological processes \citep{shargel2015applied}, climate modeling \citep{lynch2006emergence}, and other fields. Consequently, advancing our comprehension of the natural systems requires the ability to discover governing PDEs from real-world observations.

Sparse regression-based methods \citep{rudy2017data,messenger2021weak} are the most established methods for PDE discovery. These methods have shown remarkable efficacy in discovering physical laws from data, thereby facilitating the modeling of complex dynamical systems. Despite this success, these methods suffer from two limitations. First, regression-based methods typically rely on a regular grid to estimate derivatives or integral terms. However, real-world observational data are always sparse and noisy, which poses a challenge for accurate numerical differentiation or integration. Second, limited by the pre-defined library of candidate terms, these methods lack the ability to discover open-form PDEs.

The first limitation involves robustness when dealing with imperfect data. Recent methods \citep{xu2019dl,chen2022symbolic} leverage Deep Neural Networks (DNNs) as continuous surrogates to robustly estimate derivatives or integrals from sparse observations. This reliance on deep learning shifts the challenge from data processing to function approximation, but introducing new challenge related to model expressivity. Specifically, when implemented as standard Multilayer Perceptrons (MLPs), these networks often suffer from spectral bias \citep{rahaman2019spectral}, prioritizing low-frequency components and struggling to capture high-frequency behaviors without prohibitive computational costs \citep{rahaman2019spectral}. Rational Neural Networks (RatNNs) have been introduced to enhance approximation capabilities \citep{stephany2024pde,stephany2024weak}, but their dependence on finite-degree polynomials can lead to numerical instability. Consequently, it remains an unresolved issue that stable and accurate approximations from sparse and noisy observations.

Beyond robustness, the  flexibility of these methods is restricted by the pre-defined candidate library.  To enable open-form PDEs discovery, recent works \citep{chen2022symbolic,du2024discover} represent PDEs using symbolic forest and employ genetic algorithms or reinforcement learning to search for optimal trees within the functional term space. While promising, these methods face two critical issues. Primarily, they are discrete and cannot be seamlessly integrated into differentiable neural network training, resulting in a decoupled and two-stage strategy. This means that the discovered equations cannot be embedded into the learning process of the response function concurrently. Furthermore, discovering multivariable PDEs is difficult because the vast search space promotes overfitting, while the decoupled discovery of each state variable fails to maintain global physical consistency. Consequently, there is a clear need for an end-to-end differentiable framework capable of discovering physically consistent open-form PDEs.

To address these challenges, we propose Weak-PDE-Net, an end-to-end differentiable framework designed for the robust discovery of open-form PDEs from sparse and noisy data. Weak-PDE-Net consists of two main modules: a forward response learner that embeds learnable Gaussian embeddings into a lightweight MLP, and a weak-form PDE generator that couples the symbolic network with the integral module. The training process of Weak-PDE-Net involves three phases: Searching, Pruning, and Tuning. The core contributions of this work are as follows:

\begin{itemize}

\item We introduce adaptive Gaussian kernels to dynamically adjust the frequency spectrum of the response learner. This design effectively mitigates the spectral bias common in standard MLPs, allowing the model to capture complex system dynamics accurately even from sparse and noisy data.

\item  We propose a differentiable symbolic network architecture search strategy in searching phase that identifies network architectures best adapted to the observed data. This strategy dynamically constructs the corresponding library of function terms, enabling the discovery of open-form PDEs.

\item We incorporate physical-Informed refinement, specifically Galilean Invariance constraint and symmetry equivariance hypothesis. This refinement screens out spurious terms and supplements missing dynamics during training, ensuring that the discovered equations for complex multivariable systems adhere to physical laws.

\item We integrate the above components into an end-to-end differentiable framework. This integration couples the learning of the response function with the discovery of the equation, enabling them to constrain and reinforce each other, significantly enhancing robustness against sparse and noisy data.
\end{itemize}

The rest of this paper is organized as follows: Section \ref{sec: Problem Statement} formally describes the problem at hand, including the definition of the PDEs form we aim to discover. In Section \ref{sec: Related Works}, we survey related works on discovering scientific laws from data, with a specific focus on PDE discovery. Section \ref{sec: Methodology} provides a detailed description of the Weak-PDE-Net framework. In Section \ref{sec: Experiments} and \ref{sec: Results}, we evaluate the performance of Weak-PDE-Net by discovering several benchmarks from limited and noisy observations. Section \ref{sec: Discussion} discusses the results and the rationale behind our algorithm. Finally, concluding remarks and potential future directions are provided in Section \ref{sec: Conclusion}.

\section{Problem Statement}
\label{sec: Problem Statement}
Weak-PDE-Net discovers a governing PDE from sparse and noisy observations of the system response function $U$. To provide a precise mathematical description, we start with the definition of the system response function and the properties of the available dataset. This leads to the specific formulation of our open-form PDE, which represents the underlying dynamics. The section concludes by formalizing the optimization objective required to determine both the symbolic structure and the unknown coefficients.
\subsection{System Response Function}

Let $T > 0$ and $\Omega \subseteq \mathbb{R}^d$ be a compact spatial domain. We assume there is a function $U: [0, T] \times \Omega \rightarrow \mathbb{R}^{h}$, called the system response function. We refer to $\Omega$, $[0, T]$, and $\Omega_T = [0, T] \times \Omega$ as the spatial, temporal, and spatio-temporal domains, respectively. The output dimension $h$ denotes the number of state variables. When $h=1$, $U$ represents a scalar field, and when $h \ge 2$, $U$ corresponds to a vector field, in which case we refer to the system as a multivariable system.

\subsection{Sparse and Noisy Observation Data}

In realistic scenarios, the continuous field $U$ is inaccessible. Instead, we observe the system at a set of discrete, scattered coordinates. We consider a dataset $\mathcal{D} = \{(t_i, X_i, \tilde{U}_i)\}_{i=1}^{N_{\text{data}}}$, where $X_i \in \Omega$ denotes the spatial coordinate and $\tilde{U}_i \in \mathbb{R}^h$ represents the noisy observation:

\begin{equation}
\tilde{U}_i = U(t_i, X_i) + \epsilon_i,
\end{equation}
where $\epsilon_i$ denotes independent noise. Since the observation points $\{(t_i, X_i)\}$ are sparse and irregularly distributed, they do not form a structured grid. This precludes the use of standard finite difference methods for derivative estimation, as well as classical numerical integration methods.

\subsection{Open-form PDE}
\label{sec: Open-form PDE Formulation}
We assume the system response function $U$ are governed by a partial differential equation. Different from traditional methods that select from a fixed library of candidate terms, we adopt an \textit{open-form} setting in which the equation structure is not pre-defined but is constructed adaptively. To formulate this discovery process, we define a generalized governing equation represented as:

\begin{equation}
\label{eq: pde_strong_form}
D^{\alpha^{(0)}} F_{0}(U) = \sum_{s=1}^{S} \xi_{s} D^{\alpha^{(s)}} F_{s}(U),  \quad (t, X) \in \Omega_T,
\end{equation}
where $\alpha^{(s)} = (\alpha_0^{(s)}, \dots, \alpha_d^{(s)})$ is a multi-index representing the derivative orders. Notably, instead of choosing $F_s$ from a fixed dictionary, each term $F_s$ is a composite function constructed from a symbolic space spanned by a set of fundamental operators $\mathcal{O}$ (e.g., $\{+,-,\times,\sin\}$).

\subsection{Optimization Objective}

With the representation in Section \ref{sec: Open-form PDE Formulation}, the discovery task is formulated as a joint optimization problem over the symbolic structures $\{F_s\}$, the derivative orders $\{\alpha^{(s)}\}$, and the sparse coefficients $\{\xi_s\}$. This problem is challenging due to the interdependence between structure discovery and parameter estimation. Unlike traditional regression where the library is fixed, our method dynamically construct the function terms from scratch.

\section{Related Works}
\label{sec: Related Works}

\subsection{Robust PDE Discovery from Sparse and Noisy Data}
\label{sec: Robust PDE Discovery from Sparse and Noisy Data}

Strong-form methods identify governing equations via the differential form. Early strong-form methods, such as SINDy \citep{brunton2016discovering} and PDE-FIND \citep{rudy2017data}, rely on derivative calculations to construct a candidate library. And they employ sparse regression on this pre-defined library to discover the governing equations. Although effective in certain conditions, these methods require observational data to be acquired on a regular grid, which greatly limits their application in real-world scenarios. In addition, these methods depend on numerical differentiation to estimate derivatives, which tends to amplify noise. Consequently, higher noise levels significantly reduce their effectiveness.

To deal with the sparse and noisy data in real-world scenarios, {DL-PDE} \citep{xu2019dl} pioneered the use of DNNs as continuous surrogates to approximate the system response function, treating equation discovery as a subsequent optimization step using genetic algorithms. {DeepMoD} \citep{both2021deepmod} similarly employs a DNN surrogate but utilizes automatic differentiation to construct a candidate library. By enforcing sparsity on the library coefficients through loss regularization, it fuses response learning and equation discovery into a unified framework. {PDE-LEARN} \citep{stephany2024pde} further improved this {end-to-end framework} by employing RatNNs \citep{boulle2020rational} to enhance approximation accuracy. Most of these methods fundamentally rely on constructing the candidate library via automatic differentiation. Although these neural surrogate models help smooth the underlying signal and alleviate the impact of noise on differentiation, their robustness still remains insufficient to handle higher noise levels.

In response to the challenge in robustness, {weak-form} methods have emerged as a promising alternative. The weak-form PDE transfers derivative operations from the noisy data to the test function, thereby avoiding direct differentiation of the response function. Recently, {Weak-PDE-LEARN} \citep{stephany2024weak} advanced this direction by converting the original PDE-LEARN from a strong-form to a weak-form method. This method replaces direct differentiation with integration over local regions, which naturally smooths out noise in the data. As a result, Weak-PDE-LEARN achieves much better robustness to noise than the original strong-form version.

Despite these improvements, neural surrogates still face a common problem. Both standard DNNs and RatNNs suffer from spectral bias \citep{rahaman2019spectral}, meaning they learn low-frequency functions much faster than high-frequency ones. Consequently, these networks act like low-pass filters and smooth out sharp physical features. This issue is even more severe for weak-form methods. While the integral nature of the weak form helps handle noise, it also acts as a natural smoother, which further hides fine physical details. Therefore, we add adaptive Gaussian kernels to our Weak-PDE-Net. Since the kernel parameters are learnable, the model can adjust its focus to capture both smooth patterns and sharp high-frequency details.

\subsection{Open-form PDE Discovery via Symbolic Regression}
Methods in Section \ref{sec: Robust PDE Discovery from Sparse and Noisy Data} rely on a pre-defined candidate library. This pre-defined library limits the flexibility of the method and prevents open-form PDE discovery. To overcome the constraints of the pre-defined library, symbolic regression methods have been introduced to adaptively expand the search space. {DLGA-PDE} \citep{xu2020dlga} uses genetic algorithms to search for function terms. Although its mutation operations generate new combinations, the resulting set of function terms still remains restricted to a limited space. To overcome this, {SGA-PDE} \citep{chen2022symbolic} pioneered the discovery of open-form PDEs by representing function terms as symbolic trees composed of operators and operands, thereby enabling exploration within an infinite space. Similarly, inspired by Reinforcement Learning-based methods, {Discover} \citep{du2024discover} employs an LSTM-based controller to generate symbolic trees, utilizing reinforcement learning to guide the generation direction towards the optimal trees.

Nevertheless, the discrete and non-differentiable nature of symbolic optimization poses a challenge for end-to-end integration. These strategies typically rely on a decoupled and two-stage process: employing a neural network to estimate the system response and its derivatives, and subsequently performing symbolic search on these estimates. Therefore, the discovered equations cannot be embedded into the learning process of the response function. This  disconnection leaves the process susceptible to error propagation from the initial approximation step. Differentiable symbolic methods offer a promising path toward end-to-end learning. EQL \citep{sahoo2018learning} first proposed symbolic networks by replacing traditional activation functions with mathematical operators (e.g., $\sin, \times$). Then PDE-Net 2.0 \citep{long2019pde} integrates symbolic networks to learn differential operators. However, its operator set is limited to $\{+,\times\}$, thereby restricting the discovery scope exclusively to polynomial forms. Furthermore, its fixed symbolic architecture makes the model susceptible to overfitting, often yielding complex identified equations.

To overcome these limitations, we expand the symbolic operator set to include a broader range of functions (e.g., $\{+,-,\times,\div,\sin\}$). Meanwhile, we design a differentiable architecture search strategy that adaptively identifies lightweight network architecture from data, thereby enabling open-form PDE discovery and mitigating overfitting.

\subsection{Search Space Refinement via Physical Consistency}

In scientific discovery, the search space for potential equation terms grows exponentially with the number of variables. Due to this combinatorial explosion, purely data-driven methods often struggle to identify governing equations for multivariable or high-dimensional systems. To overcome this challenge, AI-Feynman \citep{udrescu2020ai} introduces dimensional analysis into symbolic regression, simplifying tasks by reducing high-dimensional problems into lower-dimensional ones. In the field of ordinary differential equations (ODEs) discovery, \citet{yang2024symmetry} leverage symmetries in equation discovery to compress the search space, thereby improving the accuracy and simplicity of the learned equations. In the problem of PDE discovery, {ICNet} \citep{chen2024invariance} proposed an {invariance-constrained} framework that leverages fundamental physical properties, Galilean and Lorentz invariance, to prune the candidate library. 

In our work, the discovery strategy is adapted to the system's complexity. For scalar fields ($h=1$), a purely data-driven method suffices. As for the multivariable system ($h\ge 2$), the discovery process becomes significantly more challenging for two main reasons. First, the search space for potential equation terms expands exponentially with the number of state variables. Second, discovering the governing equation for each state variable independently often leads to physical inconsistency especially in complex-valued systems. Drawing inspiration from ICNet, we explicitly introduce {Galilean Invariance} to guide the discovery process. This physical constraint is employed not only to filter out invalid function items but also to supplement necessary invariant terms. Furthermore, we incorporate the anti-symmetric structure derived from symmetry equivariance as a structural hypothesis for complex-valued systems. By providing symmetry-aware candidates, our framework enables the method to adaptively synchronize the learning of real and imaginary components.

\section{Methodology}
\label{sec: Methodology}

In this section, we present the methodology of Weak-PDE-Net. To start with, we establish the theoretical foundation by deriving the weak-form PDE and defining test functions in Sections \ref{sec: The weak-form PDE}, \ref{sec: The selection of test functions}. On this basis, Section \ref{sec: The architecture of Weak-PDE-Net} focuses on the network architecture designed for equation discovery. The optimization mechanism, including the loss function and training strategy, is then detailed in Sections \ref{sec: The loss function of Weak-PDE-Net}, \ref{sec: The training strategy of Weak-PDE-Net}. Finally, we incorporate physical information in Section \ref{sec: The Physics-Informed Refinement of Weak-PDE-Net}.

\subsection{The Derivation of the Weak-form PDE}
\label{sec: The weak-form PDE}
To avoid the instability of directly estimating high-order derivatives from sparse and noisy data, we derive the weak formulation \citep{evans2022partial} of the strong-form PDE in Eq.~\eqref{eq: pde_strong_form}.To start with, we multiply both sides of the strong-form Eq.~\eqref{eq: pde_strong_form} by a smooth test function $\Phi(t,X) \in C_c^{\infty}([0,T] \times \Omega)$ and integrate over the spacetime domain. This yields

\begin{equation}
\label{eq: integration_form}
\int_{0}^{T} \int_{\Omega} \Phi(t,X) D^{\alpha^{(0)}} F_{0}(U) dX dt = \sum_{s=1}^{S} \xi_s \int_{0}^{T} \int_{\Omega} \Phi(t,X) D^{\alpha^{(s)}} F_{s}(U) dX dt.
\end{equation}

By repeatedly applying Green’s lemma \citep{langtangen2003computational}, we can transfer the derivatives from the potentially noisy terms to the smooth test function. Since $\Phi(t,X)$ is assumed to have compact support within the domain, it vanishes on the boundary, thereby eliminating the boundary integral terms. Consequently, the weak formulation is obtained as:
\begin{equation}
    \label{eq: Applying Green’s lemma}
    \int_{0}^{T}\int_{\Omega}
(-1)^{|\alpha^{(0)}|} D^{\alpha^{(0)}} \Phi(t,X) F_{0}(U(t,X))dXdt=\sum_{s=1}^{S} \xi_{s}
\int_{0}^{T}\int_{\Omega}
(-1)^{|\alpha^{(s)}|} D^{\alpha^{(s)}} \Phi(t,X)F_{s}(U(t,X))dXdt,
\end{equation}
where $|\alpha^{(s)}| = \alpha_0^{(s)} + \cdots + \alpha_{d}^{(s)}.
$
Compared to the strong form, it eliminates the high-order derivatives computation of \(U\), making the formulation more robust to noise.

To transform the weak-form PDE into a computable algebraic system, we select a finite set of smooth test functions $\{\Phi_k\}_{k=1}^{K} \subset C_c^{\infty}([0,T] \times \Omega)$. This transition is driven by two main objectives. First, it allows us to aggregate local information from each test function's support into a global PDE representation. Second, it converts the discovery problem into a regression task that is more amenable to neural network optimization.

For each $k \in \{1, 2, \dots, K\}$, we define the integrated quantity $b_k$ as:
\begin{equation}
\label{eq: b_k}
b_{k}= \int_{0}^{T} \int_{\Omega} (-1)^{|\alpha^{(0)}|} D^{\alpha^{(0)}} \Phi_k(t,X) F_{0}(U(t,X)) dX dt,
\end{equation}
which corresponds to the left-hand side of Eq.~\eqref{eq: Applying Green’s lemma}. Similarly, for each candidate term $s \in \{1, \dots, S\}$, we define the matrix elements $G_{k,s}$ as:
\begin{equation}
\label{eq: G_{k,s}}
G_{k,s} = \int_{0}^{T} \int_{\Omega} (-1)^{|\alpha^{(s)}|} D^{\alpha^{(s)}} \Phi_k(t,X) F_{s}(U(t,X)) dX dt,
\end{equation}
representing the contribution of the $s$-th candidate term on the right-hand side.

By assembling these integrated quantities into the vector $b = (b_1, \dots, b_K)^{\top} \in \mathbb{R}^{K}$ and the matrix $G = (G_{k,s}) \in \mathbb{R}^{K \times S}$, the weak formulation is converted into a computable algebraic system,
\begin{equation}
\label{eq: b=Gxi}
b = G\xi,
\end{equation}
where $\xi = (\xi_1, \dots, \xi_S)^{\top} \in \mathbb{R}^{S}$ is the vector of coefficients to be identified. 

\subsection{The Selection of Test Functions}
\label{sec: The selection of test functions}
The effective selection of test functions is critical to the accuracy of the weak-form PDE discovery process, as these functions determine how local physical information is extracted and aggregated. According to Section \ref{sec: The weak-form PDE}, the selection of test functions satisfies the condition:
\( \Phi\in C_c^{\infty}([0,T]\times\Omega) \).
To construct functions meeting these requirements, we introduce the concept of a bump function $\phi(x)$, a smooth function defined by the following properties,
\begin{equation*}
\phi \in C^{\infty}(\mathbb{R}), \qquad 
\operatorname{supp}(\phi) \subset [-1,1], \qquad
\phi(x) 
\begin{cases}
> 0, & |x| < 1, \\[4pt]
=0, & |x| \ge 1,
\end{cases}
\end{equation*}
where \(C^{\infty}(\mathbb{R})\) denotes the space of infinitely differentiable functions defined on \(\mathbb{R}\), and \(\operatorname{supp}(\phi)\) is the support of \(\phi\). 

Polynomial-type and exponential-type functions are commonly adopted in the numerical implementation of weak-form PDE solvers as two representative families of prototype bump functions \citep{evans2022partial}. The standard polynomial-type prototype is defined as:
\begin{equation}
\label{eq: poly bump}
\phi_{\mathrm{poly}}(x) =
\begin{cases}
(1-x^2)^p, & |x|<1,\\[4pt]
0, & |x|\ge 1,
\end{cases}
\end{equation}
with \(p\in\mathbb{N}\). Since $\phi_{\mathrm{poly}} \in C^{p-1}(\mathbb{R})$, increasing $p$ leads to higher-order continuity at the boundary of its support. The polynomial-type function is preferred for its low computational cost. On the other hand, the exponential-type function achieves maximal ($C^\infty$) smoothness. The standard exponential-type prototype is defined as:
\begin{equation}
\label{eq: exp-bump}
\phi_{\mathrm{exp}}(x)=
\begin{cases}
\displaystyle \exp\!\left(-\frac{1}{1-x^2}\right), & |x|<1,\\[6pt]
0, & |x|\ge 1.
\end{cases}
\end{equation}
In this work, the choice between these two forms depends on the required balance between smoothness and computational efficiency.

Notably, these standard prototypes are restricted to the fixed support $[-1, 1]$. To construct test functions centered at arbitrary positions with flexible scales, we apply an affine transformation to these prototype functions. Specifically, centering at \(x_0\) with radius \(r>0\) yields
\begin{equation*}
    \phi_{x_0,r}(x)=\phi\left(\frac{x-x_0}{r}\right),
\end{equation*}
so that its support satisfies \(\operatorname{supp}(\phi_{x_0,r})\subset[x_0-r,x_0+r]\).
Although this transformation handles arbitrary positioning in a single dimension, most physical systems are defined over multidimensional spacetime domains. To extend the construction to such settings, we adopt a separable extension based on the tensor product of one-dimensional functions. For a spacetime center $(t_0, X_0)$ and a radius $r$, the final multidimensional test function is defined as:
\begin{equation}
\label{eq: final test function}
\Phi_{(t_0,X_0),r}(t,X)
=\phi_{t_0,r}(t)
\prod_{i=1}^d\phi_{x_i,r}\left(x_i\right).
\end{equation}

With the final form of the test functions established in Eq. \eqref{eq: final test function}, the remaining task is to select the parameters $\{\mathbf{c}_k=(t_{0,k}, X_{0,k}), r_k\}$ for each function in the set $\{\Phi_k\}_{k=1}^K$. These parameters define the location and coverage of each test function within the spacetime domain. To this end, we follow the sampling strategy described below. Let the input domain be defined as:
\begin{equation*}
\Omega_T=[0, T] \times \Omega 
= [a_1, b_1] \times [a_2, b_2] \times \cdots \times [a_{d+1}, b_{d+1}] 
\subset \mathbb{R}^{d+1},
\end{equation*}
and denote its shortest side length by
\begin{equation*}
L_{\min} = \min_{1 \le i \le d+1} (b_i - a_i).
\end{equation*}
For the $k$-th test function $\Phi_k$, its support is defined as a ball centered at $\mathbf{c}_k$ with radius $r_k$,
\begin{equation*}
\mathrm{supp}(\Phi_k) = 
\left\{\, \mathbf{x} \in \Omega_T \;\middle|\; 
\|\mathbf{x} - \mathbf{c}_k\| \le r_k 
\,\right\}.
\end{equation*}
The radius \(r_k\) is uniformly sampled as a fraction of the smallest domain length,
\begin{equation*}
r_k \sim \mathcal{U}(\eta_\text{min} L_{\min}, \eta_\text{max} L_{\min}),\qquad  0<\eta_\text{min}<\eta_\text{max} < 0.5,
\end{equation*}
each radius is randomly drawn between $\eta_\text{min}$ and $\eta_\text{max}$ fractions of $L_{\min}$. 
Given the radius \(r_k\), the center \(\mathbf{c}_k\) is then sampled to ensure that the entire support lies within the computational domain,
\begin{equation*}
\mathbf{c}_k \sim \mathcal{U}(\Omega_r), \qquad \Omega_r = [a_1 + r_k,\, b_1 - r_k] 
\times \cdots \times 
[a_{d+1} + r_k,\, b_{d+1} - r_k].
\end{equation*}

\subsection{The Architecture of Weak-PDE-Net}
\label{sec: The architecture of Weak-PDE-Net}
\begin{figure}[!h]
	\centering
	\includegraphics[width=\textwidth]{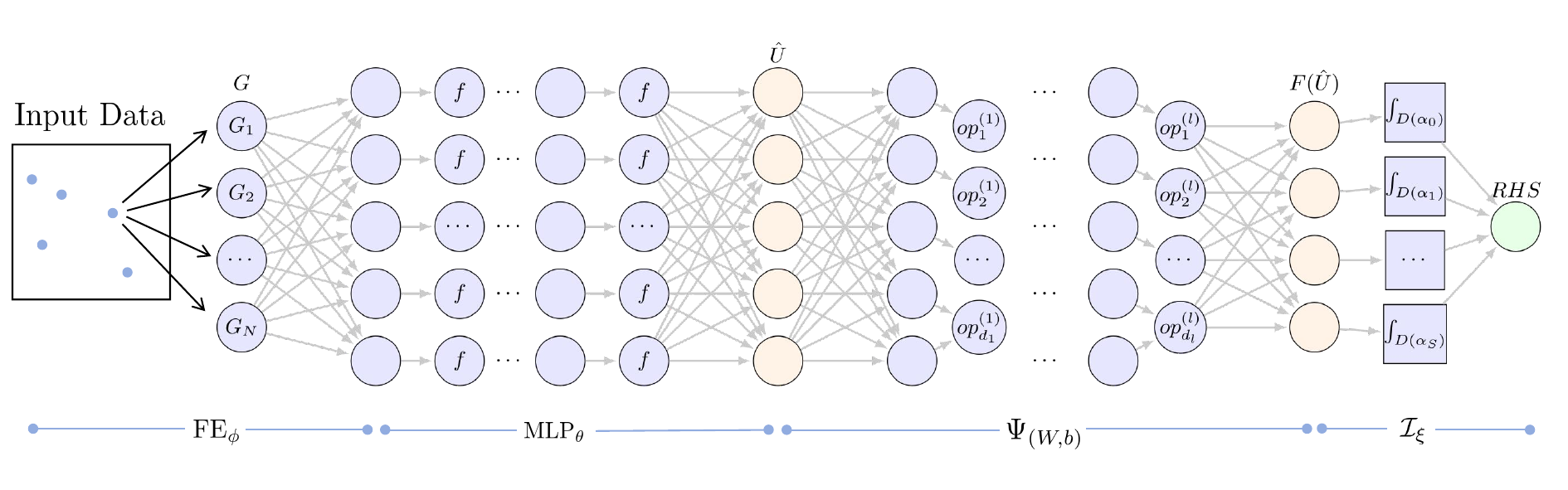}
	\caption{The overall architecture of Weak-PDE-Net.}
	\label{Figure_1}
\end{figure}
As shown in Fig. \ref{Figure_1}, Weak-PDE-Net consists of four primary components: a learnable embedding $\mathrm{FE}_{\phi}$, a lightweight $\mathrm{MLP}_{\theta}$, a symbolic network $\Psi_{(W,b)}$, and an integration module $\mathcal{I}_\xi$. The first two components act as a forward response learner to reconstruct the state $\hat{U}$ from sparse observations. The latter two function as a weak-form PDE generator, where $\Psi_{(W,b)}$ produces the candidate functional terms $F(\hat{U})$, and $\mathcal{I}_\xi$ computes the spacetime integrals to construct the right-hand side ($\mathrm{RHS}$) of Eq. \eqref{eq: b=Gxi}. In the following, we introduce each of these components in detail.

\subsubsection{Learnable Embedding}
\label{sec: Learnable embedding}
Standard MLPs often struggle to capture high-frequency components or sharp gradients due to spectral bias. To address this, we introduce an enhanced learnable Gaussian feature embedding $\mathrm{FE}_{\phi}$ as the input layer. This module maps the raw spacetime coordinates $\mathbf{x}=(t,X) \in \mathbb{R}^{d+1}$ into a $d'$-dimensional feature space, where the $j$-th feature component is constructed as a weighted sum of Gaussian kernels:
\begin{equation}
\label{eq:FE_phi}
\mathrm{FE}_{j}(\mathbf{x}; \phi) = \sum_{i=1}^{N} f_{i,j} G_{i,j}(\mathbf{x}), \quad G_{i,j}(\mathbf{x}) = e^{-\frac{1}{2}(\mathbf{x} - \mu_{i,j})^\top \Sigma_{i,j}^{-1} (\mathbf{x} - \mu_{i,j})}.
\end{equation}
The set of learnable parameters $\phi = \{\mu_{i,j}, \Sigma_{i,j}, f_{i,j}\}$ represents the centers, covariance matrices, and weights of the Gaussians, respectively. By making these parameters learnable, the embedding layer exhibits a crucial locality property. Compared with global basis functions, these Gaussians can adjust their shapes to focus on difficult regions, such as those with high-frequency details. This process works similarly to adaptive mesh refinement in classical numerical methods. By automatically concentrating on complex areas of the solution, the embedding layer helps the model accurately reconstruct the field $\hat{U}$.

\subsubsection{Lightweight MLP}
\label{sec: Lightweight MLP}
While the Gaussian embedding captures local features, a single embedding layer lacks the depth to represent complex nonlinear relationships. To enhance the representation, we follow the embedding with a lightweight multi-layer perceptron (MLP):
\begin{equation}
     \label{eq: mlp}
     \hat{U} = \mathrm{MLP}_{\theta}\big( \mathrm{FE}_{\phi}(\textbf{x}) \big),
\end{equation}
where $\mathrm{MLP}_{\theta}: \mathbb{R}^{d'} \rightarrow \mathbb{R}^{h}$ denotes the network with learnable parameters $\theta$. By processing the embedded features through a few hidden layers, this module significantly increases the model's expressive power. Because the MLP is kept lightweight, it adds minimal computational cost while allowing the framework to accurately reconstruct the field $\hat{U}$ from the high-dimensional feature space.

\subsubsection{Symbolic Network}
\label{sec: Symbolic network}
To construct the $\{F_s(\hat{U})\}_s$ in Eq. \eqref{eq: b=Gxi}, we incorporate a differentiable symbolic network $\Psi_{(W,b)}$ into the Weak-PDE-Net.  Different from traditional fully connected neural networks, the symbolic network employs interpretable operators as activation functions instead of commonly used ones such as $\tanh$ or ReLU. This unique design provides both interpretability and differentiability, enabling the model to discover explicit mathematical forms while maintaining end-to-end training. 

We first formally define the architecture of the symbolic network $\mathcal{A}$, followed by the forward propagation that define the mapping between its layers.
\begin{definition}[Symbolic Network Architecture]
\label{def: symbolic_network_architecture}
The architecture of the symbolic network is defined by the set $\mathcal{A} = \{L, \{\{op_k^{(l)}\}_{k=1}^{d_l}\}_{l=1}^{L}\}$, where $L$ denotes the number of hidden layers. Each layer $l$ consists of $d_l$ operators, where the first $p_l$ operators are unary operators $\{op_k^{(l)}\}_{k=1}^{p_l} \subset \mathcal{O}_u$ and the remaining $d_l - p_l$ operators are binary operators $\{op_k^{(l)}\}_{k=p_l+1}^{d_l} \subset \mathcal{O}_b$.
\end{definition}
Since each unary operator consumes one input node, whereas each binary operator requires two input nodes, the output after linear transformation must provide $p_l+2(d_l-p_l)$ input nodes to satisfy the computational
requirements of this layer. This above output state, denoted
as $y^{(l)}$, therefore contains $p_l+2(d_l-p_l)$ nodes prior to operator application. Following the application of operators, the resulting
vector $z^{(l)}$ contains $d_l$ nodes, the forward propagation of the symbolic network $\Psi_{(W,b)}(\hat{U}): \mathbb{R}^h \rightarrow \mathbb{R}^S$ is formulated as:
\begin{equation}
\begin{aligned}
    &z^{(0)}=\hat{U},\quad \hat{U}\in\mathbb{R}^{h},\\
    &{y}^{(l)} = {W}^{(l)}{z}^{(l-1)} + {b}^{(l)}, \quad l = 1, 2, \dots, L, \\
    &z_k^{(l)}= op_k^{(l)}(y_k^{(l)}), \quad k = 1, 2, \dots, p_l, \\
    &z_k^{(l)}= op_{k+p_l}^{(l)}(y_{p_l+2k-1}^{(l)},y_{p_l+2k}^{(l)}), \quad k = 1,2, \dots, d_l-p_l, \\
    &F(\hat{U})=z^{(l)},\quad z^{(l)}\in\mathbb{R}^{S}.
\end{aligned}
\label{eq: EQL forward}
\end{equation}  
It is important to note that during the training of Weak-PDE-Net, the network architecture $\mathcal{A}$ is not pre-defined. Instead, we employ Differentiable Neural Architecture Search (DNAS) \citep{liu2018darts} to adaptively determine the optimal network architecture from the data. Since different network architectures represent different functional combinations, searching for the most suitable architecture is equivalent to searching for the optimal set of function terms. This data-driven search allows the Weak-PDE-Net to explore a vast space of mathematical expressions, which is essential for achieving open-form PDE discovery.
Furthermore, this adaptive process helps mitigate overfitting by identifying lightweight network architecture, ensuring that the discovered equations remain concise. While the architecture evolves during the DNAS process, once the search converges, the architecture $\mathcal{A}$ becomes fixed and follows the propagation rules defined in Eq. \eqref{eq: EQL forward}. More details of the DNAS will be presented in Section \ref{sec:Differentiable neural architecture search}.

\subsubsection{Integration Module}
\label{sec: Integration module}
To bridge the terms $\{F_s(\hat{U})\}_s$ with the weak-form PDE, the integration module $\mathcal{I}_{\xi}$ transforms the candidate library $F(\hat{U})$ into the right-hand side ($\mathrm{RHS}$) of Eq. \eqref{eq: b=Gxi}. Given a set of test functions $\{\Phi_k\}_{k=1}^{K}$, the integration module $\mathcal{I}_\xi: \mathbb{R}^S \rightarrow \mathbb{R}^K$ computes the weighted spacetime integrals as follows:
\begin{equation}
\label{eq:int_module}
\mathrm{RHS}=(\mathrm{RHS}_k)_{1\le k\le K},\ \mathrm{RHS}_k=\sum_{s=1}^{S}\xi_s\int_{\Omega_T}D^{\alpha^{(s)}}(\Phi_k(\textbf{x}))F_s(\hat{U}(\textbf{x}))d\textbf{x} \approx\sum_{s=1}^S \xi_s\sum_{j} D^{\alpha^{(s)}}(\Phi_k(\textbf{x}_j)) F_s(\hat{U}(\textbf{x}_j)) \Delta \textbf{x}_j,
\end{equation}
where $\textbf{x}=(t,X)$, and $\{\textbf{x}_j\}$ are discrete points on the integration grid, and $\Delta \textbf{x}_j$ denotes the volume element between $\textbf{x}_j$ and $\textbf{x}_{j+1}$.

\subsection{The Loss Function of Weak-PDE-Net}
\label{sec: The loss function of Weak-PDE-Net}
To optimize the parameters of the Weak-PDE-Net, we define a composite loss function. This objective ensures the model reconstructs the field and adheres to the underlying physical laws. The total loss $\mathcal{L}$ is primarily driven by data consistency and the weak-form residual, while a regularization term may be optionally introduced at specific training phase to promote sparsity.

First, the data loss $\mathcal{L}_{\text{data}}$ is essential for ensuring that the reconstructed solution $\hat{U}$ remains consistent with the sparse and noisy observations $(t_i, X_i)$. It is defined as:
\begin{equation}
\label{eq: data loss}
\mathcal{L}_{\text{data}} = \frac{1}{N_\text{data}} \sum_{i=1}^{N_\text{data}} \big(\hat{U}(t_i, X_i) - \tilde{U}(t_i, X_i)\big)^2,
\end{equation}
where \(\hat{U}(t_i, X_i)\) denotes the network-predicted solution and \(U(t_i, X_i)\) represents the ground-truth data.

Second, the weak-form loss $\mathcal{L}_\text{weak}$ is mandatory to ensure the discovered dynamics to satisfy the physical law in an integral sense. Rather than enforcing the PDE at discrete points, this term minimizes the residual of the weak-form equation over the entire spacetime domain. The discovery process seeks an equilibrium between the left-hand side ($\mathrm{LHS}=(\mathrm{LHS}_k)_{1\le k\le K}$) of Eq. \eqref{eq: b=Gxi} and the $\mathrm{RHS}$ in Section \ref{sec: The architecture of Weak-PDE-Net}. This loss is formulated as:
\begin{equation}
\label{eq: Weak_Loss}
\begin{aligned}
\mathcal{L}_{\text{weak}}
&= \sum_{k=1}^K \big( \mathrm{RHS}_k - \mathrm{LHS}_k \big)^2 \\[4pt]
&= \sum_{k=1}^K \Bigg(
    \sum_{s=1}^{S} \xi_s
    \int_{\Omega_T} D^{\alpha^{(s)}}\!\big(\Phi_k(\mathbf{x})\big)
    F_s\!\big(\hat{U}(\mathbf{x})\big)\, d\mathbf{x}
    - 
    \int_{\Omega_T} D^{\alpha^{(0)}}\!\big(\Phi_k(\mathbf{x})\big)
    \hat{U}(\mathbf{x})\, d\mathbf{x}
\Bigg)^{\!2} \\[4pt]
&\approx
\sum_{k=1}^K \Bigg(
    \sum_{s=1}^S \xi_s
    \sum_{j} D^{\alpha^{(s)}}\!\big(\Phi_k(\mathbf{x}_j)\big)
    F_s\!\big(\hat{U}(\mathbf{x}_j)\big)\, \Delta \mathbf{x}_j
    -
    \sum_{j} D^{\alpha^{(0)}}\!\big(\Phi_k(\mathbf{x}_j)\big)
    \hat{U}(\mathbf{x}_j)\, \Delta \mathbf{x}_j
\Bigg)^{\!2}.
\end{aligned}
\end{equation}

Beyond these core objectives, a regularization loss $\mathcal{L}_{\text{reg}}$ can be optionally incorporated depending on the training phase. Once the network architecture is fixed, the loss is introduced to eliminate redundant connections, thereby promoting the discovery of a concise equation. It typically applies $L_1$ penalties to the coefficients $\xi$ in integral module $I_{\xi}$ and the symbolic network parameters $(W, b)$,
\begin{equation}
\label{eq: reg_Loss}
\mathcal{L}_{\text{reg}} =\sum_s |\xi_s|+\sum |W|+\sum|b|.
\end{equation}

In summary, the total training objective is formulated as $\mathcal{L} = \mathcal{L}_{\text{data}} + \lambda_w \mathcal{L}_{\text{weak}} + \lambda_r \mathcal{L}_{\text{reg}}$. While $\mathcal{L}_{\text{data}}$ and $\mathcal{L}_{\text{weak}}$ are always present to guide the training process, the coefficient $\lambda_r$ is non-zero only during the pruning phase to ensure the conciseness of the discovered law. The specific strategy for employing these loss functions across different training phases is detailed in Algorithm \ref{alg: Weak-PDE-Net training}.

\subsection{The Training Strategy of Weak-PDE-Net}
\label{sec: The training strategy of Weak-PDE-Net}

In Section \ref{sec: The architecture of Weak-PDE-Net}, it is mentioned that Weak-PDE-Net consists of a forward response learner ($\mathrm{FE}_{\phi}$ and $\mathrm{MLP}_{\theta}$) and a PDE generator ($\Psi_{(W,b)}$ and $\mathcal{I}_{\xi}$). The training of Weak-PDE-Net proceed the PDE learning and PDE discovery processes simultaneously. Specifically, the PDE learning process utilizes the forward response learner to learn the system response function $\hat{U}$ and reconstruct full grid data, providing the foundation for subsequent evaluation of spacetime integrals. Meanwhile, the PDE discovery process employs the weak-form PDE generator to optimize the equation structure and parameters, determining the explicit mathematical form of the governing equations. 

\begin{figure}[!h]
	\centering
	\includegraphics[width=\textwidth]{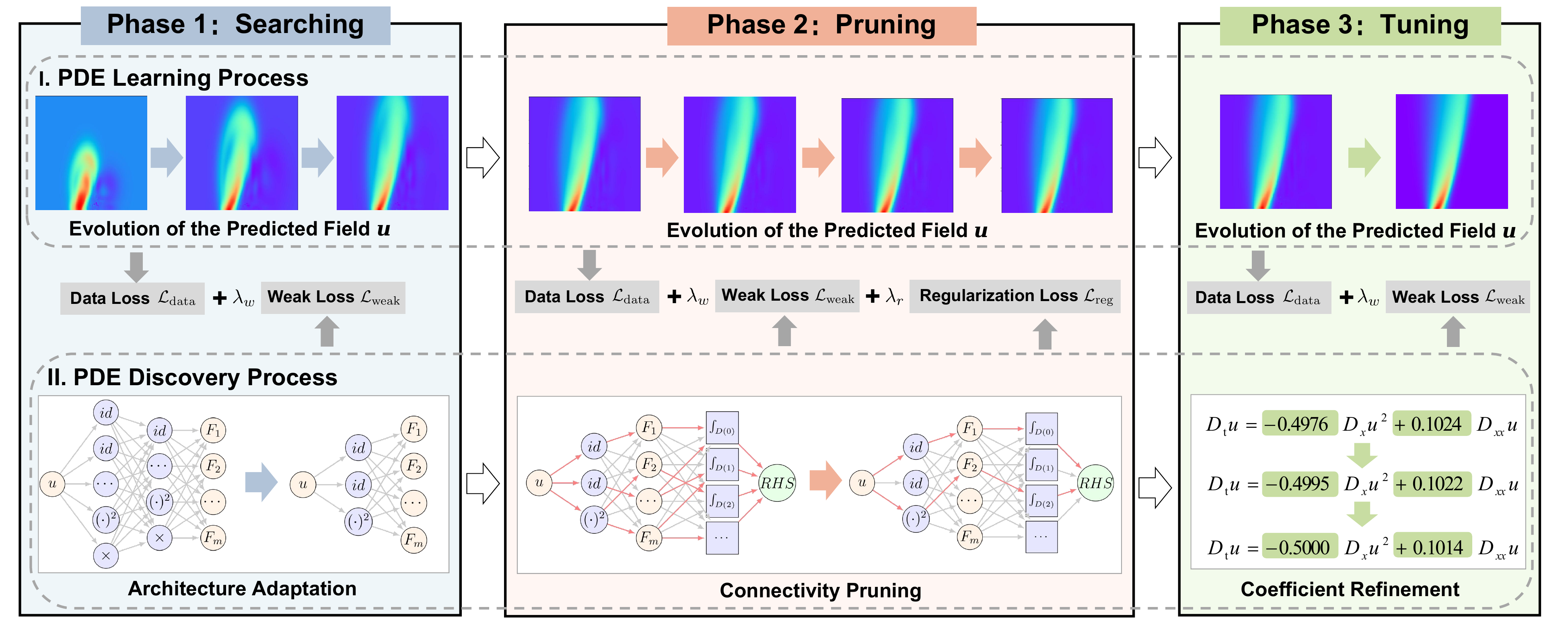}
	\caption{Overview of the training process, exemplified by the Burgers equation. The training progress through two parallel tracks. The PDE Learning Process focuses on learning the PDE solution, consistently reconstructing the full grid data. Concurrently, the PDE Discovery Process evolves from symbolic network architecture adaptation, to connectivity pruning for structural sparsity, and finally to precise coefficient refinement. These two tracks are jointly optimized via a shared loss.
    }
	\label{Figure_2}
\end{figure}
As illustrated in Fig. \ref{Figure_2}, these two tightly coupled processes evolve concurrently across three distinct training phases: differentiable symbolic network architecture search (Searching), structural pruning for equation simplification (Pruning), and sparse regression for coefficient tuning (Tuning). In the following, we detail each of these three phases.

\subsubsection{Differentiable Symbolic Networks Architecture Search}
\label{sec:Differentiable neural architecture search}
In this searching phase, the objective of training is to perform \textbf{architecture adaptation} for the symbolic network, while the architectures and forward propagation of other components (learnable embedding, lightweight MLP and integration module) within WeakPDE-Net remain pre-defined. We conduct this search primarily for two reasons. First, since different network architectures represent distinct functional combinations, searching for the most suitable architecture is equivalent to identifying the optimal set of function terms. This data-driven strategy enables Weak-PDE-Net to explore a vast space of mathematical expressions, which is essential for achieving open-form PDE discovery. Second, this adaptive process helps mitigate overfitting by identifying lightweight network architectures, ensuring that the discovered equations remain concise.

We design a specialized differentiable neural architecture search strategy to automatically identify a suitable and lightweight architecture for the symbolic network. We first define a complete symbolic network with a maximum depth of $L_{\text{max}}$. Each layer in this network incorporates a comprehensive set of operators $\mathcal{O} = \mathcal{O}_u \cup \mathcal{O}_b$, consisting of $m$ distinct operator types. For each type $i \in \{1, \dots, m\}$, a specific number of terms, denoted by $N_i$, is allocated. To enable architecture search during the continuous training process, we introduce two sets of learnable parameters, thereby modifying the standard forward propagation $\Psi_{(W,b)}$ (Eq. \eqref{eq: EQL forward}) of the complete symbolic network into $\Psi_{(W,b,\alpha,\beta)}$. These learnable parameters govern the architecture selection through gradient backpropagation:
\begin{itemize}
    \item The parameters $\alpha=\{\alpha^{(l)}_{i,j}\}$ determine the number of the $i$-th operator type in the $l$-th layer.
    \item The parameters $\beta=\{\beta_l\}$ control the  depth of the network.
\end{itemize}

\begin{figure}[!h]
    \centering
    \includegraphics[width=\textwidth]{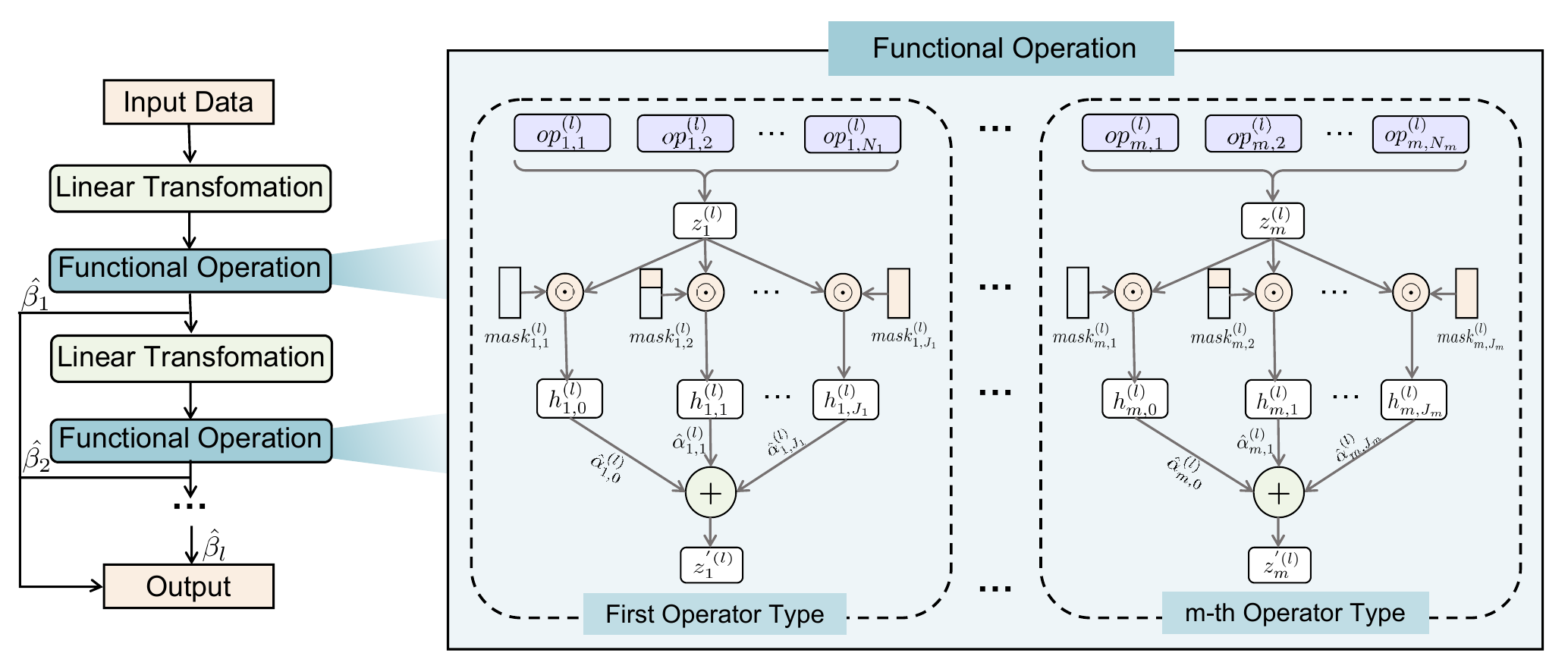}
    \caption{The forward propagation of the symbolic network during the searching phase. The left panel illustrates the overall forward propagation framework; the right panel provides the specific details of the functional operation.}
    \label{Figure_3}
\end{figure}

We then detail how these learnable parameters are integrated into the modified forward propagation. As shown in Fig. \ref{Figure_3}, the forward propagation in each layer consists of two parts: a linear transformation and a functional operation. The linear transformation remains consistent with Eq. \eqref{eq: EQL forward}, whereas the functional operation is no longer a simple application of activation functions. Instead, it incorporates additional operations following the activation stage. We denote the neurons immediately after the activation functions as $z^{(l)}$. To align specific subsets of neurons with their respective operator types, we partition the vector $z^{(l)}$ into contiguous segments:
\begin{equation}
z_i^{(l)} = z^{(l)}\bigg[\bigg(\sum_{\tau=1}^{i-1} N_\tau\bigg)+1 : \sum_{\tau=1}^{i} N_\tau \bigg],
\end{equation}
where $z_i^{(l)} \in \mathbb{R}^{N_i}$ represents the subspace processed by the $i$-th operator type, and the colon $[:]$ denotes the slicing operator that extracts elements from the start index to the end index.

Let $\mathcal{K}_i = \{k_{i,1}, k_{i,2}, \dots, k_{i, J_i}\}$ be the candidate set of active node counts for the $i$-th operator type, where $0 \le k_{i,j} \le N_i$. Here, $J_i$ denotes the total number of candidate choices for the $i$-th operator type. Correspondingly, to implement the $j$-th choice (i.e., retaining exactly $k_{i,j}$ nodes), we define a binary mask vector $\mathrm{mask}_{i,j}^{(l)} \in \{0, 1\}^{N_i}$. Its $p$-th element is defined as:
\begin{equation}
(\mathrm{mask}_{i,j}^{(l)})_p =
\begin{cases}
1, & \text{if } 1 \le p \le k_{i,j}, \\
0, & \text{otherwise}.
\end{cases}
\end{equation}
This mask ensures that only the top $k_{i,j}$ terms are preserved for the $j$-th choice. We then apply these masks to the input segments $z_i^{(l)}$:
\begin{equation}
\label{eq:mask}
h_{i,j}^{(l)} = \mathrm{mask}_{i,j}^{(l)} \odot z_i^{(l)}.
\end{equation}
where $\odot$ denotes the Hadamard product. 

To determine the count of active terms for the each operator type, we compute the weighted sum of these candidates using the learnable parameters $\alpha$. This yields the intermediate output for the $i$-th operator type, denoted as $z_i'^{(l)}$:
\begin{equation}
\label{eq:dnas_alpha}
z_i'^{(l)} = \sum_{j=1}^{J_i} \hat{\alpha}_{i,j}^{(l)} h_{i,j}^{(l)}, \quad \text{with } \hat{\alpha}_{i,j}^{(l)} = \frac{\exp(\alpha_{i,j}^{(l)})}{\sum_{j=1}^{J_i} \exp(\alpha_{i,j}^{(l)})}.
\end{equation}
By concatenating these $z_i'^{(l)}$, the fully processed output of layer $l$ is obtained as $z'^{(l)} = \text{Concat}(z_1'^{(l)}, \dots, z_m'^{(l)})$. Upon completion of the searching phase, we obtain the optimized parameters $\hat{\alpha}$. Based on these values, the discrete count of active terms for the $i$-th operator type is determined by $\hat{n}_i^{(l)} = k_{i, j^*}$, where $j^* = \arg\max_j \hat{\alpha}_{i,j}^{(l)}$.  In other words, the specific number of active terms for the $i$-th operator type in the $l$-th layer is fixed as $\hat{n}_i^{(l)}$, such that the total number of active operators in the $l$-th layer is $d_l = \sum_{i=1}^m \hat{n}_i^{(l)}$. This determines the operator configuration across all layers, denoted as $\{\{{op}_k^{(l)}\}_{k=1}^{d_l}\}_{l=1}^{{L_\text{max}}}$.

In addition to modifying the functional operations within each layer, we also update the final output of the forward propagation. To automatically determine the optimal network depth, the final output of the symbolic network is computed as a weighted sum of the processed outputs $z'^{(l)}$ aggregated across all layers ($l = 1, \dots, L_{\text{max}}$),
\begin{equation}
\label{eq:dnas_beta}
    \mathrm{Output} = \sum_{l=1}^{L_{\text{max}}} \hat{\beta}_l z'^{(l)}, \quad \text{with } \hat{\beta}_{l} = \frac{\exp(\beta_{l})}{\sum_{k=1}^{L_{\text{max}}} \exp(\beta_k)}.
\end{equation}
Upon completion of searching phase, utilizing the optimized parameters $\hat{\beta}$, the optimal number of hidden layers is determined by $L^* = \operatorname{argmax}_l \hat{\beta}_l$.  

Once the searching phase is completed, the symbolic network architecture is reduced and fixed as $\mathcal{A} = \{L^*, \{\{\text{op}_k^{(l)}\}_{k=1}^{d_l}\}_{l=1}^{L^*}\}$, and the forward propagation of the symbolic network is simplified accordingly to $\Psi_{(W,b)}$ in Eq. \eqref{eq: EQL forward}.

\subsubsection{Structural Pruning for Equation Simplification}
\label{sec: Structural Pruning for Equation Simplification}
In this pruning phase, the architecture of Weak-PDE-Net remains fixed, exactly as defined in Section \ref{sec: The architecture of Weak-PDE-Net}. The objective of training is to perform \textbf{connectivity pruning} for the symbolic network and integration module within Weak-PDE-Net, thereby ensuring the conciseness of the final discovered PDEs. Therefore, the regularization loss $\mathcal{L}_\text{reg}$ is introduced to force insignificant weights toward zero, thereby ensuring that the resulting network structure is sufficiently sparse and the discovered equation remains concise. The detailed training procedure is provided in Algorithm \ref{alg: Weak-PDE-Net training}. 

Once the pruning phase is completed, the active terms $\{D^{\alpha^{(s)}}(F_s)\}_s$ in the PDE are determined according to the parameters $(W,b)$ of the symbolic network and the coefficients $\xi$ of the integration module. These identified terms are then projected into the weak form to serve as the basis for subsequent tuning phase. Accordingly, this establishes the symbolic formulation of the candidate library $G$ in Eq. \eqref{eq: b=Gxi}, while its detailed numerical evaluation is deferred to the next section.

\subsubsection{Sparse Regression for Coefficient Tuning}
\label{sec: Sparse Regression for Coefficient Tuning}
In this tuning phase, the objective of training is to perform \textbf{coefficient refinement} for the parameters $\xi$ within the discovered PDE. With the candidate library $G$ identified in Section \ref{sec: Structural Pruning for Equation Simplification}, the PDE discovery task is transformed into a linear regression problem. At this stage, the symbolic network and the integration module are no longer required. Therefore, we retain only the response function learner within the Weak-PDE-Net.

Specifically, we first utilize the response function learner to estimate of the field variables on a regular grid, denoted as $\hat{U}$. With the reconstructed data in hand, we then compute the numerical values of the data matrix ${G} \in \mathbb{R}^{K \times S}$ and the target vector ${b} \in \mathbb{R}^{K}$ as follows:
\begin{equation}
\label{eq: G_k,b_k}
G_{k,s} \approx \sum_{j} (-1)^{|\alpha^{(s)}|} D^{\alpha^{(s)}} \Phi_k(\mathbf{x}_j) F_s(\hat{U}(\mathbf{x}_j)) \Delta \mathbf{x}j, \quad b_k \approx \sum{j} (-1)^{|\alpha^{(0)}|} D^{\alpha^{(0)}} \Phi_k(\mathbf{x}_j) \hat{U}(\mathbf{x}_j) \Delta \mathbf{x}_j.
\end{equation}
Subsequently, with these estimated values, the coefficient set $\xi$ is determined by solving the linear regression problem in Eq. \eqref{eq: b=Gxi} via least-squares estimation \citep{hastie2009elements}. The detailed training process is provided in Algorithm \ref{alg: Weak-PDE-Net training}. Once the tuning phase is completed, we obtain both the completed PDE and the optimized forward response function learner.

\subsection{The Physics-Informed Refinement of Weak-PDE-Net}
\label{sec: The Physics-Informed Refinement of Weak-PDE-Net}
Generalizing open-form PDE discovery to multivariable systems faces two major practical problems. First, the candidate search space expands dramatically. In multivariable systems, we must consider not only each individual variable but also their inter-variable combinations such as $uv$. This leads to a combinatorial explosion of potential terms, making it more difficult for optimization algorithms to identify the correct model.
Second, discovering the governing equation for each state variable independently often leads to physical inconsistency especially in complex-valued systems. Existing methods treat each variable in a system as an isolated task, severing their inherent physical connections. Inevitably, this leads to a systemic failure. The model might identify a very accurate equation for one variable but fail completely for another.

To cope with these two problems, we introduce {physical consistency refinement} to narrow the search space and ensure physical validity:
\begin{itemize}
\item \textbf{Galilean Invariance Constraints}: For systems involving fluid flow or transport, Galilean Invariance requires that physical laws remain consistent across all inertial frames. In our framework, we use this invariance as \textbf{constraints} to filter out spurious terms and supplement the equation with the necessary terms.

\item \textbf{Symmetry Equivariance Hypothesis}: For complex-valued systems, $U(1)$ symmetry requires that the dynamical evolution remains equivariant under phase rotations. In our framework, we treat this equivariance as a structural \textbf{hypothesis} to pair real and imaginary components via anti-symmetry structure.
\end{itemize}

These refinements act as physics-informed priors, effectively effectively narrowing the vast search space and promoting physical consistency. Notably, these physical priors are applied during the tuning phase of the training process, rather than in other phases. In the following, we detail the implementation of these refinements.

\subsubsection{Galilean Invariance Constraints}
\label{sec: Galilean Invariance Constraints}

For systems governing fluid dynamics and transport phenomena within the Newtonian framework, the dynamical laws obey {Galilean Invariance} (GI). This fundamental principle mandates that the mathematical form of equations remains invariant under Galilean transformations between inertial frames.

Once the pruning phase is completed, we obtain an initial candidate library ${G}$ (see Section \ref{sec: Structural Pruning for Equation Simplification}), which is retained throughout the tuning phase. During this stage, GI serves as a fundamental constraint that continuously governs the refinement of ${G}$. Specifically, we eliminate terms that are mathematically plausible but physically invalid, while simultaneously incorporating missing terms necessary for invariance. The following derivation details the selection rules for velocity fields and then extends these principles to vortex dynamics.

Let $X$ and $U(t, X)$ denote the coordinates and the vector field in a stationary reference frame. We then consider a moving frame traveling at a constant velocity $\boldsymbol{c}$. The coordinates $\bar{X}$ and the transformed vector field $\bar{U}(\bar{t}, \bar{X})$ in this moving frame are defined as:
\begin{equation}
    \label{eq:galilean_transform}
    \bar{X} = X - \boldsymbol{c}t, \quad \bar{t} = t, \quad \bar{U}(\bar{t}, \bar{X}) = U(t, X) - \boldsymbol{c},
\end{equation}
where variables with an overbar represent quantities in the moving frame. By applying the chain rule, we can relate the differential operators in the two frames. Since the coordinates in the original frame are related to the transformed coordinates $(\bar{t}, \bar{X})$ in the moving frame as defined in Eq. \eqref{eq:galilean_transform}, the original time derivative $\partial_t$ transforms as follows,
\begin{equation}
\frac{\partial}{\partial t} = \frac{\partial \bar{t}}{\partial t}\frac{\partial}{\partial \bar{t}} + \frac{\partial \bar{X}}{\partial t} \cdot \nabla_{\bar{X}} = \frac{\partial}{\partial \bar{t}} - \boldsymbol{c} \cdot \nabla,
\end{equation}
where ${\nabla}$ denotes the gradient with respect to $\bar{X}$. Note that the spatial gradient remains invariant, i.e., $\nabla{X} = \nabla{\bar{X}} \equiv \nabla$, as the frames differ only by a time-dependent translation. Applying these transformations to a general governing equation $U_t = \mathcal{F}(U, \nabla U, \dots)$, the time derivative becomes,
\begin{equation}
    \frac{\partial U}{\partial t} = \left(\frac{\partial}{\partial \bar{t}} - \boldsymbol{c} \cdot \nabla\right) (\bar{U} + \boldsymbol{c}) = \frac{\partial \bar{U}}{\partial \bar{t}} - (\boldsymbol{c} \cdot \nabla) \bar{U}.
\end{equation}
This reveals that the temporal term $U_t$ generates a frame-dependent convective shift $-(\boldsymbol{c} \cdot \nabla) \bar{U}$. To preserve the form invariance of the equation, the right-hand side $\mathcal{F}$ must retain terms that exactly cancel out this extra shift, while removing any other terms that would introduce uncancelable shift. This requirement gives us clear principles to refine our library $G$:
\begin{itemize}
\item \textbf{Terms to be Retained: } We must keep terms that can neutralize the shift from the time derivative. The most common example is the convective term $\nabla \cdot (U \otimes U)$. In our framework, this term is essential because its transformation generates an additional $\boldsymbol{c} \cdot \nabla U$ term that perfectly offsets the $-\boldsymbol{c} \cdot \nabla U$ shift from the time derivative. Without these terms, the discovered equation would violate Galilean Invariance.
\item \textbf{Terms to be Removed: } We exclude terms dependent on absolute velocity, such as $U$, $|U|^2$, or $U^3$, as they vary with the reference frame. For example, a simple $U$ term transforms into $\bar{U} + \boldsymbol{c}$, introducing a constant velocity $\boldsymbol{c}$ that cannot be canceled or absorbed. Including such terms would make the equation frame-dependent, violating the core principle of Galilean Invariance.
\end{itemize}

Building upon the aforementioned principles, we now extend these Galilean Invariance constraints to vorticity-based equations. Unlike the velocity field $U$, the vorticity field $\boldsymbol{\omega} = \nabla \times U$ is itself a Galilean invariant,
\begin{equation}
    \bar{\boldsymbol{\omega}} = \nabla \times \bar{U} = \nabla \times (U - \boldsymbol{c}) = \boldsymbol{\omega},
\end{equation}
since $\nabla \times \boldsymbol{c} = \mathbf{0}$. However, the time derivative of vorticity still generates an extra term due to the moving reference frame, similar to our previous derivation,
\begin{equation}
    \frac{\partial \boldsymbol{\omega}}{\partial t} = \frac{\partial \boldsymbol{\omega}}{\partial \bar{t}} - (\boldsymbol{c} \cdot \nabla) \boldsymbol{\omega}.
\end{equation}
To ensure the discovered vorticity equation satisfies Galilean Invariance, we apply similar constraints to the library $G_{\omega}$ as derived for the velocity field:
\begin{itemize}
\item \textbf{Terms to be Retained:} We must keep terms that can neutralize the shift from the time derivative. The most common example is the advective flux term $\nabla \cdot (\boldsymbol{\omega} \otimes U)$. In our framework, this term is essential because its transformation generates an additional $(\boldsymbol{c} \cdot \nabla) \boldsymbol{\omega}$ term that perfectly offsets the $-(\boldsymbol{c} \cdot \nabla) \boldsymbol{\omega}$ shift from the time derivative. 
\item \textbf{Terms to be Removed:} We exclude terms that depend on absolute velocity, such as $|U|^2$ or $U \times \boldsymbol{\omega}$. For example, the term $U \times \boldsymbol{\omega}$ transforms into $(\bar{U} + \boldsymbol{c}) \times \boldsymbol{\omega}$, introducing a frame-dependent component proportional to $\boldsymbol{c}$ that cannot be canceled. 
\end{itemize}

In our experiments, we enforce these Galilean Invariance constraints to guide the discovery of the 2D incompressible Navier-Stokes equation in its vorticity form.

\subsubsection{Symmetry Equivariance Hypothesis}
\label{sec: Symmetry Equivariance Hypothesis}
In systems involving complex fields $\psi = u + iv$, such as quantum fluids, the governing laws typically exhibit $U(1)$ equivariance. This principle ensures that the dynamic operators commute with global phase rotations, meaning the underlying physics remains consistent regardless of the choice of phase reference. Specifically, let $\mathcal{N}$ denote the dynamical operator and $g_\phi$ be a global phase rotation such that $g_\phi \psi = e^{i\phi}\psi$, the $U(1)$ equivariance mandates the following commutative relationship,
\begin{equation}
\label{eq: appendix_commute}
\mathcal{N}(g_\phi \psi) = g_\phi \mathcal{N}(\psi).
\end{equation}
This mathematical identity ensures that any phase shift applied to the input field results in an identical rotation of its temporal evolution.

Unlike Galilean Invariance, which serves as universal constraints across all Newtonian systems, phase equivariance is a system-specific property that primarily exists in complex-valued dynamics. Given that $U(1)$ equivariance is not universal across all physical systems, we treat it as an adaptive structural hypothesis rather than a mandatory constraint. This allows our framework to verify the presence of such symmetry directly from the data. For the systems exhibit $U(1)$ equivariance, we derive a specific matching rule for the real and imaginary parts of the equation. This rule requires the two components to be coupled in an anti-symmetric structure,
\begin{equation}
\label{eq: equivariant_identity}
\mathcal{N}_{\text{im}}(u, v) = -\mathcal{N}_{\text{re}}(-v, u).
\end{equation}
The formal derivation from the $U(1)$ equivariance to this anti-symmetric structure is detailed in Appendix \ref{appendix: Detailed Derivation}.

Guided by this Eq. \eqref{eq: equivariant_identity}, we implement a symmetric library augmentation strategy to the candidate library $G=(G_u,G_v)$ in tuning phase. For each candidate term $T(u, v)$ included in the real-part library $G_u$, we automatically construct and include its rotationally consistent counterpart in the imaginary-part library $G_v$. Rather than imposing symmetry as a rigid constraint, we relax it into a  structural hypothesis. By providing these equivariant couplings to the regression algorithm in tuning phase, the model adaptively decides whether to retain them according to the data. This method prioritizes physically consistent structures in symmetric systems while maintaining the flexibility to discover dynamics where the $U(1)$ assumption may not apply.

\section{Experiments}
\label{sec: Experiments}
\subsection{Datasets}
\label{sec: Datasets}
To comprehensively evaluate the generalization capability and robustness of the proposed model, we establish comprehensive benchmarks that covers many different physical systems and mathematical structures.
\begin{itemize}
    \item In the 1D scalar setting ($d=1, h=1$), we select equations with increasing complexity, the second-order Burgers equation (dissipative), the third-order Korteweg-de Vries (dispersive), and the fourth-order Kuramoto-Sivashinsky equation (chaotic). Besides, we include the Chafee-Infante equation to cover reaction-diffusion phenomena.
    \item For multivariable systems ($d=1, h=2$), we consider the Nonlinear Schr\"{o}dinger equation, which is an important model for complex-valued systems.
    \item In 2D space ($d=2$), our benchmarks include the Wave equation and the Sine-Gordon equation. Besides, we test the incompressible Navier-Stokes equation.
\end{itemize}

These diverse benchmarks cover a range of physical phenomena, including diffusion, chaos, and complex fluid coupling, thereby allowing us to test our model against different levels of nonlinearity and derivative orders. More details about these benchmarks can be found in Section \ref{sec: Robustness Analysis}.

\subsection{Metrics}
\label{sec: Metrics}
We evaluate the ability of our method to identify the terms with nonzero coefficients using the true positivity ratio. The true positivity ratio ($\mathrm{TPR}$) is defined as:
\begin{equation*}
\mathrm{TPR} = \frac{\mathrm{TP}}{\mathrm{TP} + \mathrm{FN} + \mathrm{FP}},
\end{equation*}
where $\mathrm{TP}$ denotes the number of correctly identified nonzero coefficients, $\mathrm{FN}$ is the number of coefficients that are incorrectly identified as zero, and $\mathrm{FP}$ is the number of coefficients that are incorrectly identified as nonzero. A higher TPR value indicates a more accurate identification of the nonzero terms. Perfect structural recovery is marked by a TPR of 1, indicating that all nonzero terms are correctly identified without any false positives.

We use two metrics, $E_{\infty}(\hat{{\xi}})$ and $E_{2}(\hat{{\xi}})$, to evaluate the accuracy of the recovered coefficients. The $E_{\infty}(\hat{{\xi}})$ and $E_2(\hat{{\xi}})$ are defined as:
\begin{equation*}
E_{\infty}(\hat{{\xi}}) := \max_{{j: \xi^*_j\neq 0}} \frac{|\hat{\xi}_j - \xi_j^*|}{|\xi_j^*|},\qquad E_2(\hat{{\xi}}) :=
\frac{\|\hat{{\xi}} - {\xi}^*\|_2}{\|{\xi}^*\|_2}.
\end{equation*}
\(E_\infty\) measures the maximum relative error among individual coefficients, reflecting the worst-case recovery accuracy, whereas \(E_2\) quantifies the overall normalized root-mean-square error, indicating the average accuracy of coefficient recovery. Lower values of \(E_{\infty}(\hat{\xi})\) and \(E_2(\hat{\xi})\) indicate more accurate coefficient recovery.

We use reconstruction error \(\mathcal{L}(\hat{U}, U)\) to evaluate the model’s ability to learn (U). The mean squared error (MSE) on the test set \({\hat{U}(t_i, X_i)}_{i=1}^{N_{\text{test}}}\) is defined as
\begin{equation}
\mathcal{L}(\hat{U},U) =
\frac{1}{N_{\text{test}}}
\sum_{i=1}^{N_{\text{test}}}
\bigg(\hat{U}(t_i, X_i) - U(t_i, X_i)\bigg)^2,
\end{equation}
where \(\hat{U}(t_i, X_i)\) denotes the model prediction, \(U(t_i, X_i)\) is the ground truth, and \(N_{\text{test}}\) represents the number of test samples.

\subsection{Baselines}
\label{sec: Baselines}

To evaluate the performance of our proposed method, we select Weak-PDE-LEARN \citep{stephany2024weak} as the baseline.

\begin{itemize}
    \item \textbf{Weak-PDE-LEARN: }  This method employs Rational Neural Networks (RatNNs) to approximate the system response function while simultaneously identifying governing equations via an adaptive integral-based loss. Despite its efficacy, it remains constrained by the need for a pre-defined candidate library.
\end{itemize}

Additionally, we used the standard hyperparameters provided in the open-source implementations for Weak-PDE-LEARN\footnote{\url{https://github.com/punkduckable/Weak_PDE_LEARN.git}}.

\subsection{Settings}
\label{sec: Settings}
We assume the data points are randomly distributed across the full spatio-temporal grid. To measure the sparsity, we define the sampling ratio $r$ as follows,
\begin{equation}
    r = \frac{N_{\text{data}}}{N_{\text{total}}},
\end{equation}
where $N_{\text{data}}$ is the number of randomly sampled points and $N_{\text{total}} = N_x \times N_t$ represents the total size of the spatio-temporal grid.

For the sampled data, the noisy observations are generated as,
\begin{equation*}
    \tilde{U} = U + \sigma_\text{NR}|U|_\text{RMS}\boldsymbol{\epsilon},
\end{equation*}
where \(U\) denotes the noise-free data, \(\boldsymbol{\epsilon}\) is a Gaussian noise with zero mean and unit variance. The parameter \(\sigma_\text{NR}\) controls the relative noise level with respect to the RMS magnitude of the clean data. 

The computing infrastructure and hyperparameter settings \textcolor{black}{used in the} experiments are \textcolor{black}{provided} in Appendix \ref{Appendix: Computing Infrastructure and Hyperparameter Settings}.

\section{Results}
\label{sec: Results}
\subsection{Robustness Analysis}
\label{sec: Robustness Analysis}
Since the governing equations exhibit different dynamic characteristics, their sensitivities to data sparsity and noise also differ. To provide a comprehensive evaluation, we present the results for each equation separately, showing Weak-PDE-Net's performance at different sample ratio and noise levels.

\subsubsection{Burgers Equation}
\label{sec: Burgers Equation}

The Burgers equation originated in fluid mechanics as a model of the fundamental dynamics of viscous flows \citep{bateman1915some}. In the intervening years, the Burgers equation has become a classic model for studying the balance between non-linear shock formation and viscous smoothing. It is now widely applied across diverse fields, including nonlinear acoustics, gas dynamics, and traffic flow \citep {basdevant1986spectral}. 

\begin{table}[!b]
\centering
\caption{Performance evaluation on Burgers equation.}
\label{tab:results_burgers}
\resizebox{0.85\textwidth}{!}{
\begin{tabular}{cc c ccc l}
\toprule
$\sigma_{\text{NR}}$ & $r$ & TPR & ${E_{\infty}}$ & ${E_2}$ & ${\mathcal{L}(\hat{U},U)}$ & Discovered Equation \\
\midrule
\multicolumn{7}{c}{\textbf{Varying Sample Ratio}} \\
\midrule
0\% & 2.5\% & 1.00 & 0.0589 & 0.0557 & $2.97\times10^{-4}$ & $\partial_t u = -0.4722\partial_x u^2 + 0.1059\partial_{xx}u$ \\
0\% & 5.0\% & 1.00 & 0.0184 & 0.0182 & $1.54\times10^{-4}$ & $\partial_t u = -0.4908\partial_x u^2 + 0.0992\partial_{xx}u$ \\
0\% & 10\%  & 1.00 & 0.0096 & 0.0095 & $7.14\times10^{-5}$ & $\partial_t u = -0.5048\partial_x u^2 + 0.1007\partial_{xx}u$ \\
0\% & 25\%  & 1.00 & 0.0213 & 0.0047 & $5.07\times10^{-5}$ & $\partial_t u = -0.5011\partial_x u^2 + 0.1021\partial_{xx}u$ \\
0\% & 50\%  & 1.00 & 0.0012 & 0.0008 & $4.84\times10^{-5}$ & $\partial_t u = -0.4996\partial_x u^2 + 0.1001\partial_{xx}u$ \\
\midrule
\multicolumn{7}{c}{\textbf{Varying Noise Level ($r=10\%$)}} \\
\midrule
20\%  & 10\% & 1.00 & 0.0127 & 0.0125 & $3.14\times10^{-4}$ & $\partial_t u = -0.4936\partial_x u^2 + 0.1003\partial_{xx}u$ \\
40\%  & 10\% & 1.00 & 0.0467 & 0.0453 & $9.74\times10^{-4}$ & $\partial_t u = -0.4774\partial_x u^2 + 0.0953\partial_{xx}u$ \\
60\%  & 10\% & 1.00 & 0.0792 & 0.0567 & $2.69\times10^{-3}$ & $\partial_t u = -0.4722\partial_x u^2 + 0.1079\partial_{xx}u$ \\
80\%  & 10\% & 1.00 & 0.1627 & 0.1233 & $4.59\times10^{-3}$ & $\partial_t u = -0.4393\partial_x u^2 + 0.0837\partial_{xx}u$ \\
100\% & 10\% & 1.00 & 0.1844 & 0.1814 & $6.45\times10^{-3}$ & $\partial_t u = -0.4078\partial_x u^2 + 0.1073\partial_{xx}u$ \\
\bottomrule
\end{tabular}
}
\end{table}

As a classic test case for evaluating PDE discovery methods, we consider the 1D viscous Burgers equation:
\begin{equation}
    \label{burgers Equation}
    \partial_t u = -0.5\partial_{x}u^2 + \nu \partial_{xx} u,
\end{equation}
where $\nu$ is the diffusion coefficient.

Following the setting of \citet{rudy2017data}, we focus on the Burgers equation in the dissipative setting ($\nu = 0.1$). The dataset is defined on the domain $(t,x) \in [0, 10]\times [-8, 8] $ with a $101 \times 256$ grid resolution. We evaluated our method under two challenging conditions. We first varied the sampling ratio $r$ from $2.5\%$ to $50\%$ to test for sparsity. And then we fixed $r=10\%$ and then introduced Gaussian noise ranging from $20\%$ up to $100\%$.

Table \ref{tab:results_burgers} summarizes the identification results and corresponding metrics for these cases. Our method achieves a TPR of 1.00 and identifies the correct equation structure in all cases. Weak-PDE-Net identifies the equation from data with up to $100\%$ noise. The accuracy of the identified coefficients for the diffusion and non-linear terms decreases as the noise level increases. Additionally, Weak-PDE-Net is able to recover the governing equation in low-data setting with a sample ratio of $r=2.5\%$. These results demonstrate that our method achieves robust discovery of PDEs from sparse and noisy observations.

\subsubsection{Korteweg-de Vries}
\label{sec: KdV Equation}
The Korteweg-De Vries (KdV) equation is a third-order non-linear PDE. It was originally proposed to describe the motion of 1D and shallow water waves \citep{korteweg1895xli}. The KdV equation also finds application in plasma physics and other non-linear systems \citep{zabusky1965interaction}.

The KdV equation introduces third-order derivative terms to the system: 
\begin{equation}
    \label{KdV Equation}
    \partial_t u = -3\partial_{x}u^2 - \partial_{xxx} u,
\end{equation}
where $u(t,x)$ denotes the wave amplitude.

We used the dataset produced by \citet{rudy2017data}. The problem is defined on the domain $(t,x) \in [0, 20]\times  [-30, 30]$ with a spatio-temporal resolution of $ 201\times 512$ grid points. For the KdV equation, we conducted sparsity experiments by varying the sampling ratio across the set $r \in \{2.5\%, 5.0\%, 10\%, 25\%, 50\%\}$. Subsequently, to evaluate noise robustness, we fixed the sampling ratio at $r=25\%$ and adjusted the noise level $\sigma_{\text{NR}}$ within the range $\{20\%, 40\%, 60\%, 80\%, 100\%\}$.

Table \ref{tab:results_kdv} summarizes the identification performance on the KdV equation, highlighting our method's robustness against both data sparsity and noise. In the sparsity tests, our method Weak-PDE-Net successfully recovers the correct equation structure and accurate coefficients even with a sampling ratio as low as $2.5\%$. Under noisy conditions ($r=25\%$), Weak-PDE-Net maintains a perfect identification rate (TPR=1.00) up to an $100\%$ noise level, demonstrating significant robustness. 

\begin{table}[!h]
\centering
\caption{Performance evaluation on the KdV equation.}
\label{tab:results_kdv}
\resizebox{0.85\textwidth}{!}{
\begin{tabular}{cc c ccc l}
\toprule
$\sigma_{\text{NR}}$ & $r$ & TPR & ${E_{\infty}}$ & ${E_2}$ & ${\mathcal{L}(\hat{U},U)}$ & Discovered Equation \\
\midrule
\multicolumn{7}{c}{\textbf{Varying Sample Ratio}} \\
\midrule
0\% & 2.5\% & 1.00 & 0.0719 & 0.0259 & $1.38\times10^{-4}$ & $\partial_t u = -3.0394\partial_x u^2 - 0.9281\partial_{xxx}u$ \\
0\% & 5.0\% & 1.00 & 0.0373 & 0.0139 & $1.48\times10^{-4}$ & $\partial_t u = -2.9766\partial_x u^2 - 1.0373\partial_{xxx}u$ \\
0\% & 10\%  & 1.00 & 0.0254 & 0.0120 & $4.12\times10^{-5}$ & $\partial_t u = -2.9718\partial_x u^2 - 1.0254\partial_{xxx}u$ \\
0\% & 25\%  & 1.00 & 0.0376 & 0.0119 & $6.49\times10^{-5}$ & $\partial_t u = -3.0013\partial_x u^2 - 1.0376\partial_{xxx}u$ \\
0\% & 50\%  & 1.00 & 0.0274 & 0.0103 & $1.47\times10^{-5}$ & $\partial_t u = -3.0174\partial_x u^2 - 1.0274\partial_{xxx}u$ \\
\midrule
\multicolumn{7}{c}{\textbf{Varying Noise Level ($r=25\%$)}} \\
\midrule
20\%  & 25\% & 1.00 & 0.0159 & 0.0151 & $2.11\times10^{-4}$ & $\partial_t u = -2.9523\partial_x u^2 - 0.9987\partial_{xxx}u$ \\
40\%  & 25\% & 1.00 & 0.0207 & 0.0067 & $2.49\times10^{-4}$ & $\partial_t u = -2.9960\partial_x u^2 - 1.0207\partial_{xxx}u$ \\
60\%  & 25\% & 1.00 & 0.0772 & 0.0283 & $3.72\times10^{-4}$ & $\partial_t u = -2.9549\partial_x u^2 - 0.9228\partial_{xxx}u$ \\
80\%  & 25\% & 1.00 & 0.0713 & 0.0298 & $2.09\times10^{-4}$ & $\partial_t u = -2.9384\partial_x u^2 - 0.9287\partial_{xxx}u$ \\
100\% & 25\% & 1.00 & 0.1240 & 0.0401 & $1.04\times10^{-3}$ & $\partial_t u = -3.0264\partial_x u^2 - 1.1240\partial_{xxx}u$ \\
\bottomrule
\end{tabular}
}
\end{table}

\subsubsection{Kuramoto-Sivashinsky Equation}
\label{sec: KS Equation}
The Kuramoto-Sivashinsky (KS) equation is a fourth-order PDE. It was originally derived to model instabilities in laminar flame fronts \citep{michelson1977nonlinear}. The KS equation is also used to describe plasma physics, chemical physics, and combustion dynamics \citep{hyman1986kuramoto}. When the diffusion coefficient is negative, the solutions to the KS equation can exhibit chaotic behavior.

We employ the fourth-order KS equation as follows as a test case:
\begin{equation}
    \label{KS Equation}
    \partial_t u =  - 0.5\partial_x u^2 -\partial_{xx} u - \partial_{xxxx} u,
\end{equation}
where the interplay between the instability-inducing negative diffusion $-\partial_{xx}u$ and the stabilizing fourth-order dissipation $-\partial_{xxxx}u$ generates rich chaotic dynamics.

Following the setup in Weak-PDE-LEARN \citep{stephany2024weak}, we analyzed the system on the domain $(t,x) \in [0, 50] \times [-10, 10]$ with a spatio-temporal resolution of $ 251\times 256$ grid points. The simulation employs periodic boundary conditions, with the initial state being $u(x,0) = -\sin(\pi x/10)$. The sampling ratio and noise configuration are identical to those used in the KdV equation in our experiments.
\begin{table}[!h]
\centering
\caption{Performance evaluation on the KS equation. }
\label{tab:results_ks}
\resizebox{0.95\textwidth}{!}{
\begin{tabular}{cc c ccc c}
\toprule
$\sigma_{\text{NR}}$ & $r$ & TPR & ${E_{\infty}}$ & ${E_2}$ & ${\mathcal{L}(\hat{U},U)}$ & Discovered Equation \\
\midrule
\multicolumn{7}{c}{\textbf{Varying Sample Ratio}} \\
\midrule
0\% & 2.5\% & 1.00 & 0.0173 & 0.0149 & $4.47\times10^{-3}$ & $\partial_t u = -0.4916\partial_x (u)^2 - 1.0115\partial_{xx} u - 1.0173\partial_{xxxx} u$ \\
0\% & 5.0\% & 1.00 & 0.0076 & 0.0045 & $2.13\times10^{-3}$ & $\partial_t u = -0.4962\partial_x(u^2) - 0.9946\partial_{xx} u - 0.9989\partial_{xxxx} u$ \\
0\% & 10\%  & 1.00 & 0.0120 & 0.0042 & $2.77\times10^{-3}$ & $\partial_t u = -0.4940\partial_x(u^2) - 0.9998\partial_{xx} u - 1.0012\partial_{xxxx} u$ \\

0\% & 25\%  & 1.00 & 0.0055 & 0.0021 & $1.66\times10^{-3}$ & $\partial_t u = -0.5027 \partial_x(u^2) - 0.9985\partial_{xx} u - 0.9995\partial_{xxxx} u$ \\
0\% & 50\%  & 1.00 & 0.0032 & 0.0029 & $1.01\times10^{-3}$ & $\partial_t u = -0.4998 \partial_x(u^2) - 1.0032\partial_{xx} u - 1.0030\partial_{xxxx} u$ \\

\midrule
\multicolumn{7}{c}{\textbf{Varying Noise Level ($r=25\%$)}} \\
\midrule
20\%  & 25\% & 1.00 & 0.0054 & 0.0034 & $4.34\times10^{-3}$ & $\partial_t u = -0.5027\partial_x(u^2) - 1.0022\partial_{xx} u - 0.9964\partial_{xxxx} u$ \\
40\%  & 25\% & 1.00 & 0.0165 & 0.0113 & $8.63\times10^{-3}$ & $\partial_t u = -0.4968\partial_x(u^2) - 0.9979\partial_{xx} u - 0.9835\partial_{xxxx} u$ \\
60\%  & 25\% & 1.00 & 0.0405 & 0.0288 & $3.57\times10^{-2}$ & $\partial_t u = -0.4860\partial_x(u^2) - 0.9943\partial_{xx} u - 0.9595\partial_{xxxx} u$ \\
80\%  & 25\% & 1.00 & 0.0622 & 0.0314 & $5.74\times10^{-2}$ & $\partial_t u = -0.4689\partial_x(u^2) - 1.0338\partial_{xx} u - 1.0102\partial_{xxxx} u$ \\
100\% & 25\% & 1.00 & 0.0882 & 0.0678 & $1.18\times10^{-1}$ & $\partial_t u =-0.4711\partial_x(u^2) - 0.9584\partial_{xx} u - 0.9118\partial_{xxxx} u$ \\
\bottomrule
\end{tabular}
}
\end{table}

The results for the KS equation are detailed in Table \ref{tab:results_ks}. As shown in the table, our method Weak-PDE-Net successfully identifies the correct equation structure in all cases, maintaining a TPR of 1.00. Weak-PDE-Net recovers the governing laws from limited data with a sample ratio as low as $r=2.5\%$. Regarding robustness to noise, the identified coefficients remain close to their true values even under high noise levels. For instance, at $60\%$ noise, the recovered coefficient for the fourth-order term is $-0.9595$, which is a slight deviation from the true value of $-1.0$. While the reconstruction error $\mathcal{L}$ increases as the noise level rises, reaching $1.18\times10^{-1}$ at $100\%$ noise, Weak-PDE-Net consistently filters out non-physical terms and identifies the correct equation structure. These results demonstrate that Weak-PDE-Net is capable of identifying higher-order governing equations from sparse and noisy observations.

\subsubsection{Chafee-Infante Equation}
\label{sec: Chafee-Infante Equation}
The Chafee-Infante (CI) equation is a reaction-diffusion PDE used to study bifurcation phenomena \citep{chafee1974bifurcation}. While the Burgers equation involves shock formation and the KdV equation describes solitons, the CI equation models the evolution of phase boundaries and steady-state transitions. Unlike the KS equation, which exhibits spatio-temporal chaos, this system is governed by the competition between diffusion and non-linear reaction terms. The equation also finds application in material transport and particle diffusion \citep{tahir2021exact}.

We further extend our evaluation using the CI equation:
\begin{equation}
    \label{Chafee-Infante Equation}
    \partial_t u  =  \partial_{xx} u - u + u^3.
\end{equation}

Following the configuration in \citet{xu2020dlga}, we analyze the system on the spatio-temporal domain $(t,x) \in [0, 0.5] \times [0, 3]$, discretized into a $200 \times 301$ grid. The simulation enforces zero Dirichlet boundary conditions $u(0,t)=u(3,t)=0$ and starts from a sinusoidal initial state $u(x,0) = \sin(x)$.

For the CI equation, we conducted sparsity experiments by varying the sampling ratio over $r \in \{2.5\%, 5.0\%, 10\%\\, 25\%, 50\%\}$. Subsequently, to evaluate noise robustness, we fixed the sampling ratio at $r=25\%$ and adjusted the noise level $\sigma_{\text{NR}}$ within the range $\{1\%, 5\%, 10\%\}$. 

\begin{table}[!b]
\centering
\caption{Performance evaluation on the CI equation. The symbol $--$ indicates that forward learning is not required under full sampling conditions; therefore, the reconstruction MSE is not applicable.}
\label{tab:results_ci}
\resizebox{0.95\textwidth}{!}{
\begin{tabular}{cc c ccc l}
\toprule
$\sigma_{\text{NR}}$ & $r$ & TPR & ${E_{\infty}}$ & ${E_2}$ & ${\mathcal{L}(\hat{U},U)}$ & Discovered Equation \\
\midrule
\multicolumn{7}{c}{\textbf{Varying Sample Ratio}} \\
\midrule
0\% & 2.5\% & 1.00 & 0.2149 & 0.1523 & $9.25\times10^{-4}$ & $\partial_t u = 0.9298u^3 - 0.7851u + 0.8643\partial_{xx}u$ \\
0\% & 5.0\% & 1.00 & 0.1520 & 0.1189 & $1.93\times10^{-5}$ & $\partial_t u = 0.9200u^3 - 0.8480u + 0.8865\partial_{xx}u$ \\
0\% & 10\%  & 1.00 & 0.1139 & 0.0968 & $2.37\times10^{-5}$ & $\partial_t u = 0.9314u^3 - 0.8861u + 0.8977\partial_{xx}u$ \\
0\% & 25\%  & 1.00 & 0.1224 & 0.0955 & $6.06\times10^{-6}$ & $\partial_t u = 0.9317u^3 - 0.8776u + 0.9123\partial_{xx}u$ \\
0\% & 50\%  & 1.00 & 0.1121 & 0.0926 & $1.00\times10^{-5}$ & $\partial_t u = 0.9297u^3 - 0.8879u + 0.9094\partial_{xx}u$ \\
0\% & 100\%  & 1.00 & 0.0003 & 0.0003 & $--$ & $\partial_t u = 0.9998u^3 - 0.9997u + 0.9997\partial_{xx}u$ \\

\midrule
\multicolumn{7}{c}{\textbf{Varying Noise Level}} \\
\midrule
1\%  & 25\% & 1.00 & 0.9916 & 0.6093 & $1.15\times10^{-3}$ & $\partial_t u = 0.6397u^3-1.0254u+0.0084\partial_{xx} u$ \\
5\%  & 25\%  & 1.00 & 0.9950& 0.6158 &$4.81\times10^{-3}$ & $\partial_t u = 0.6451u^3-0.8536u+0.0050\partial_{xx} u$ \\
10\%  & 25\%  & 0.50 & 1.0000& 0.6391 & $6.67\times10^{-3}$ & $\partial_t u = 0.5796u^3-0.7799u$ \\
10\%  & 100\%  & 0.20 & 1.0000& 0.8369 & $--$ & $\partial_t u = 1.2557u^3-0.0189\partial_{x}u^3+0.1888\partial_{x}u$ \\
\bottomrule
\end{tabular}
}
\end{table}

Table \ref{tab:results_ci} summarizes the results of the CI equations. Under noise-free conditions, our method exhibits excellent robustness to data sparsity, accurately recovering the equation structure (TPR=1.00) even at sample ratio as low as $r=2.5\%$. However, compared to other systems (e.g., the Burgers equation or the KS equation), the identification results for the CI equation exhibit significantly higher sensitivity to noise. Under noisy conditions, the accuracy of the diffusion term $\partial_{xx}u$ drops rapidly. For example, at a noise level of $1\%$, the identified coefficients are suppressed to $\approx 0.008$, almost disappearing compared to the true value of 1.0.

To investigate the underlying cause of this higher sensitivity, we examine the performance at higher noise levels ($\sigma_{\text{NR}}=10\%$). A notable anomaly arises where the full-sampling case ($r=100\%$) actually performs worse (TPR=0.20) than the sparse case ($r=25\%$, TPR=0.50). While Weak-PDE-Net identifies the correct physics on clean data, the introduction of high noise forces the weak-form integral to rely more heavily on its smoothing effect. The integral operation acts as a low-pass filter. While generally beneficial for noise suppression, it tends to over-smooth the sharp, localized variations characteristic of the CI equation. In this process, the integral cannot distinguish between random noise and actual high-frequency signals. Consequently, the subtle signal of the diffusion term is washed out, causing the model to identify spurious terms like $\partial_x u^3$. This highlights a trade-off in weak-form methods, while the smoothing effect effectively suppresses noise, it may inevitably attenuate critical physical features when the noise level is too high.

\subsubsection{Nonlinear Schr\"{o}dinger Equation}
\label{sec: NLS equation}
To ensure a comprehensive evaluation, particularly for complex-valued systems, we introduce the Nonlinear Schr\"{o}dinger (NLS) equation as a test case,
\begin{equation}
    \label{NLS Equation}
    i \psi_t = -\frac{1}{2}\psi_{xx} - |\psi|^2 \psi.
\end{equation}
where $\psi(t,x)$ is a complex-valued field. Originally derived to describe quantum mechanical systems, the NLS equation has become a universal model for the evolution of slowly varying wave packets in weakly nonlinear dispersive media \citep{sulem2007nonlinear}. It is widely applied in fields such as fiber optics, deep-water gravity waves, and plasma physics \citep{zakharov1968stability}.

For the purpose of PDE Discovery, we decomposed the complex field into its real and imaginary parts, $\psi(t,x) = u(t,x) + i v(t,x)$, resulting in a coupled system of real-valued PDEs:
\begin{equation}
    \label{NLS Coupled}
    \left\{
    \begin{aligned}
        u_t &= \frac{1}{2}\partial_{xx}v + (u^2+v^2)v, \\
        v_t &= -\frac{1}{2}\partial_{xx}u - (u^2+v^2)u.
    \end{aligned}
    \right.
\end{equation}
We used the dataset produced by \citet{brunton2016discovering}. The problem is defined on the domain $(t,x) \in [0, \pi] \times [-5, 5]$ with a spatio-temporal resolution of $251 \times 256$ grid points. 

\begin{table}[!H]
\centering
\caption{Performance evaluation on the NLS equation.}
\label{tab:results_nls}
\resizebox{0.85\textwidth}{!}{
\begin{tabular}{cc c ccc l}
\toprule
$\sigma_{\text{NR}}$ & $r$ & TPR & ${E_{\infty}}$ & ${E_2}$ & ${\mathcal{L}(\hat{U},U)}$ & Discovered Equation \\
\midrule
\multicolumn{7}{c}{\textbf{Varying Sample Ratio}} \\
\midrule
0\% & 10\% & 
\begin{tabular}{@{}c@{}} $1.00$ \\ $1.00$ \end{tabular} & 
\begin{tabular}{@{}c@{}} $0.0124$ \\ $0.0101$ \end{tabular} & 
\begin{tabular}{@{}c@{}} $0.0096$ \\ $0.0073$ \end{tabular} & 
\begin{tabular}{@{}c@{}} $1.64\times10^{-4}$ \\ $1.57\times10^{-4}$ \end{tabular} & 
$\begin{cases} \partial_t u =0.4955 \partial_{xx} v +0.9876u^2 v + 1.0056v^3 \\ \partial_t v = -0.4993 \partial_{xx} u - 0.9899u v^2 - 1.0041u^3 \end{cases}$ \\ \addlinespace

0\% & 25\% & 
\begin{tabular}{@{}c@{}} $1.00$ \\ $1.00$ \end{tabular} & 
\begin{tabular}{@{}c@{}} $0.0049$ \\ $0.0034$ \end{tabular} & 
\begin{tabular}{@{}c@{}} $0.0034$ \\ $0.0025$ \end{tabular} & 
\begin{tabular}{@{}c@{}} $3.17\times10^{-5}$ \\ $2.69\times10^{-5}$ \end{tabular} & 
$\begin{cases} \partial_t u =  0.4990 \partial_{xx} v + 0.9951u^2 v + 1.0008v^3  \\ \partial_t v = -0.4991 \partial_{xx} u - 1.0034u v^2 - 0.9987u^3 \end{cases}$ \\ \addlinespace

0\% & 37.5\% & 
\begin{tabular}{@{}c@{}} $1.00$ \\ $1.00$ \end{tabular} & 
\begin{tabular}{@{}c@{}} $0.0043$ \\ $0.0022$ \end{tabular} & 
\begin{tabular}{@{}c@{}} $0.0031$ \\ $0.0015$ \end{tabular} & 
\begin{tabular}{@{}c@{}} $1.21\times10^{-5}$ \\ $1.06\times10^{-5}$ \end{tabular} & 
$\begin{cases} \partial_t u = 0.4988\partial_{xx}v+1.0043u^2v+0.9986v^3 \\ \partial_t v = -0.5001 \partial_{xx} u - 1.0022u v^2 - 0.9995u^3 \end{cases}$ \\ \addlinespace

0\% & 50\% & 
\begin{tabular}{@{}c@{}} $1.00$ \\ $1.00$ \end{tabular} & 
\begin{tabular}{@{}c@{}} $0.0041$ \\ $0.0024$ \end{tabular} & 
\begin{tabular}{@{}c@{}} $0.0021$ \\ $0.0017$ \end{tabular} & 
\begin{tabular}{@{}c@{}} $1.56\times10^{-5}$ \\ $1.12\times10^{-5}$ \end{tabular} & 
$\begin{cases} \partial_t u = 0.4980 \partial_{xx} v + 0.9977u^2 v + 0.9998v^3 \\ \partial_t v = -0.4997 \partial_{xx} u - 0.9976u v^2 - 1.0006u^3 \end{cases}$ \\ 
\midrule
\multicolumn{7}{c}{\textbf{Varying Noise Level ($r=25\%$)}} \\
\midrule

20\% & 25\% & 
\begin{tabular}{@{}c@{}} $1.00$ \\ $1.00$ \end{tabular} & 
\begin{tabular}{@{}c@{}} $0.0142$ \\ $0.0305$ \end{tabular} & 
\begin{tabular}{@{}c@{}} $0.0100$ \\ $0.0219$ \end{tabular} & 
\begin{tabular}{@{}c@{}} $1.71\times10^{-3}$ \\ $1.64\times10^{-3}$ \end{tabular} & 
$\begin{cases} \partial_t u = 0.4998 \partial_{xx} v + 1.0142u^2 v + 0.9953v^3 \\ \partial_t v = -0.4834 \partial_{xx} u - 0.9976u v^2 - 0.9871u^3 \end{cases}$ \\ \addlinespace

40\% & 25\% & 
\begin{tabular}{@{}c@{}} $1.00$ \\ $1.00$ \end{tabular} & 
\begin{tabular}{@{}c@{}} $0.0750$ \\ $0.1376$ \end{tabular} & 
\begin{tabular}{@{}c@{}} $0.0554$ \\ $0.1242$ \end{tabular} & 
\begin{tabular}{@{}c@{}} $7.23\times10^{-3}$ \\ $6.95\times10^{-3}$ \end{tabular} & 
$\begin{cases} \partial_t u = 0.5133 \partial_{xx} v+1.0750u^2v +0.9668v^3 \\ \partial_t v = -0.4386 \partial_{xx} u - 1.1376u v^2 - 0.8904u^3 \end{cases}$ \\ \addlinespace

60\% & 25\% & 
\begin{tabular}{@{}c@{}} $1.00$ \\ $1.00$ \end{tabular} & 
\begin{tabular}{@{}c@{}} $0.0979$ \\ $0.1516$ \end{tabular} & 
\begin{tabular}{@{}c@{}} $0.0876$ \\ $0.1409$ \end{tabular} & 
\begin{tabular}{@{}c@{}} $1.62\times10^{-2}$ \\ $1.74\times10^{-2}$ \end{tabular} & 

$\begin{cases} \partial_t u = 0.4699\partial_{xx}v+1.0979u^2v+0.9177v^3 \\ \partial_t v = -0.4550 \partial_{xx} u - 1.1516u v^2 - 0.8597u^3 \end{cases}$ \\ \addlinespace

80\% & 25\% & 
\begin{tabular}{@{}c@{}} $1.00$ \\ $1.00$ \end{tabular} & 
\begin{tabular}{@{}c@{}} $0.1288$ \\ $0.2336$ \end{tabular} & 
\begin{tabular}{@{}c@{}} $0.0815$ \\ $0.1725$ \end{tabular} & 
\begin{tabular}{@{}c@{}} $2.72\times10^{-2}$ \\ $2.74\times10^{-2}$ \end{tabular} & 
$\begin{cases} \partial_t u = 0.4356 \partial_{xx} v+0.9934u^2v + 0.8962v^3\\ \partial_t v = -0.4270 \partial_{xx} u - 0.7664u v^2 - 0.9162u^3 \end{cases}$ \\ \addlinespace

100\% & 25\% & 
\begin{tabular}{@{}c@{}} $1.00$ \\ $1.00$ \end{tabular} & 
\begin{tabular}{@{}c@{}} $0.2414$ \\ $0.3755$ \end{tabular} & 
\begin{tabular}{@{}c@{}} $0.2097$ \\ $0.1887$ \end{tabular} & 
\begin{tabular}{@{}c@{}} $4.66\times10^{-2}$ \\ $4.85\times10^{-2}$ \end{tabular} & 

$\begin{cases} \partial_t u = 0.3885 \partial_{xx} v+1.1679 u^2v + 0.7586v^3\\ \partial_t v = -0.3122 \partial_{xx} u - 0.9082u v^2 - 0.8091u^3 \end{cases}$ \\
\bottomrule
\end{tabular}
}
\end{table}

We evaluated Weak-PDE-Net's performance by randomly subsampling the full grid data, with sample ratio $r$ selected from $\{10\%, 25\%, 37.5\%, 50\%\}$. To further evaluate our method's robustness to measurement noise, we fixed the sampling ratio at $r=25\%$ and introduce Gaussian noise of varying intensities, where the noise ratio $\sigma_{\text{NR}}$ ranges from $20\%$ to $100\%$.

For the NLS equation, a complex-valued system, we incorporated the symmetry equivariance hypothesis (detailed in Section \ref{sec: Symmetry Equivariance Hypothesis}). Without this hypothesis, especially when the noise level exceeds $40\%$, and the sampling ratio is below $25\%$, the identification results become inconsistent. Specifically, one equation can be accurately recovered, while the other fails to converge. This problem can be effectively solved by introducing structural symmetry, thereby achieving robust synchronous convergence.

Table \ref{tab:results_nls} reports the experimental results for the nonlinear Schrödinger equation. As shown in the Table \ref{tab:results_nls}, our method identifies the correct equation structure in all cases, maintaining a TPR of 1.00. Weak-PDE-Net successfully identifies the coupled terms between the real and imaginary parts without retrieving any non-physical terms. Regarding data efficiency, Weak-PDE-Net identifies the governing equation in the setting with a sample ratio as low as $r=10\%$. In these cases, the parameter estimation remains precise, with the $E_\infty$ coefficient error kept around $10^{-2}$. For noise robustness, the identified coefficients deviate only slightly from the true values at noise levels between $20\%$ and $60\%$. Notably, our method recovers the correct governing laws even from data contaminated with $100\%$ noise. These results indicate that our method can recover the underlying dynamics of this coupled system even from highly sparse and noisy observations.

\subsubsection{Wave Equation}
\label{sec: Wave Equation}
The 2D Wave equation is a fundamental prototype of second-order partial differential equations and is crucial for describing wave propagation phenomena in multidimensional space. Different from 1D PDEs, this equation can capture the dynamics of systems involving multiple spatial variables, and it follows this form,
\begin{equation}
    \label{eq: wave}
    \partial_{tt}u=c^2(\partial_{xx}u+\partial_{yy}u).
\end{equation} 

We used the dataset from \citet{stephany2024pde}, focusing on the 2D wave equation with unit wave speed ($c^2=1$). The problem is defined on the spatio-temporal domain $(t,x,y) \in [0, 10]\times [-5, 5]\times [-5, 5]$ with a resolution of $100 \times 64 \times 64$ grid points. 

We performed random subsampling on the full grid data, varying the sampling ratio $r$ across the set $\{1.0\%, 2.5\%, 5.0\%,\\ 10\%, 25\%\}$. Furthermore, to assess robustness against measurement noise, we fixed the sampling ratio at $r=5\%$ and introduced varying levels of Gaussian noise, with noise ratios $\sigma_{\text{NR}}$ ranging from $20\%$ to $100\%$. 

\begin{table}[!b]
\centering
\caption{Performance evaluation on the 2D Wave equation.}
\label{tab:results_wave}
\resizebox{0.85\textwidth}{!}{
\begin{tabular}{cc c ccc l}
\toprule
$\sigma_{\text{NR}}$ & $r$ & TPR & ${E_{\infty}}$ & ${E_2}$ & ${\mathcal{L}(\hat{U},U)}$ & Discovered Equation \\
\midrule
\multicolumn{7}{c}{\textbf{Varying Sample Ratio}} \\
\midrule
0\% & 1.0\% & 1.00 & 0.0181 & 0.0135 & $7.12\times10^{-3}$ & $\partial_{tt}u=0.9943\partial_{xx}u+0.9819\partial_{yy}u$ \\
0\% & 2.5\% & 1.00 & 0.0158 & 0.0133 & $8.35\times10^{-4}$ & $\partial_{tt}u=0.9842\partial_{xx}u+0.9899\partial_{yy}u$ \\
0\% & 5.0\% & 1.00 & 0.0080 & 0.0056 & $9.19\times10^{-4}$ & $\partial_{tt}u=0.9999\partial_{xx}u+0.9920\partial_{yy}u$ \\
0\% & 10\%  & 1.00 & 0.0041 & 0.0034 & $7.16\times10^{-4}$ & $\partial_{tt}u=1.0024\partial_{xx}u+1.0041\partial_{yy}u$ \\
0\% & 25\%  & 1.00 & 0.0052 & 0.0050 & $7.48\times10^{-4}$ & $\partial_{tt}u=1.0052\partial_{xx}u+1.0048\partial_{yy}u$ \\
\midrule
\multicolumn{7}{c}{\textbf{Varying Noise Level ($r=5.0\%$)}} \\
\midrule
20\%  & 5.0\% & 1.00 & 0.0554 & 0.0521 & $6.97\times10^{-3}$ & $\partial_{tt}u=0.9446\partial_{xx}u+0.9515\partial_{yy}u$ \\
40\%  & 5.0\% & 1.00 & 0.1835 & 0.1426 & $6.90\times10^{-2}$ & $\partial_{tt}u=0.9163\partial_{xx}u+0.8165\partial_{yy}u$ \\
60\%  & 5.0\% & 1.00 & 0.3611 & 0.3230 & $1.65\times10^{-1}$ & $\partial_{tt}u=0.7202\partial_{xx}u+0.6390\partial_{yy}u$ \\
80\%  & 5.0\% & 1.00 & 0.3473 & 0.3220 & $2.11\times10^{-1}$ & $\partial_{tt}u=0.7055\partial_{xx}u+0.6527\partial_{yy}u$ \\
100\% & 5.0\% & 0.67 & 0.5450 & 0.4609 & $2.28\times10^{-1}$ & $\partial_{tt}u=-0.1528u+0.6767\partial_{xx}u+0.4550\partial_{yy}u$ \\
\bottomrule
\end{tabular}
}
\end{table}
The corresponding results and error metrics for these experiments are summarized in Table \ref{tab:results_wave}. Experimental results on the 2D wave equation further highlight the robustness of Weak-PDE-Net, especially when extended to higher-dimensional systems. Regarding robustness, our method exhibits strong robustness to severe data sparsity, providing accurate coefficient estimates even at the lowest sampling ratio $r = 1.0\%$. Furthermore, the experimental results on the 2D wave equation also underscore the stability of our method. A notable feature of these results is that our method maintains good performance even under stringent conditions of low sampling density $r = 5.0\%$ and high noise. Typically, recovering dynamics from such sparse and noisy data is challenging. However, our method successfully identifies the correct control term even with noise levels as high as $80\%$, failing only when the noise reaches $100\%$. 

\subsubsection{Sine--Gordon Equation}
\label{sec: Sine--Gordon Equation}
The Sine-Gordon (SG) equation is a classic model for nonlinear hyperbolic dynamics in relativistic field theory and condensed matter physics \citep{rubinstein1970sine}. We incorporate them into our benchmark to evaluate the ability of our algorithm to identify transcendental nonlinearities. Unlike typical polynomial models, the SG equation contain a transcendental term $\sin(u)$. We focus on the 2D form:
\begin{equation}
\label{Sine-Gordon Equation}
\partial_{tt}u = \partial_{xx}u + \partial_{yy}u - \sin(u).
\end{equation}

Following the setup described in \citep{messenger2021weak}, we consider a system defined on $(x,y) \in [-\pi, \pi] \times [-1, 1]$. The dataset is generated on a $403 \times 129$ spatial grid with a time step of $\Delta t = 0.025$. We used periodic boundary conditions in the $x$ direction and homogeneous Dirichlet boundary conditions in the $y$ direction. During discovery, the trajectories are truncated to the first 70 time steps, resulting in a dataset of $70 \times 403 \times 129$ dimensions.

Random subsampling was applied to the full grid data, with the sampling ratio $r$ taking values in $\{2.5\%, 5.0\%, 10\%,\\ 25\%, 50\%\}$. In addition, to evaluate robustness to measurement noise, the sampling ratio was fixed at $r=10\%$, while Gaussian noise of increasing intensity was added, with noise ratios $\sigma_{\text{NR}}$ varying from $20\%$ to $100\%$.

\begin{table}[!b]
\centering
\caption{Performance evaluation on the SG equation.}
\label{tab:results_sg}
\resizebox{0.85\textwidth}{!}{
\begin{tabular}{cc c ccc l}
\toprule
$\sigma_{\text{NR}}$ & $r$ & TPR & ${E_{\infty}}$ & ${E_2}$ & ${\mathcal{L}(\hat{U},U)}$ & Discovered Equation \\
\midrule
\multicolumn{7}{c}{\textbf{Varying Sample Ratio}} \\
\midrule
0\% & 2.5\% & 1.00 & 0.3004 & 0.1800 & $1.02\times10^{-2}$ & $\partial_{tt}u = 0.6996\partial_{xx}u+0.9995\partial_{yy}u-1.0839\sin(u)$ \\
0\% & 5.0\% & 1.00 & 0.2608 & 0.1804 & $9.68\times10^{-3}$ & $\partial_{tt}u = 0.8280\partial_{xx}u+0.9965\partial_{yy}u-0.7392\sin(u)$ \\
0\% & 10\%  & 1.00 & 0.1341 & 0.1052 & $3.08\times10^{-3}$ & $\partial_{tt}u = 0.8775\partial_{xx}u+0.9855\partial_{yy}u-1.1341\sin(u)$ \\
0\% & 25\%  & 1.00 & 0.1772 & 0.1026 & $3.21\times10^{-3}$ & $\partial_{tt}u = 1.0049\partial_{xx}u+0.9875\partial_{yy}u-0.8228\sin(u)$ \\
0\% & 50\%  & 1.00 & 0.1689 & 0.0976 & $3.95\times10^{-3}$ & $\partial_{tt}u = 1.0042\partial_{xx}u+1.0030\partial_{yy}u-1.1689\sin(u)$ \\
\midrule
\multicolumn{7}{c}{\textbf{Varying Noise Level ($r=10\%$)}} \\
\midrule
20\%  & 10\% & 1.00 & 0.1351 & 0.0822 & $3.13\times10^{-3}$ & $\partial_{tt}u = 1.0439\partial_{xx}u+0.9904\partial_{yy}u-0.8649\sin(u)$ \\
40\%  & 10\% & 1.00 & 0.1845 & 0.1296 & $3.01\times10^{-3}$ & $\partial_{tt}u = 1.1845\partial_{xx}u+0.9899\partial_{yy}u-0.8724\sin(u)$ \\
60\%  & 10\% & 1.00 & 0.2406 & 0.1441 & $3.45\times10^{-3}$ & $\partial_{tt}u = 0.7594\partial_{xx}u+0.9787\partial_{yy}u-1.0628\sin(u)$ \\
80\%  & 10\% & 1.00 & 0.5467 & 0.3297 & $5.89\times10^{-3}$ & $\partial_{tt}u = 1.1623\partial_{xx}u+0.9719\partial_{yy}u-0.4533\sin(u)$ \\
100\% & 10\% & 1.00 & 0.5706 & 0.3466 & $6.38\times10^{-3}$ & $\partial_{tt}u = 1.1862\partial_{xx}u+0.9876\partial_{yy}u-0.4294\sin(u)$ \\
\bottomrule
\end{tabular}
}
\end{table}
Table~\ref {tab:results_sg} reports the identification results and corresponding error metrics. The SG benchmark provide evidence of Weak-PDE-Net's capacity to resolve transcendental nonlinearities. Across the entire range of testing conditions, our algorithm achieved perfect structural identification ($\text{TPR} = 1.00$) and identified the $\sin(u)$ term within a hybrid feature space. This identification proved highly resistant to interference, as the correct governing equation was recovered even when the data was corrupted by $100\%$ noise or reduced to a mere $2.5\%$ sample ratio.

\subsubsection{Incompressible Navier-Stokes Equation}
\label{sec: Navier-Stokes Equation}
The Navier-Stokes (NS) equation, proposed in the 19th century by Claude-Louis Navier \citep{navier1823memoire} and George Gabriel Stokes \citep{stokes2007theories}, describe the motion of viscous fluids. Recently, this equation is the cornerstone of fluid mechanics, widely applied in science and engineering, encompassing everything from aerodynamic design and weather forecasting to ocean current modeling. As a fundamental model of fluid dynamics, the NS equation are extensively used to evaluate the accuracy of computational schemes \citep{raissi2019physics,li2020fourier} and, in recent years, to assess the robustness of partial differential equation discovery \citep{rudy2017data,messenger2021weak}.
frameworks.

We specifically consider the 2D vorticity transport equation:
\begin{equation}
\label{eq:NS}
\partial_t \omega = -\partial_x(u\omega) -\partial_y(v\omega) + \frac{1}{Re} \partial_{xx}\omega+ \frac{1}{Re} \partial_{yy}\omega,
\end{equation}
where $\omega$ denotes the vorticity, $(u,v)$ is the velocity field, and $Re$ represents the Reynolds number. For this test case, we analyze the classic problem of flow past a cylinder at $Re=100$. 

Following the experimental procedures in \citet{taira2007immersed}, we generate the datasets using the Immersed Boundary Projection Method (IBPM). The simulation is conducted on a domain of $(x,y) \in [-1, 8] \times [-2, 2]$ with a cylinder of diameter $D=1$ centered at the origin. We used a third-order Runge-Kutta scheme with a spatiotemporal resolution of $\Delta x = \Delta t = 0.02$. After the system reaches a steady state or limit cycle, we record a trajectory of 2000 time steps. To focus on the wake dynamics, the spatial domain is restricted to $(x,y) \in [1, 7.5] \times [-1.5, 1.5]$. We then perform a $10\%$ temporal subsampling with an effective time step of $\Delta t_{\text{eff}} = 0.2$. This process results in a final data set with $101 \times 300 \times 150$ grid points. 

We evaluated Weak-PDE-Net's performance by randomly subsampling the full grid data, with the sampling ratio $r$ selected from $\{5\%, 10\%, 100\%\}$. To further evaluate Weak-PDE-Net's robustness to measurement noise, we fixed the sampling ratio at $r=10\%$ and introduced Gaussian noise with varying intensities, with the noise ratio $\sigma_{\text{NR}}$ ranging from $10\%$ to $50\%$

Importantly, we imposed Galilean Invariance constraints to refine the candidate library $G$ obtained after the pruning stage. This process involved a selective strategy, absolute velocity terms (e.g., $u, v, u^2$) were excluded due to their violation of frame invariance, whereas velocity-gradient couplings (e.g., $\partial_x(u \omega)$) were retained to preserve the advective structure. These physics-informed constraints effectively minimized spurious correlations, significantly enhancing the robustness of the identification.

\begin{table}[H]
\centering
\caption{Performance evaluation on the incompressible NS equation at $Re=100$. The symbol $--$ indicates that forward learning is not required under full sampling conditions; therefore, the reconstruction MSE is not applicable.}
\label{tab:results_ns}
\resizebox{\textwidth}{!}{
\begin{tabular}{c c c c c c c l}
\toprule
$Re$ & $\sigma_{\text{NR}}$ & $r$ & TPR & ${E_{\infty}}$ & ${E_2}$ & ${\mathcal{L}(\hat{U},U)}$ & Discovered Equation \\
\midrule
\multirow{9}{*}{100} 
& \multicolumn{7}{c}{\textbf{Varying Sample Ratio} ($\sigma_{\text{NR}}=0\%$)} \\
\cmidrule{2-8}
& 0\% & 5\%  & 1.00 & 0.0374 & 0.0141 & $1.35\times10^{-3}$ & $\partial_t \omega = -0.9824\partial_x(u\omega) -0.9904\partial_y(v\omega) + 0.0097\partial_{xx}\omega + 0.0104\partial_{yy}\omega$ \\
& 0\% & 10\%   & 1.00 & 0.0170 & 0.0052 & $1.23\times10^{-3}$ & $\partial_t \omega = -0.9927\partial_x(u\omega) -1.0004\partial_y(v\omega) + 0.0099\partial_{xx}\omega + 0.0102\partial_{yy}\omega$ \\
& 0\% & 100\% & 1.00 & 0.0051 & 0.0043 & $--$ & $\partial_t \omega = -1.0010\partial_x(u\omega) -0.9949\partial_y(v\omega) + 0.0100\partial_{xx}\omega + 0.0100\partial_{yy}\omega$ \\
\cmidrule{2-8}
& \multicolumn{7}{c}{\textbf{Varying Noise Level} ($r=10\%$)} \\
\cmidrule{2-8}
& 10\% & 10\% & 1.00 & 0.0972 & 0.0204 & $9.51\times10^{-4}$ & $\partial_t \omega = -0.9792\partial_x(u\omega) -0.9800\partial_y(v\omega) + 0.0110\partial_{xx}\omega + 0.0098\partial_{yy}\omega$ \\
& 20\% & 10\% & 1.00 & 0.0326 & 0.0306 & $2.20\times10^{-3}$ & $\partial_t \omega = -0.9674\partial_x(u\omega) -0.9716\partial_y(v\omega) + 0.0099\partial_{xx}\omega + 0.0103\partial_{yy}\omega$ \\
& 30\% & 10\% & 1.00 & 0.0872 & 0.0364 & $2.43\times10^{-3}$ & $\partial_t \omega =  -0.9666\partial_x(u\omega) -0.9609\partial_y(v\omega) + 0.0092\partial_{xx}\omega + 0.0109\partial_{yy}\omega$ \\
& 40\% & 10\% & 1.00 & 0.1757 & 0.0942 & $1.17\times10^{-2}$ & $\partial_t \omega = -0.9093\partial_x(u\omega) -0.9025\partial_y(v\omega) + 0.0118\partial_{xx}\omega + 0.0083\partial_{yy}\omega$ \\
& 50\% & 10\% & 1.00 & 0.2851 & 0.1014 & $9.61\times10^{-3}$ & $\partial_t \omega = -0.9216\partial_x(u\omega) -0.8800\partial_y(v\omega) + 0.0071\partial_{xx}\omega + 0.0099\partial_{yy}\omega$ \\
\bottomrule
\end{tabular}
}
\end{table}

Detailed identification results and corresponding metrics for these experiments are summarized in Table \ref{tab:results_ns}. Our method consistently identifies the correct symbol structure for non-stationary states across all test sample ratio and noise levels (TPR=1.00). In noise-free conditions, Weak-PDE-Net exhibits high data efficiency, achieving accurate parameter estimation even when the dataset is downsampled to $5\%$. Furthermore, our method demonstrates excellent robustness to measurement noise. Even under interference ($\sigma_{\text{NR}}=50\%$), it avoids false alarms and successfully recovers the governing equation with a diffusion coefficient of approximately $0.0071$.

\subsection{Ablation Studies}
\label{sec: Ablation Studies}
We performed the ablation studies to to evaluate the roles of the adaptive Gaussian kernel in WeakPDE-Net and the neural architecture search strategy within the training process. Here, we do not conduct specific ablation experiments on the introduce of physical information, as this aspect has already been thoroughly examined in Section \ref{sec: Robustness Analysis} using the NLS and NS equations.

\begin{figure}[!h]
	\centering
	\includegraphics[width=\textwidth]{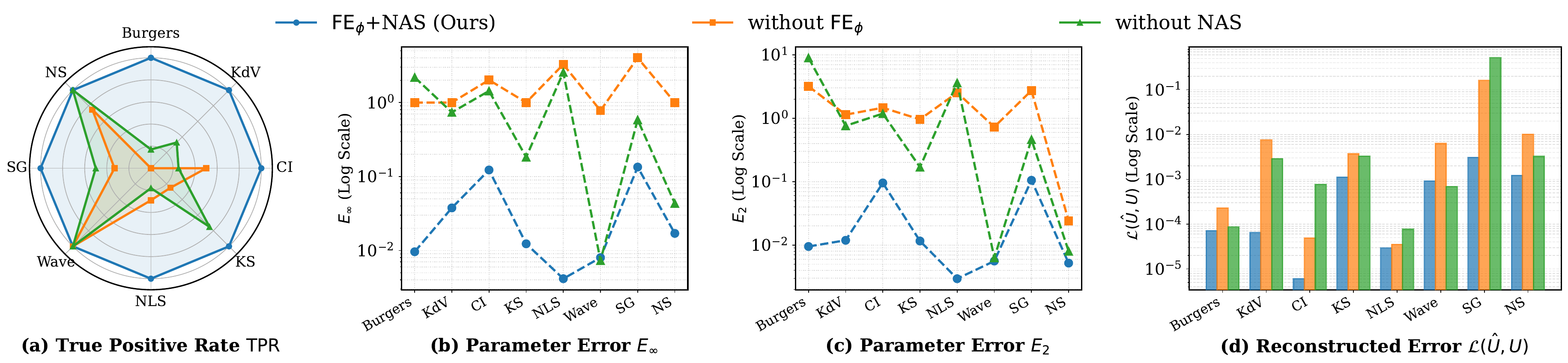}
	\caption{The results of the ablation studies. (a) A radar chart illustrating the True Positive Rate $\text{TPR}$. (b) and (c) Line plots showcasing the parameter errors $E_{\infty}$ and $E_{2}$, respectively. (d) A bar chart representing the reconstructed error $\mathcal{L}(\hat{U},U)$.}
    \label{Figure_4}
\end{figure}

We conducted experiments with two settings across all eight equations under specific sampling ratios (detailed in Table \ref{tab: ablation_results} in Appendix \ref{Appendix: Detialed Results}):
\begin{itemize}
\item {Variant without $\mathrm{FE}_{\phi}$: } This variant disables the adaptive Gaussian kernel while keeping all other parameters identical to the full model.
\item {Variant without NAS: } This variant replaces the symbolic architecture search with the complete network architecture as described in Section \ref{sec:Differentiable neural architecture search} (with $L=2$, $N_i=2$, and $\mathcal{O} = \{id, \sin, (\cdot)^2, (\cdot)^3, \times\}$). All other training settings remain identical to the full model.
\end{itemize}

We evaluate the full model against its ablated variants across four metrics. As shown in the radar chart in Fig. \ref{Figure_4}(a), the full model achieves a TPR of 1.00 across all eight benchmark equations. In contrast, the ablated variants show decreased performance, particularly on the Burgers, KdV, and NLS equation. Regarding parameter precision, Fig. \ref{Figure_4}(b) and (c) indicate that the full model maintains lower estimation errors compared to the ablated versions, with a reduction of approximately 1 to 2 orders of magnitude. Furthermore, the bar chart in Fig. \ref{Figure_4}(d) shows that the full model achieves lower reconstructed MSE in most cases. These results suggest that the integration of these components within the Weak-PDE-Net facilitates the identification of governing laws and the approximation of the system response function. The effectiveness of the full model stems from two complementary.
\begin{itemize}
    \item \textbf{The Role of  $\mathrm{FE}_{\phi}$: }The adaptive Gaussian kernel facilitates the high-quality reconstruction of $U$. Its absence results in increased MSE, suggesting that a standalone MLP may lack the capacity to capture complex spatiotemporal dynamics. 
    The adaptive kernel identifies sharp discontinuities, such as shock waves in the Burgers and NS equation, as evidenced by the visualization results in Appendix \ref{Appendix: Detialed Results} (Tables \ref{tab: results_visual_1d}, \ref{tab:results_visual_2d}, and \ref{tab:results_visual_ns}).

    \item \textbf{The Role of NAS: } Without NAS, the fixed symbolic architecture fails to prune redundant operators, yielding equations with excessive complexity and spurious terms (see Table \ref{tab: ablation_results} in Appendix \ref{Appendix: Detialed Results}). Furthermore, these incorrect physical equation constraints affect the loss function, reducing the accuracy of forward learning.
\end{itemize}

Overall, removing either component affects both equation discovery and solution learning, resulting in a decline in overall performance. Additionally, detailed results are provided in Table \ref{tab: ablation_results} in Appendix \ref{Appendix: Detialed Results}.

\subsection{Comparison Analysis}
\label{sec: Comparison Analysis}
We evaluate our method in comparison with Weak-PDE-LEARN on three PDEs of increasing complexity, namely the second-order Burgers, third-order KdV, and fourth-order KS equations. To ensure a fair comparison, the datasets in this section are generated using the same settings as Weak-PDE-LEARN \citep{stephany2024weak}. These configurations differ from the ones used in our earlier robustness tests. In this evaluation, we set the number of data points to $N_{\text{Data}} = 4,000$ for the Burgers and KdV equation, and $N_{\text{Data}} = 10,000$ for the KS equation. We  test the identification performance across four noise levels: $25\%, 50\%, 75\%,$ and $100\%$.

\begin{table}[!b]
\centering
\caption{Comparative evaluation against Weak-PDE-LEARN (WPL). The symbol \textendash \ for $E_\infty$ and $E_2$ indicates that the equation was not correctly identified ($\text{TPR} < 1$).}
\label{tab: results_comparison}
\resizebox{\textwidth}{!}{
\small 
\setlength{\tabcolsep}{4pt} 
\renewcommand{\arraystretch}{1.25} 
\begin{tabular}{c cc cc cc cc cc cc}
\toprule
\multirow{3}{*}{{Noise}} & \multicolumn{4}{c}{{Burgers}} & \multicolumn{4}{c}{{KdV}} & \multicolumn{4}{c}{{KS}} \\
\cmidrule(lr){2-5} \cmidrule(lr){6-9} \cmidrule(lr){10-13}
 & \multicolumn{2}{c}{$E_\infty$} & \multicolumn{2}{c}{$E_2$} & \multicolumn{2}{c}{$E_\infty$} & \multicolumn{2}{c}{$E_2$} & \multicolumn{2}{c}{$E_\infty$} & \multicolumn{2}{c}{$E_2$} \\
\cmidrule(lr){2-3} \cmidrule(lr){4-5} \cmidrule(lr){6-7} \cmidrule(lr){8-9} \cmidrule(lr){10-11} \cmidrule(lr){12-13}
 & WPL & {Ours} & WPL & {Ours} & WPL & {Ours} & WPL & {Ours} & WPL & {Ours} & WPL & {Ours} \\
\midrule
25\%  & 0.0130 & \textbf{0.0098} & 0.0130 & \textbf{0.0097} & 0.0070 &\textbf{0.0040} & 0.0036 &\textbf{0.0036} & {0.0100} & \textbf{0.0066} & {0.0094} & \textbf{0.0049} \\
50\%  & 0.0808 & \textbf{0.0593} & 0.0801 & \textbf{0.0117} & 0.0262 &\textbf{0.0136} & 0.0259 &\textbf{0.0062} & \textbf{0.0383} & 0.0487 &0.0400 & \textbf{0.0342} \\
75\%  & \textbf{0.1530} & {0.1746} & 0.0982 & \textbf{0.0343} & 0.0520 &\textbf{0.0393} & 0.0494 &\textbf{0.0352} & 0.1153 & \textbf{0.0895} &0.1300 & \textbf{0.0331}  \\
100\% & 0.3400 & \textbf{0.1931} & 0.1334 & \textbf{0.0552} & 0.1838 &\textbf{0.0732} & 0.1752 &\textbf{0.0719} & \textendash & \textbf{0.1340} & \textendash & \textbf{0.0973} \\
\bottomrule
\end{tabular}
}
\end{table}

Table \ref{tab: results_comparison} summarizes the comparison results between our method Weak-PDE-Net and Weak-PDE-LEARN. As shown in the Table \ref{tab: results_comparison}, our method maintains higher identification accuracy and robustness as the noise level increases. For the Burgers and KdV equation, our method achieves lower parameter errors compared to Weak-PDE-LEARN, especially under high noise conditions. Specifically, for the KdV equation, our method outperforms the other method across the entire noise level range from $25\%$ to $100\%$, with better performance in terms of both $E_\infty$ and $E_2$ errors. The performance difference is most significant for the fourth-order KS equation. While Weak-PDE-LEARN correctly identifies the KS equation at lower noise levels, it fails to recover the equation's structural form at $100\%$ noise. In contrast, our method successfully identifies the eqution structure even at $100\%$ noise, maintaining low parameter errors ($E_\infty = 0.1340$, $E_2 = 0.0973$). Additionally, the resulting equations are summarized in Table \ref{tab:results_comparison_ours} of Appendix \ref{Appendix: Detialed Results}.

\section{Discussion}
\label{sec: Discussion}

\subsection{Adaptive Gaussian Kernel and Spectral Bias}
\label{sec: Adaptive Gaussian kernel and spectral bias}
The response function learner, which embeds adaptive Gaussian kernels into MLPs, has been demonstrated to satisfy the universal approximation property and effectively mitigate the spectral bias inherent in standard MLPs \citep{kangpig}. Our ablation studies show that this architecture significantly improves the reconstruction performance.

The reconstruction performance improvement can be attributed to a primary reason. The adaptive nature of the Gaussian kernels allows the network to automatically adjust its representation during training. Specifically, this self-adjustment drives the kernels to cluster in regions with sharp gradients or singularities, thereby facilitating the resolution of high-frequency dynamics. 

\subsection{Symbolic Networks and Open-form PDEs}
\label{sec: Symbolic Networks and Open-form PDEs}
Recent works \citep{liu2023snr,wu2023discovering} have demonstrated the high efficiency of using symbolic networks to represent various functions in symbolic regression. We apply symbolic networks to the PDE discovery task. Different from conventional symbolic networks restricted to fixed architectures \citep{long2019pde}, we introduce a symbolic network architecture search strategy to achieve open-form PDE discovery, which has yielded strong experimental results.

The efficacy of this strategy can be attributed to from its adaptability. The strategy transforms discrete architecture selection into a differentiable optimization problem. This allows the symbolic network to dynamically reshape its topology via gradient descent, navigating a vast functional space without being restricted to a fixed library. 

\subsection{Weak Form and Noise}
\label{sec: Weak form and noise}
As demonstrated in Section \ref{sec: Robustness Analysis}, our method remarkably maintains its identification accuracy even when subjected to extreme noise levels of up to $80\%$. This robustness is attributed to the integral nature of the weak formulation, which does not require the partial derivatives of the system response function.

By multiplying compactly supported test functions and integration by parts, the method transfers derivative operators from the noisy observations onto the smooth test functions. Since these test functions are known and differentiable, their derivatives can be computed exactly. Our method avoids amplifying the noise associated with numerical differentiation. Furthermore, since the Riemann integral is a scaled average over the integration domain, it effectively filters out zero-mean noise.

\subsection{Limitations}
\label{sec: limitations}
While the Weak-PDE-Net successfully enables the discovery of open-form PDEs, a limitation remains its inability to represent nested differential operators. To illustrate this problem, we consider the Porous Medium Equation:
$$\frac{\partial u}{\partial t} = \frac{\partial}{\partial x} \left( u \frac{\partial u}{\partial x} \right).$$
Within our standard weak-form representation, applying integration by parts shifts the spatial derivatives from the flux terms to the test function $\Phi$:
$$\int_{\Omega} \Phi \frac{\partial u}{\partial t} \, d\Omega = - \int_{\Omega} \frac{\partial \Phi}{\partial x} \left(u \frac{\partial u}{\partial x}\right)d\Omega.$$
The term $u \partial_x u$ cannot be expressed as a derivative-free function $F(U)$ of $U$. To express such terms, one potential avenue in future works is to extend the network architecture (Fig. \ref{Figure_1}) by incorporating a differential operator layer $\{D^{\alpha^{(p)}}\}_{p=1}^{S}$, and a symbolic network following the output $F(\hat{U})$. This extension expands the function space to a generalized composition form $G(D^{\alpha^{(1)}}F_1, \dots, D^{\alpha^{(S)}}F_S)$, allowing the network to represent nested differential terms. For systems with more complex structures, this architecture can be extended by stacking additional differential and symbolic layers to represent operators of greater depth.

While extending the network architecture improves the model's representational capacity, it introduces two primary challenges. First, the introduce of the differential operator layer affects the algorithm's noise robustness. Nevertheless, this is partially mitigated since the extended weak form still requires lower-order derivatives than its strong-form. Second, accommodating these additional differential and symbolic layers significantly increases the architectural complexity, which inevitably drives up the computational cost. Consequently, our future research will focus on optimizing these nested structures incrementally, possibly through block-wise training strategies to ensure computational efficiency.

Additionally, the weak-form integral acts as a low-pass filter that averages the system response over the integration domain \citep{stephany2024weak}. While this mechanism suppresses high-frequency noise, it also attenuates small-scale dynamics. Consequently, for equations characterized by localized features, such as the phase interfaces in the CI equation, the physical signal may be difficult to distinguish from noise. This smoothing effect can limit the identification accuracy in settings where the governing dynamics depend on fine length-scale variations.

\section{Conclusion and Future Works}
\label{sec: Conclusion}

In this paper, we introduce Weak-PDE-Net, an end-to-end differentiable framework for discovering open-form PDEs from sparse and noisy data. The framework incorporates a learnable Gaussian embedding to adaptively capture multi-scale features while addressing the spectral bias inherent in standard MLPs. By using a Differentiable Neural Architecture Search strategy, Weak-PDE-Net automates the identification of governing equations without requiring a pre-defined candidate library. To ensure physical consistency, particularly in multivariable systems, we integrated physics-informed refinements including Galilean Invariance constraints and the symmetry equivariance hypothesis. Experimental results across various benchmarks demonstrate that Weak-PDE-Net can recover PDEs from sparse and noisy observations.

Future research will focus on expanding this framework to model a broader class of complex physical systems. To support the representation of nested differential operators, we aim to enhance the expressive power of Weak-PDE-Net through efficient network architectures and targeted optimization strategies, including block-wise training. Furthermore, we plan to investigate the systematic integration of additional physical priors, thereby elevating the model's capability and ensuring physical consistency across diverse scientific applications.

\section*{CRediT authorship contribution statement}
\textbf{Xinxin Li:} Conceptualization, Formal analysis, Investigation, Methodology, Software, Visualization, Writing-original draft. \textbf{Xingyu Cui:} Formal analysis, Investigation, Methodology, Writing–-review \& editing \textbf{Jin Qi:} Formal analysis, Methodology, Validation. \textbf{Juan Zhang}: Investigation, Resources, Supervision, Validation. \textbf{Da Li:} Methodology, Software, Writing–-review \& editing. \textbf{Junping Yin:} Supervision, Project administration, Funding acquisition.

\section*{Data availability}
Data and code will be made available on request.

\section*{Acknowledgements}
This work was supported by the Major Program of the National Natural Science Foundation of China Nos. 12292980, 12292984 and No. 12201024.

\appendix
\section*{Appendix}
\section{Detailed Algorithm}
\begin{algorithm}[h]
\caption{Weak-PDE-Net Training}
\label{alg: Weak-PDE-Net training}
    \begin{algorithmic}[1]
        \STATE {\bfseries Input:} Weak-PDE-Net $\mathcal{N}_{\Theta}=(\mathrm{FE}_\phi,\ \mathrm{\mathrm{MLP}_\theta,\Psi_{(W,b,\alpha,\beta)},\mathcal{I}_{\xi}})$, training data $\{\tilde{U}(t_i,X_i)\}_{i=1}^{N_\text{data}}$, searching learning rate $\gamma_s$,  pruning learning rate $\gamma_p$, 
        tuning learning rate $\gamma_t$, searching epochs  $K_s$, pruning epochs $K_p$, tuning epochs $K_t$, Weak loss coefficient $\lambda_w$, Regularization loss coefficient $\lambda_r$
        \STATE \textbf{Phase 1: Differentiable symbolic networks architecture search}
        \FOR{$k = 1$ {\bfseries to} $K_s$}
            \STATE $\hat{U}\leftarrow\mathrm{MLP}_\theta(\mathrm{FE}_{\phi}(t,X))$, $\mathrm{RHS}\leftarrow\mathcal{I}_\xi(\Psi_{(W,b,\alpha,\beta)}(\hat{U}))$
            \quad$\triangleright$\COMMENT{Weak-PDE-Net forward propagation.}
            \STATE $\mathcal{L}_{\text{train}} \leftarrow \mathcal{L}_\text{data}(\hat{U})+\lambda_w\mathcal{L}_\text{weak}(\mathrm{RHS}) $
            \quad$\triangleright$\COMMENT{Compute the value of the loss function.}
            \STATE Update $\Theta$ using $\Theta \leftarrow \Theta - \gamma_s \nabla_{\Theta}\mathcal{L}_{\text{train}}$
            \quad$\triangleright$\COMMENT{Update the parameters in the Weak-PDE-Net.}
        \ENDFOR
    \STATE Compute $\mathcal{A}=\{L,\ \{\{op_k^{(l)}\}_{k=1}^{d_l}\}_{l=1}^{L}\}$ from $(\beta, \alpha)$
    \quad$\triangleright$\COMMENT{Select the symbolic network architecture.}
    \STATE Update $\Psi_{(W,b)}\leftarrow \Psi_{(W,b,\alpha,\beta)}$
    \quad$\triangleright$\COMMENT{Updating the forward propagation of the symbol network.}
    \STATE \textbf{Phase 2: Structural pruning for equation simplification}
    \FOR{$k = 1$ {\bfseries to} $K_p$}
            \STATE $\hat{U}\leftarrow\mathrm{MLP}_\theta(\mathrm{FE}_{\phi}(t,X))$, $\mathrm{RHS}\leftarrow\mathcal{I}_\xi(\Psi_{(W,b)}(\hat{U}))$
            \quad$\triangleright$\COMMENT{Weak-PDE-Net forward propagation.}
            \STATE $\mathcal{L}_{\text{train}} \leftarrow \mathcal{L}_\text{data}(\hat{U})+\lambda_w\mathcal{L}_\text{weak}(\mathrm{RHS})+\lambda_r\mathcal{L_\text{reg}}(\xi,W,b) $
            \quad$\triangleright$\COMMENT{Compute the value of the loss function.}
            \STATE Update $\Theta$ using $\Theta \leftarrow \Theta - \gamma _p\nabla_{\Theta}\mathcal{L}_{\text{train}}$
            \quad$\triangleright$\COMMENT{Update the parameters in the Weak-PDE-Net.}
        \ENDFOR
    \STATE Build the set $G$ from $(W,b)$ \quad$\triangleright$\COMMENT{Based on the Eq. \eqref{eq: EQL forward}.}
    \STATE \textbf{Phase 3: Sparse regression for coefficient tuning}
    \FOR{$k = 1$ {\bfseries to} $K_t$}
            \IF{$h \ge 2$}
                \STATE $\mathcal{I} \leftarrow \operatorname{SelectInvariance}(\hat{\xi})$
                \quad$\triangleright$\COMMENT{Determine dominant symmetry (Galilean vs. Phase).}
                \STATE $G \leftarrow \operatorname{Refine}(G, \mathcal{I})$
                \quad$\triangleright$\COMMENT{Filter spurious terms and supplement missing ones.}
            \ENDIF
            \STATE $\hat{U}\leftarrow\mathrm{MLP}_\theta(\mathrm{FE}_{\phi}(t,X))$, 
            \quad$\triangleright$\COMMENT{PDE Learning process in the Weak-PDE-Net.}
            \STATE {Compute} $G=\{G_{k,s}\}_{1\le k\le K,1\le s \le S}$ and $b=\{b_{k}\}_{1\le k\le K}$ \quad$\triangleright$\COMMENT{Based on the Eq. \eqref{eq: G_k,b_k}.}
            \STATE {Least-squares Regression} $\xi\leftarrow\min_{\xi} |G\xi - b|_2^2$, $\mathcal{L}_\text{weak}=|G\xi - b|_2^2$
            \quad$\triangleright$\COMMENT{Compute Coefficients.}
            \STATE $\mathcal{L}_{\text{train}} \leftarrow \mathcal{L}_\text{data}(\hat{U})+\lambda_w\mathcal{L}_\text{weak}$
            \COMMENT{Compute the value of the loss function.}
            \STATE {Update} $\Theta'=(\phi,\theta)$ using $\Theta' \leftarrow \Theta' - \gamma_t \nabla_{\Theta'}\mathcal{L}_{\text{train}}$
            \quad$\triangleright$\COMMENT{Update the parameters in the PDE Learning process.}
    \ENDFOR
    \STATE {Infer} PDE from $(F,\xi)$ 
    \STATE {\bfseries Output:} PDE, $\mathrm{MLP}_{\theta}$, $\mathrm{FE}_{\phi}$
    \end{algorithmic}
\end{algorithm}
We employ a three-stage training strategy consisting of Searching, Pruning, and Tuning phases. The detailed algorithmic procedure is presented in Algorithm \ref{alg: Weak-PDE-Net training}.

\section{Detailed Derivation} 
\label{appendix: Detailed Derivation}
Consider a complex-valued field $\psi(t,X) = u(t, X) + iv(t,X)$ governed by a equation $\partial_t \psi = \mathcal{N}(\psi)$, where the operator $\mathcal{N}$ can be decomposed into its real and imaginary components as $\mathcal{N}(\psi) = \mathcal{N}_{\text{re}}(u, v) + i\mathcal{N}_{\text{im}}(u, v)$. The system is said to exhibit $U(1)$ equivariance if the dynamical operator commutes with a global phase rotation. Mathematically, for any constant phase $\phi \in [0, 2\pi)$, the equivariance condition is expressed as:

\begin{equation}
\label{eq: appendix_commute}
\mathcal{N}(e^{i\phi}\psi) = e^{i\phi}\mathcal{N}(\psi),
\end{equation}
where $e^{i\phi}$ represents an element of the $U(1)$ unitary group acting as a rotation in the complex plane. To derive the specific structural coupling used in our Symmetry Equivariance Hypothesis, we evaluate this identity using a discrete rotation of $\phi = \pi/2$, which corresponds to multiplication by the imaginary unit $i$.

On the left-hand side of Eq. \eqref{eq: appendix_commute}, a phase shift of $\pi/2$ transforms the input field into $\psi' = i(u + iv) = -v + iu$. Substituting this transformed state into the operator components yields:

\begin{equation*}
\mathcal{N}(i\psi) = \mathcal{N}_{\text{re}}(-v, u) + i\mathcal{N}_{\text{im}}(-v, u).
\end{equation*}

On the right-hand side, rotating the output of the original operator by the same phase results in:
\begin{equation*}
i\mathcal{N}(\psi) = i[\mathcal{N}_{\text{re}}(u, v) + i\mathcal{N}_{\text{im}}(u, v)] = -\mathcal{N}_{\text{im}}(u, v) + i\mathcal{N}_{\text{re}}(u, v).
\end{equation*}

By equating the real and imaginary parts of both sides to satisfy the equivariance requirement, we establish the following structural identities:

\begin{align*}
\mathcal{N}_{\text{re}}(-v, u) &= -\mathcal{N}_{\text{im}}(u, v), \\
\mathcal{N}_{\text{im}}(-v, u) &= \mathcal{N}_{\text{re}}(u, v).
\end{align*}

The first identity directly defines the anti-symmetric relationship:

\begin{equation}
\mathcal{N}_{\text{im}}(u, v) = -\mathcal{N}_{\text{re}}(-v, u),
\end{equation}
This derivation proves that for any physical system obeying $U(1)$ symmetry, the evolution of the imaginary component is a rotationally conjugate reflection of the real component. Our Symmetric Library Augmentation strategy leverages this result as a generative rule, populating the candidate library with terms that respect this physical structure while allowing the data-driven discovery process to adaptively validate the symmetry.

\section{Computing Infrastructure and Hyperparameter Settings}
\label{Appendix: Computing Infrastructure and Hyperparameter Settings}

The experiments in this work were executed on an Intel(R) Xeon(R) Platinum 8468H CPU @ 2.10GHz, 1.5T RAM equipped with two NVIDIA H800 GPUs 80GB. 

The detailed hyperparameter settings used in our experiments are summarized in Table \ref{tab:hyperparameters}.

\begin{table}[!b]
    \centering
    \caption{Hyperparameter settings used in the experiments.}
    \label{tab:hyperparameters}
    \begin{tabular*}{0.75\textwidth}{@{\extracolsep{\fill}}lcc}
        \toprule
        \textbf{Parameter Name} & \textbf{Symbol} & \textbf{Value} \\
        \midrule
        Test Function Type & $\Phi$ & Poly-bump (KdV) / Exp-bump (Others) \\
        Support Radius Range & $(\eta_{\text{min}}, \eta_\text{max})$ & $(0.2, 0.4)$ \\
        Number of Gaussian Kernels & $N$ & $1000$ \\
        Gaussian Feature Dimension & $d'$ & $4$ \\
        Lightweight MLP Depth & $L$ & $2$ \\
        MLP Activation Function & $\sigma$ & $\sin$ (NS) / $\tanh$ (Others) \\
        MLP Hidden Dimension & $-$ & 32 \\
        Set of Operators & $\mathcal{O}$ & $\{id, (\cdot)^2, (\cdot)^3, \sin, \times\}$ \\
        Max Symbolic Network Length & $L_\text{max}$ & $3$ \\
        Set of Operator Counts & $\mathcal{K}$ & $\{0,1,2,3\}$ \\
        \bottomrule
    \end{tabular*}
\end{table}

\section{Detialed Results}
\label{Appendix: Detialed Results}

Table \ref{tab: ablation_results} details the specific results of the ablation studies. Table \ref{tab:results_comparison_ours} presents the specific performance of our proposed method in comparative experiments. Tables \ref{tab: results_visual_1d}, \ref{tab:results_visual_2d}, and \ref{tab:results_visual_ns} demonstrate the effectiveness of the forward learning of system response functions.

\begin{table}[H]
\caption{Reconstruction performance of the learned response function 
$U$ for 1D equations.}
\label{tab: results_visual_1d}
\resizebox{0.75\textwidth}{!}{
\begin{tabular}{ccc}
\toprule
{Equation} & {Settings} & {Visualization of the true and learned system dynamics} \\
\midrule
\makecell{Burgers} & \makecell{$\sigma_\text{NR}=0, r=25\%$} & 
\includegraphics[width=0.3\textwidth, valign=m]{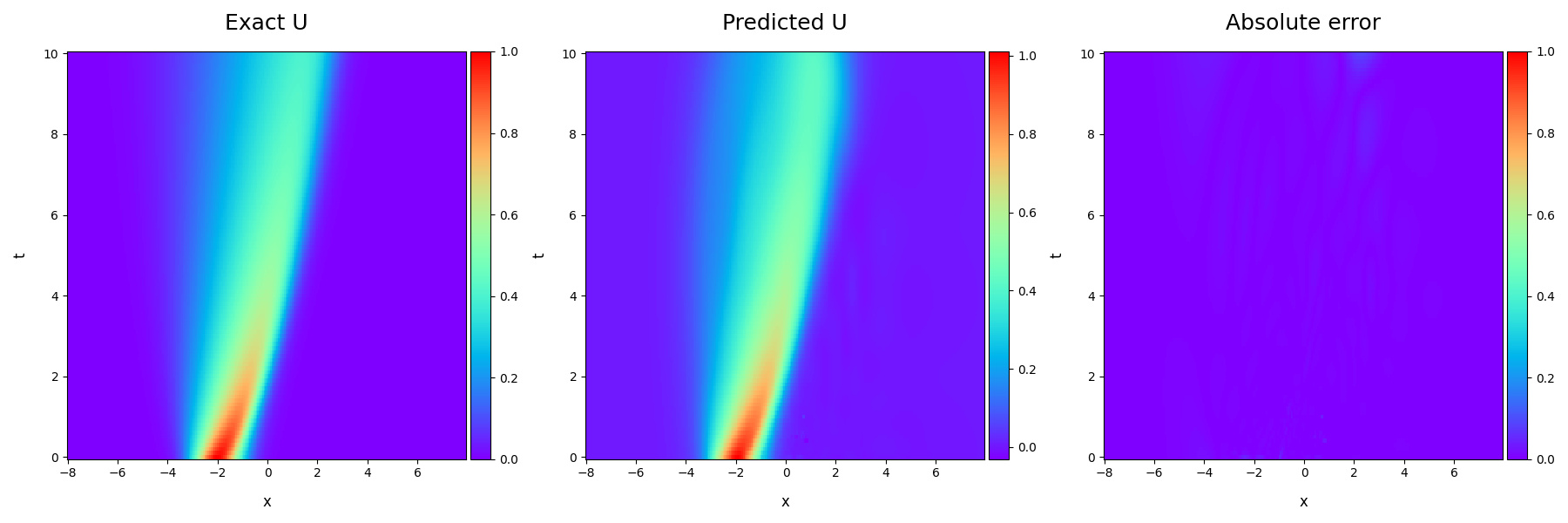} \\
\midrule
\makecell{KdV} & \makecell{$\sigma_\text{NR}=0, r=25\%$} & 
\includegraphics[width=0.3\textwidth, valign=m]{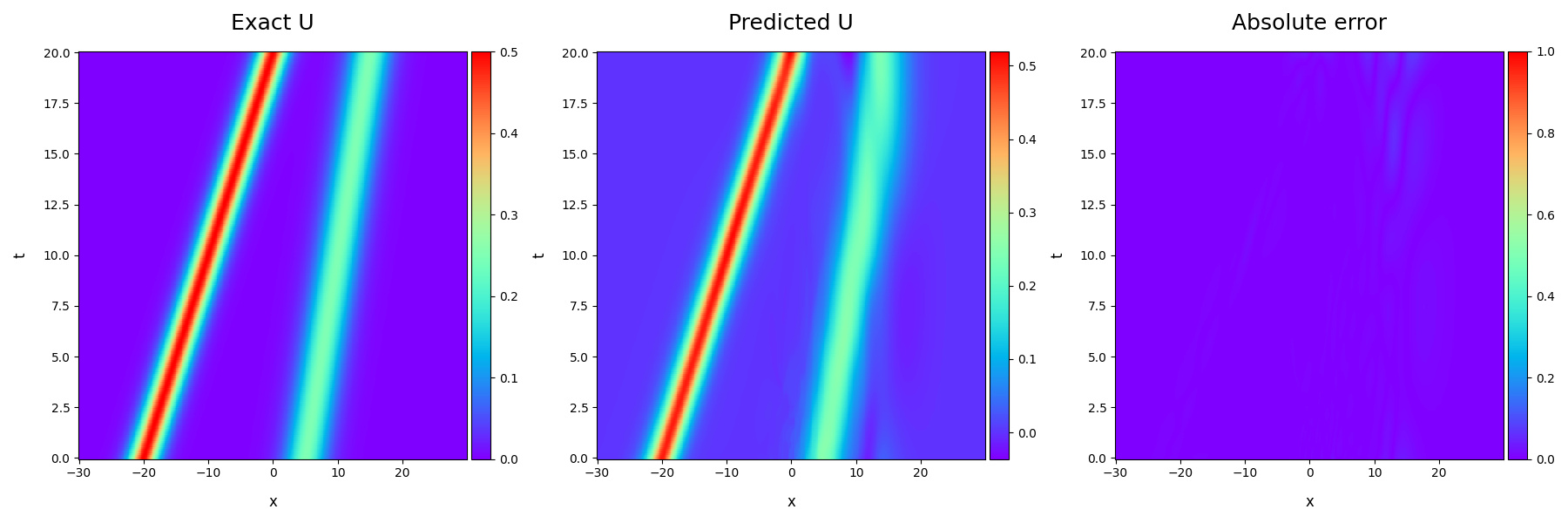} \\
\midrule

\makecell{KS} & \makecell{$\sigma_\text{NR}=0, r=25\%$} & 
\includegraphics[width=0.3\textwidth, valign=m]{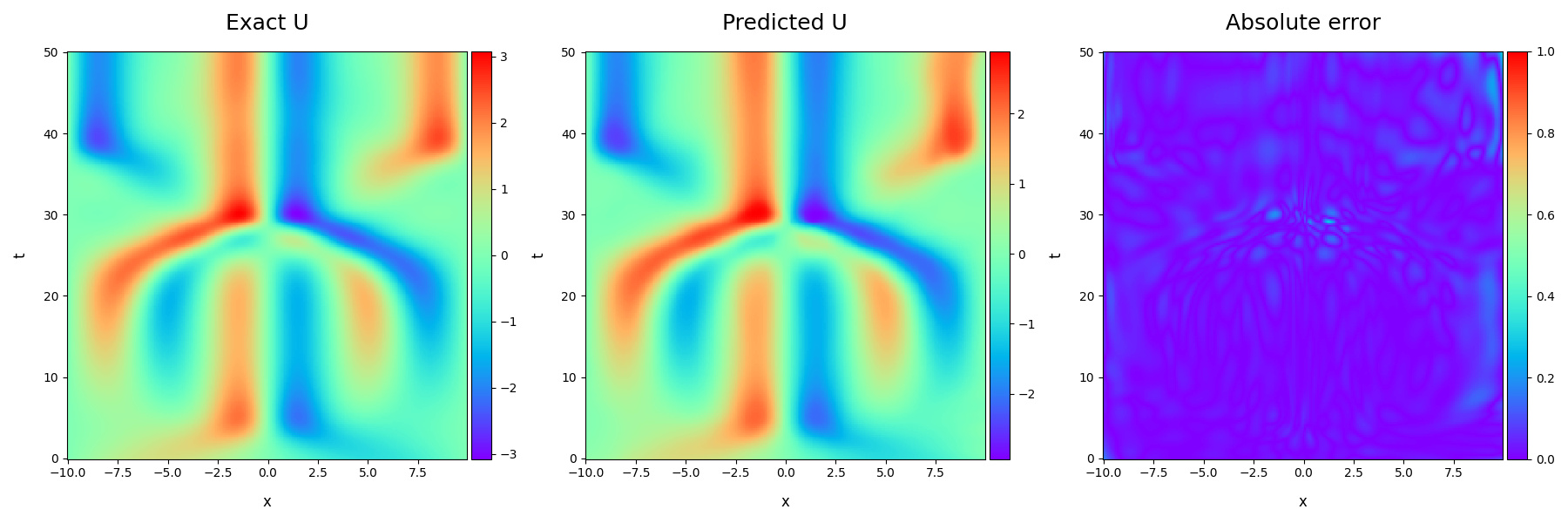} \\
\midrule
\makecell{CI} & \makecell{$\sigma_\text{NR}=0, r=25\%$} & 
\includegraphics[width=0.3\textwidth, valign=m]{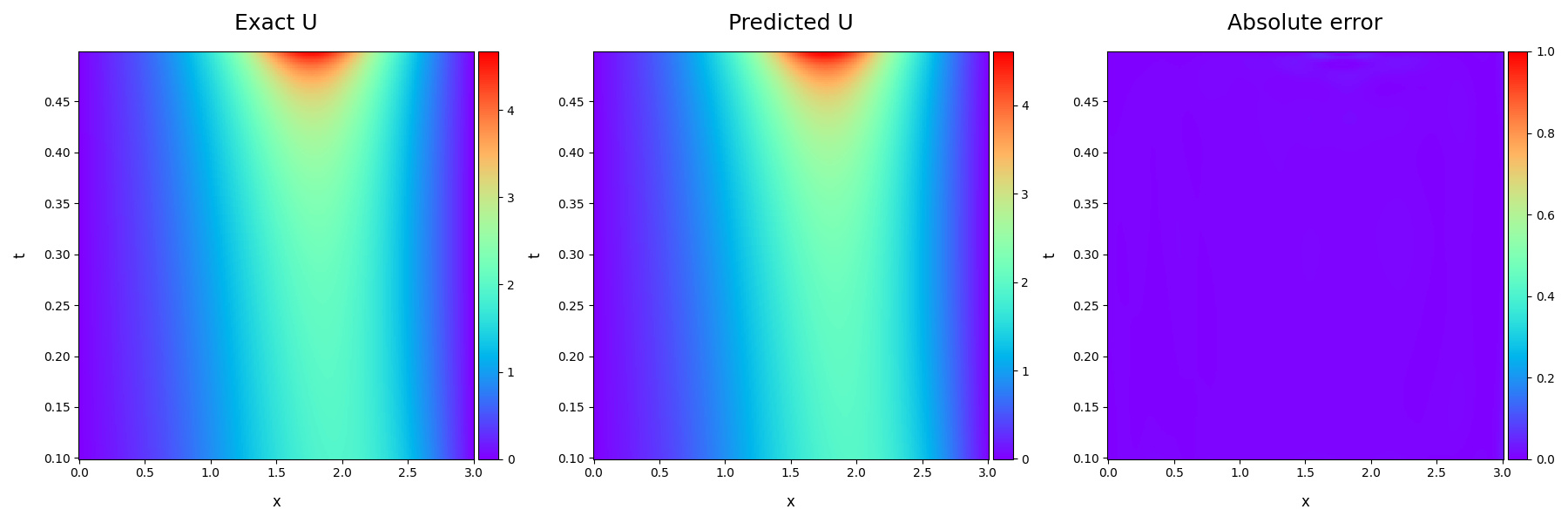} \\
\midrule
NLS & $\sigma_\text{NR}=0, r=25\%$ & 
\makecell[c]{
  \includegraphics[width=0.3\textwidth]{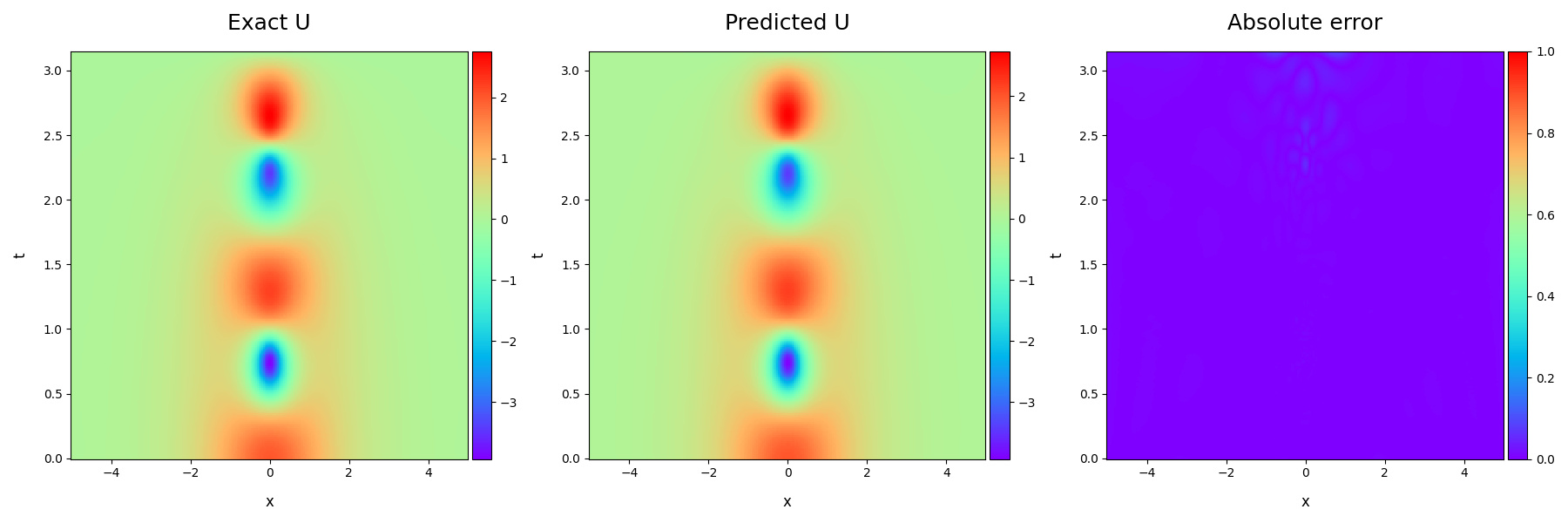} \\
  \hline
  \includegraphics[width=0.3\textwidth]{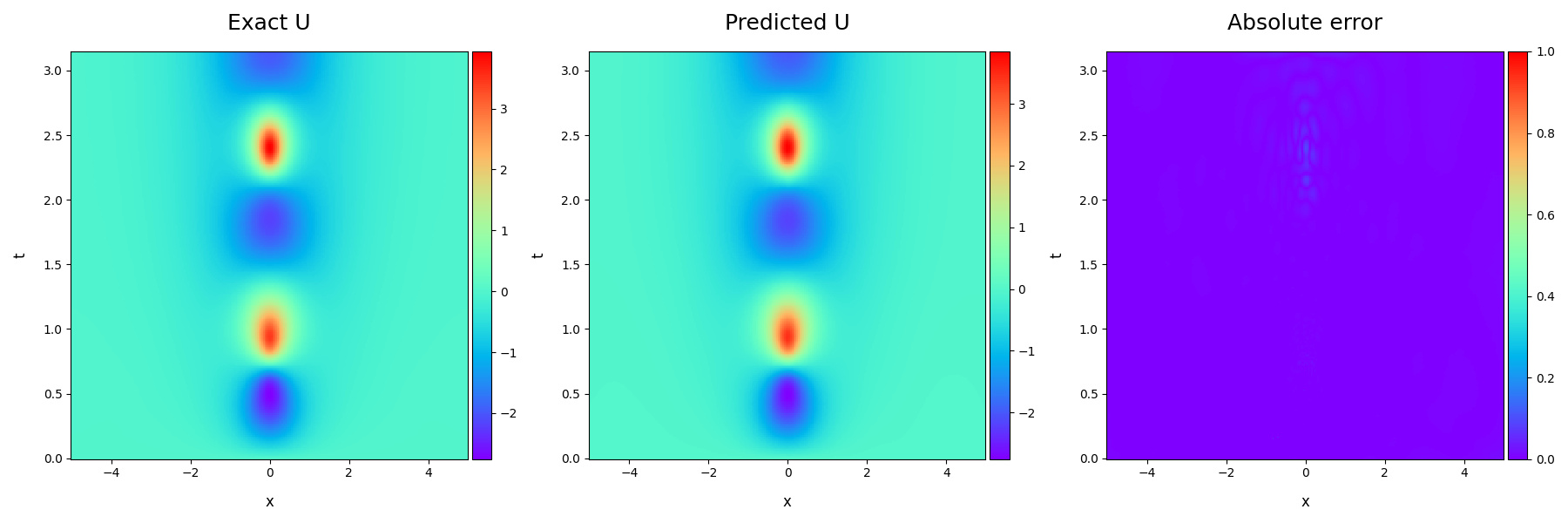} \\
} \\
\bottomrule
\end{tabular}
}
\end{table}

\begin{table}[H]
\centering
\caption{Reconstruction performance of the learned response function 
$U$ for 2D equations.}
\label{tab:results_visual_2d}
\begin{tabular}{c c c c}
\toprule
{Equation} & {Settings} & {Time} & {Visualization of the true and learned system dynamics} \\
\toprule
\multirow{18}{*}{Wave} 
& \multirow{18}{*}{$\sigma_\text{NR}=0, r=10\%$}
& $t = 20\ \text{time step}$ 
& \includegraphics[width=0.4\textwidth, valign=m]{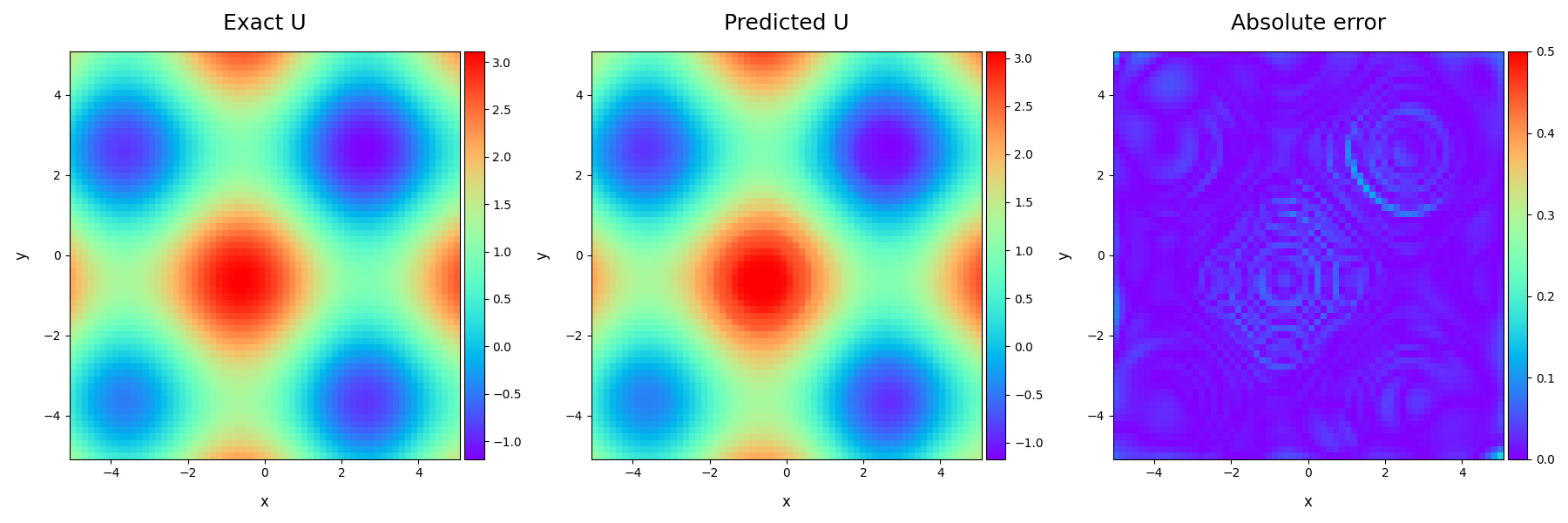} \\
\hhline{~~--}
& & $t = 40\ \text{time step}$ 
& \includegraphics[width=0.4\textwidth, valign=m]{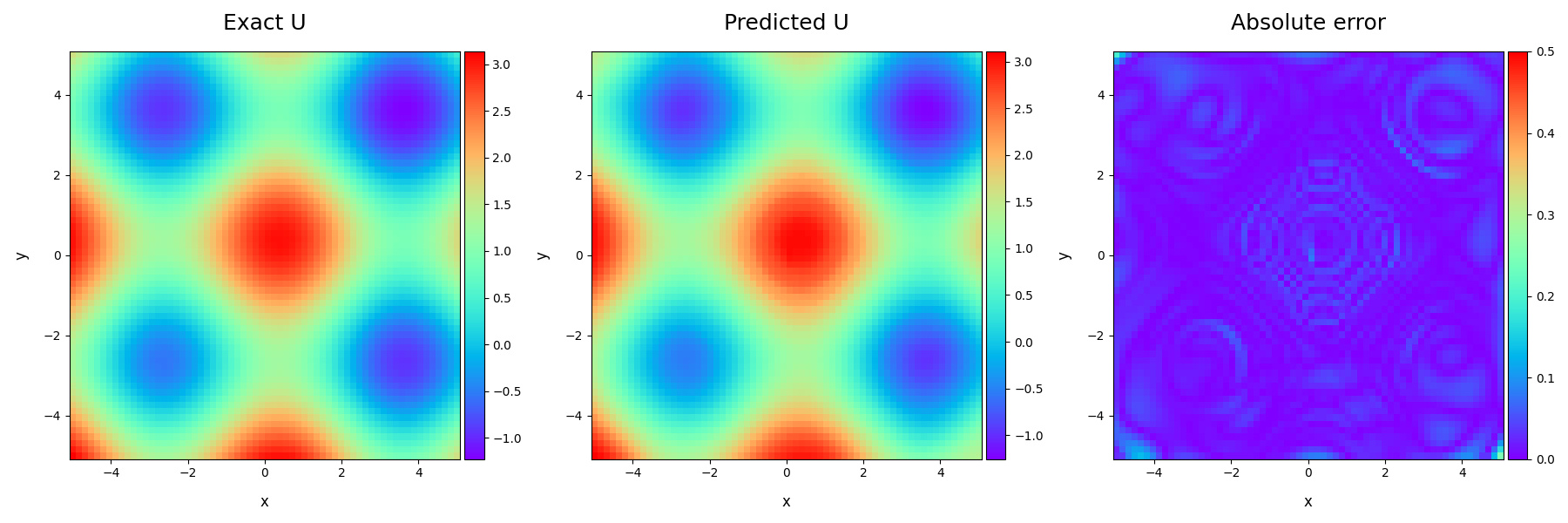} \\
\hhline{~~--}
& & $t = 60\ \text{time step}$ 
& \includegraphics[width=0.4\textwidth, valign=m]{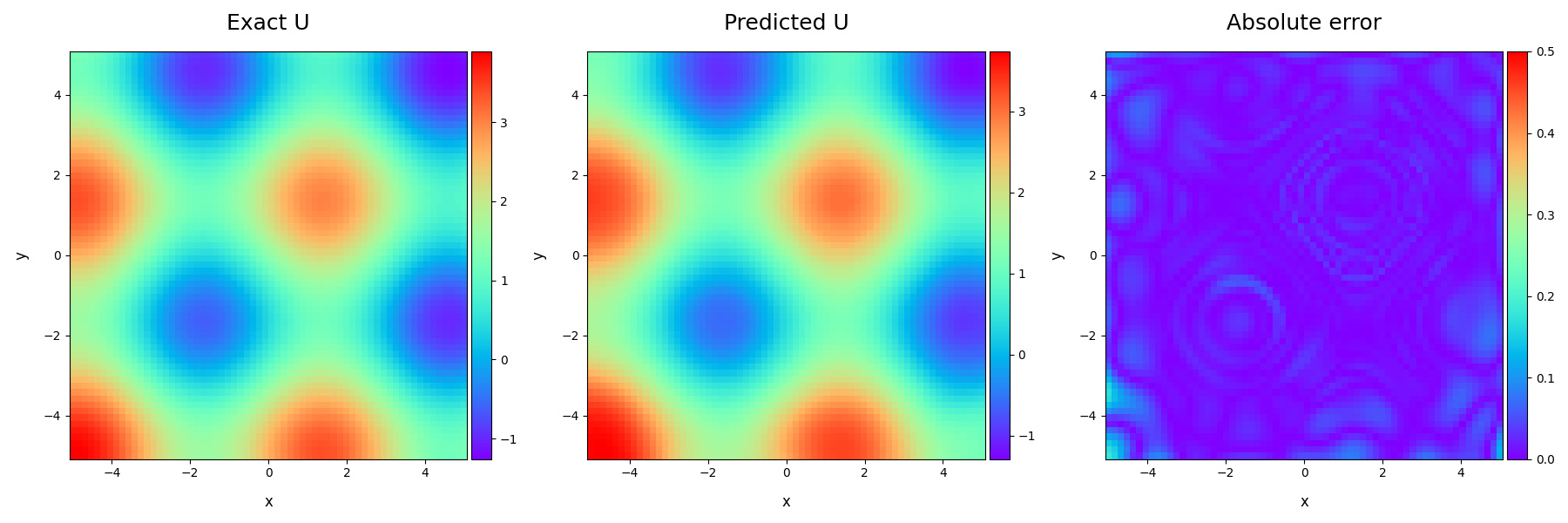} \\
\hhline{~~--}
& & $t = 80\ \text{time step}$ 
& \includegraphics[width=0.4\textwidth, valign=m]{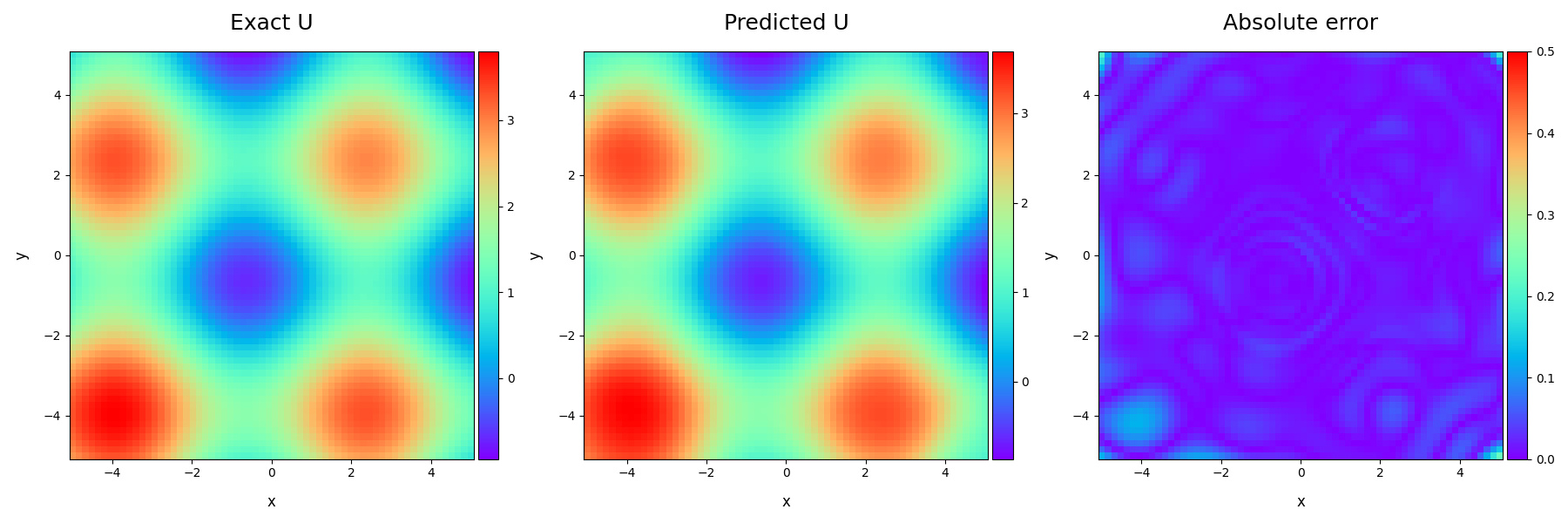} \\
\midrule
\multirow{18}{*}{SG} 
& \multirow{18}{*}{$\sigma_\text{NR}=0, r=10\%$}
& $t = 15\ \text{time step}$ 
& \includegraphics[width=0.4\textwidth, valign=m]{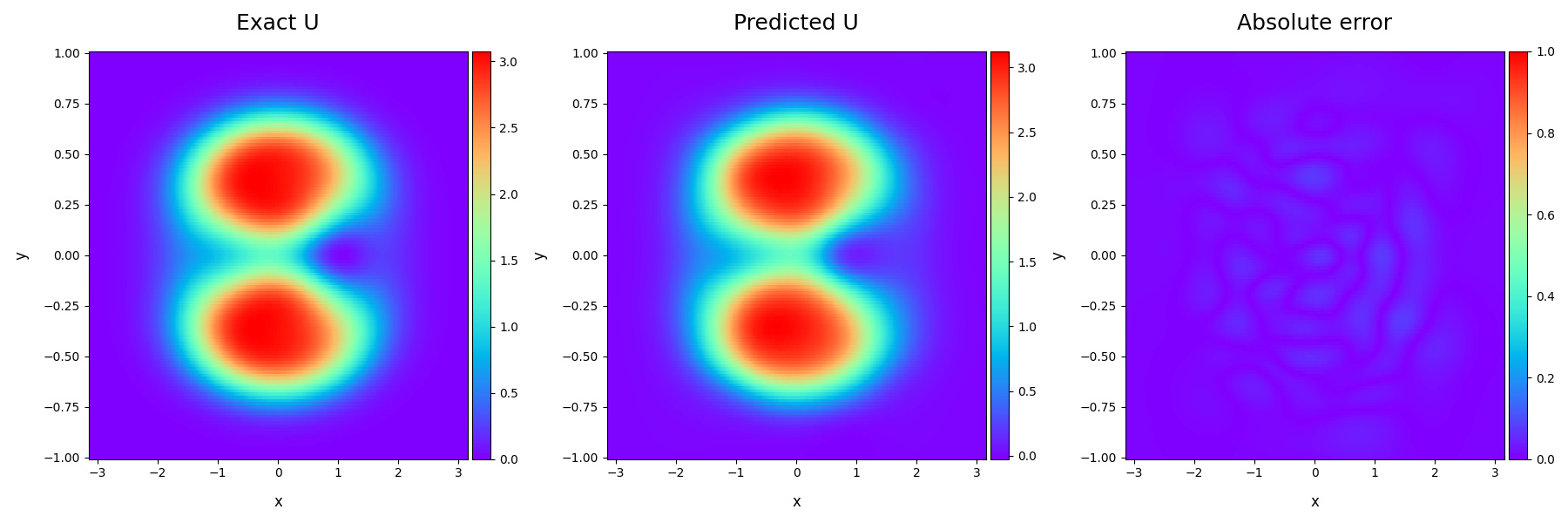} \\
\hhline{~~--}
& & $t =30\ \text{time step}$
& \includegraphics[width=0.4\textwidth, valign=m]{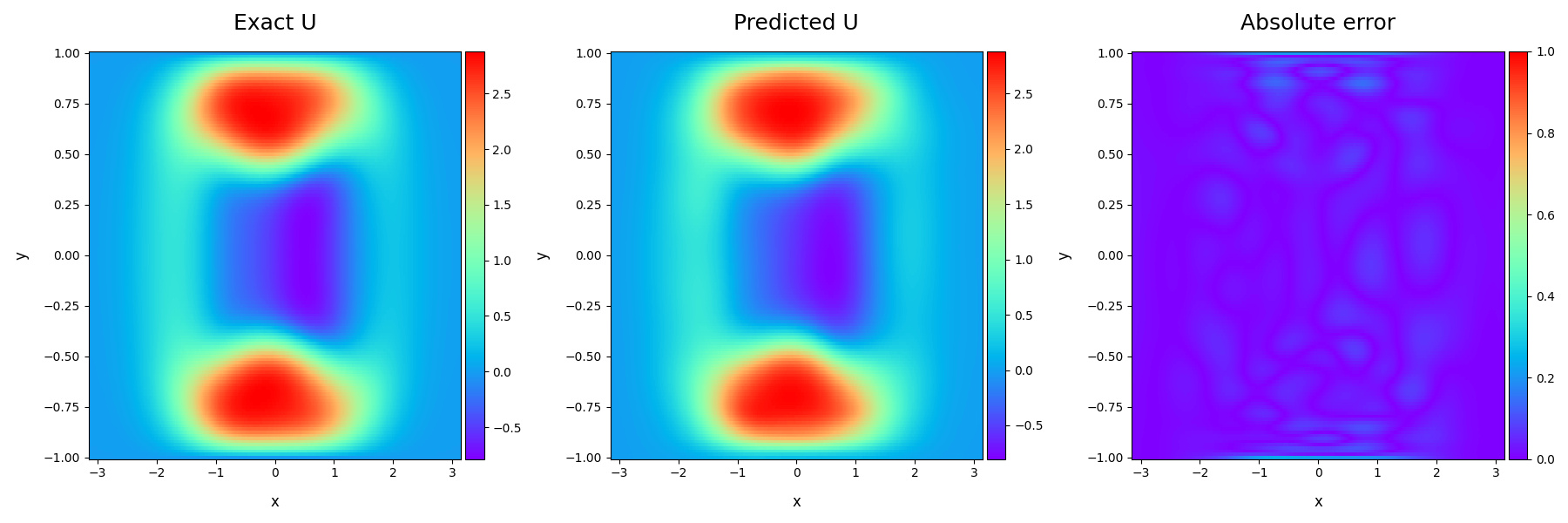} \\
\hhline{~~--}
& & $t =45\ \text{time step}$
& \includegraphics[width=0.4\textwidth, valign=m]{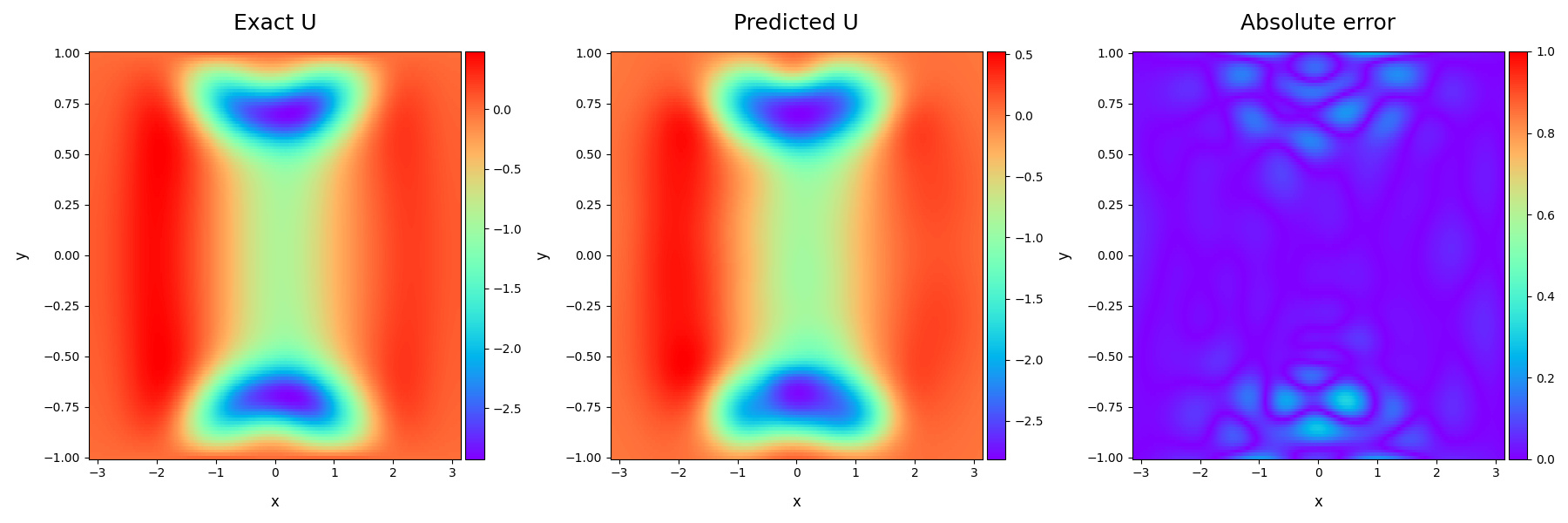} \\
\hhline{~~--}
& & $t =60\ \text{time step}$
& \includegraphics[width=0.4\textwidth, valign=m]{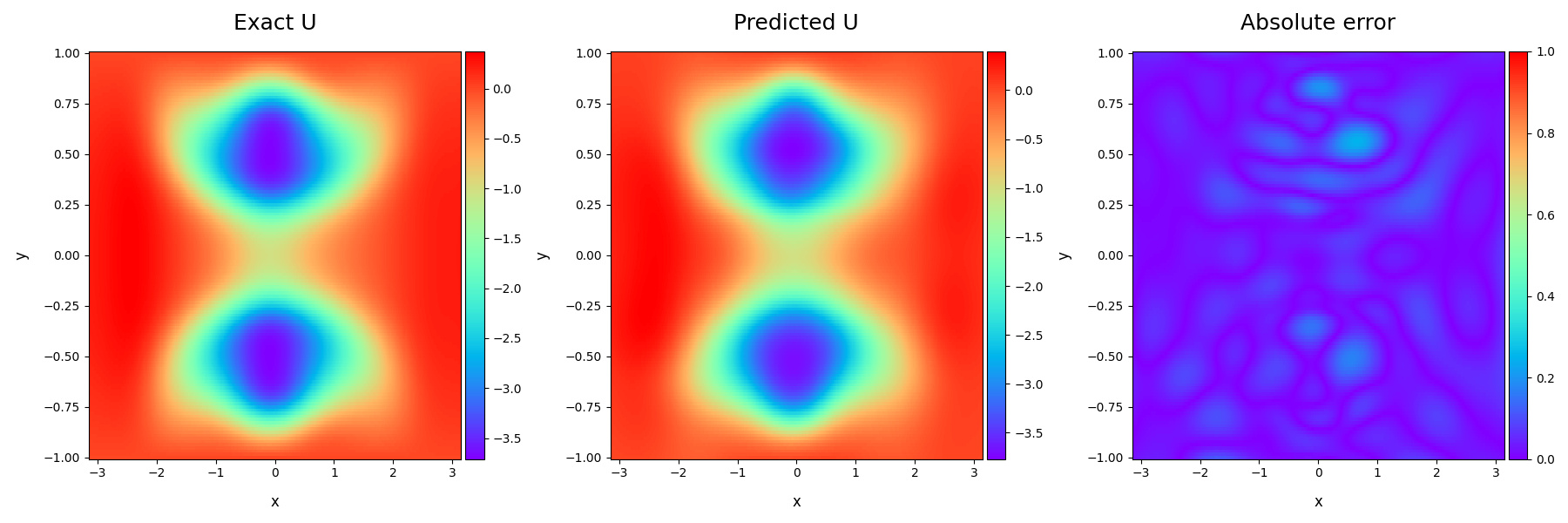} \\

\bottomrule
\end{tabular}
\end{table}

\begin{table}[H]
\centering
\caption{Reconstruction performance of the learned response function 
$U$ for 2D NS equation.}
\label{tab:results_visual_ns}
\begin{tabular}{c c c c}
\toprule
{Settings} & {Time} & {Visualization of the true and learned system dynamics} \\
\toprule
\multirow{39}{*}{$\sigma_\text{NR}=0, r=10\%$}
& $t = 0\ \text{time step}$ 
& \includegraphics[width=0.45\textwidth, valign=m]{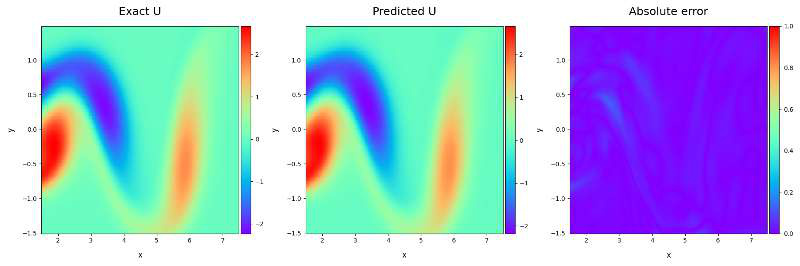} \\
\hhline{~~--}
 & $t = 15\ \text{time step}$ 
& \includegraphics[width=0.45\textwidth, valign=m]{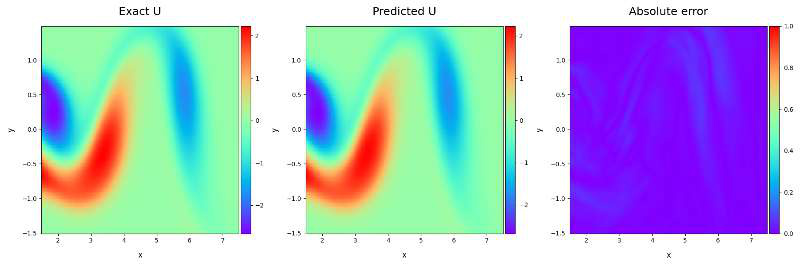} \\
\hhline{~~--}
& $t = 30\ \text{time step}$ 
& \includegraphics[width=0.45\textwidth, valign=m]{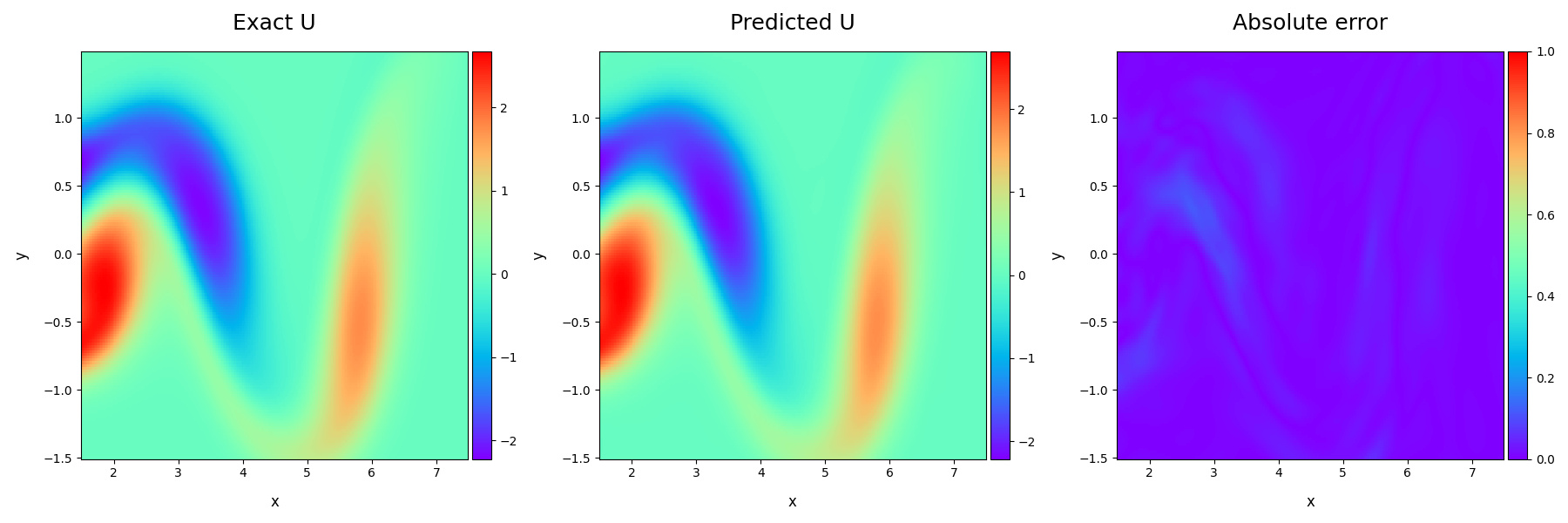} \\
\hhline{~~--}
& $t = 45\ \text{time step}$ 
& \includegraphics[width=0.45\textwidth, valign=m]{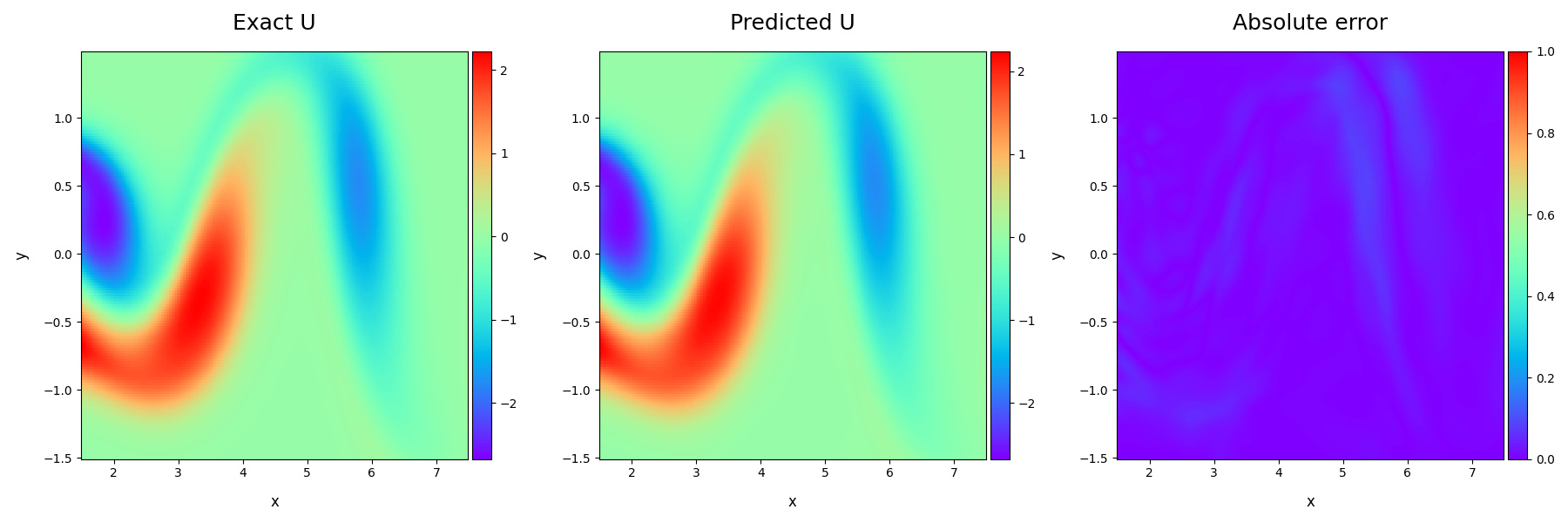} \\
& $t = 60\ \text{time step}$ 
& \includegraphics[width=0.45\textwidth, valign=m]{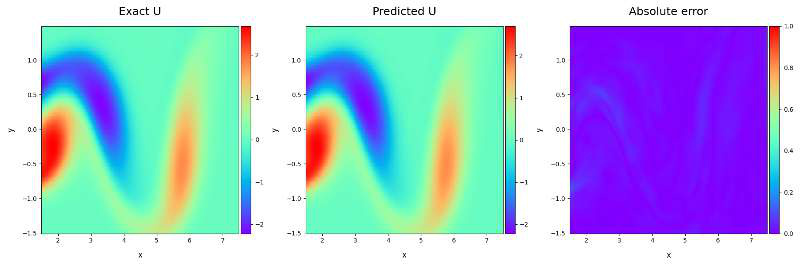} \\
\hhline{~~--}
& $t = 75\ \text{time step}$ 
& \includegraphics[width=0.45\textwidth, valign=m]{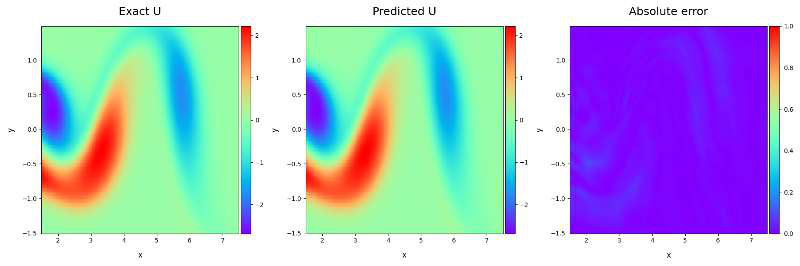} \\
\hhline{~~--}
& $t = 90\ \text{time step}$ 
& \includegraphics[width=0.45\textwidth, valign=m]{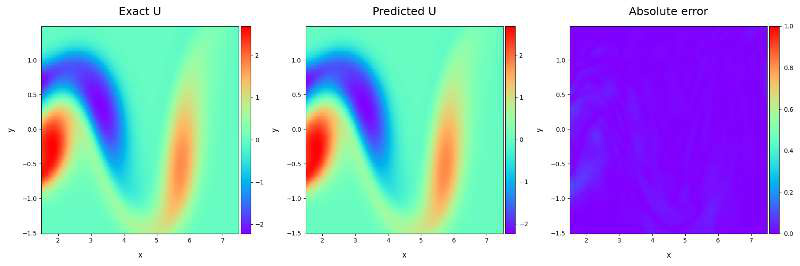} \\
\bottomrule
\end{tabular}
\end{table}

\begin{table}[H]
\centering
\caption{Detailed results of ablation studies.}
\label{tab: ablation_results}
\resizebox{\textwidth}{!}{
\begin{tabular}{lcccccl}
\toprule
{Equation} & {Settings} & ${\mathrm{TPR}}$ & ${E_{\infty}}$ & ${E_2}$ & ${\mathcal{L}(\hat{U},U)}$ & {Discovered Equation} \\
\midrule
\multirow{3}{*}{Burgers} 
& $r=10\%$, $\mathrm{FE}_{\phi}$+NAS & 1.00 & 0.0096 & 0.0095 & $7.14\times10^{-5}$ & $\partial_t u = -0.5048\partial_x u^2 + 0.1007\partial_{xx}u$ \\
& $r=10\%$, without $\mathrm{FE}_{\phi}$  & 0.00 & 1.0000 &3.1928 & $2.30\times10^{-4}$ & $\partial_t u = -1.5006\partial_x u^4 + 0.3724\partial_{xx}u^4$\\
& $r=10\%$, without NAS  & 0.17 & 2.1914 &8.9731 & $8.63\times10^{-5}$ & $\partial_t u = 3.0014\partial_x u^4 -3.2493\partial_x u^3+ 0.5957\partial_{x}u^2+\cdots$ \\
\midrule

\multirow{3}{*}{KdV} 
& $r=25\%$, $\mathrm{FE}_{\phi}$+NAS & 1.00 & 0.0376 & 0.0119 & $6.49\times10^{-5}$ & $\partial_t u = -3.0013\partial_x u^2 - 1.0376\partial_{xxx}u$ \\
& $r=25\%$, without $\mathrm{FE}_{\phi}$ & 0.00 & 1.0000 & 1.1379 & $7.55\times10^{-3}$ & $\partial_t=0.3366u^2-0.4742\partial_{x}u+1.5952\partial_{xx}u^2-0.2539\partial_{xx}u$ \\
& $r=25\%$, without NAS  & 0.33 & 0.7369 & 0.7598 & $2.90\times10^{-3}$ & $\partial_t=-0.7894\partial_{x}u^2-0.6544\partial_{xxx}u+\cdots$ \\
\midrule
\multirow{3}{*}{KS} 
& $r=25\%$, $\mathrm{FE}_{\phi}$+NAS  & 1.00 & 0.0123 & 0.0116 & $1.12\times10^{-3}$ & $\partial_t u = -0.5022 \partial_x(u^2) - 1.0121\partial_{xx} u - 1.0123\partial_{xxxx} u$\\
& $r=25\%$, without $\mathrm{FE}_{\phi}$ & 0.25 & 1.0000 & 0.9600 & $3.71\times10^{-3}$ & $\partial_t=-0.2390\partial_{x}u^2-0.0735\partial_{xx}u^3$ \\
& $r=25\%$, without NAS  & 0.75 & 0.1826 & 0.1693 & $3.29\times10^{-3}$ & $\partial_t=-0.4659\partial_x(u^2)-0.8283\partial_{xx} u-0.8174\partial_{xxxx} u+\cdots$ \\
\midrule
\multirow{3}{*}{CI} 
&  $r=25\%$, $\mathrm{FE}_{\phi}$+NAS & 1.00 & 0.1224 & 0.0955 & $6.06\times10^{-6}$ & $\partial_t=0.9317u^3-0.8776u+0.9123\partial_{xx}u$\\
& $r=25\%$, without $\mathrm{FE}_{\phi}$ &0.50 & 2.0156 & 1.4652 & $4.90\times10^{-5}$ & $\partial_t=1.0156u+0.3463\partial_{x}u+0.1218\partial_{xx}u$\\
& $r=25\%$, without NAS  & 0.25 &1.4324& 1.1841 & $7.71\times10^{-4}$ & $\partial_t=0.3931\partial_{x}u-0.4324\partial_{xx}u$ \\
\midrule
\multirow{3}{*}{NLS} 
&  $r=25\%$, $\mathrm{FE}_{\phi}$+NAS &\begin{tabular}{c}
$1.00$ \\
$1.00$
\end{tabular}&\begin{tabular}{c}
$0.0049$ \\
$0.0034$
\end{tabular}& \begin{tabular}{c}
$0.0034$ \\
$0.0025$
\end{tabular} & \begin{tabular}{c}
$3.17\times10^{-5}$ \\
$2.69\times10^{-5}$
\end{tabular} & $\begin{cases} \partial_t u =  0.4990 \partial_{xx} v + 0.9951u^2 v + 1.0008v^3  \\ \partial_t v = -0.4991 \partial_{xx} u - 1.0034u v^2 - 0.9987u^3 \end{cases}$\\
\addlinespace
&  $r=25\%$, without $\mathrm{FE}_{\phi}$ &\begin{tabular}{c}
$0.29$ \\
$0.29$
\end{tabular}&\begin{tabular}{c}
$3.2458$ \\
$3.3048$
\end{tabular}& \begin{tabular}{c}
$2.4957$ \\
$2.5672$
\end{tabular} & \begin{tabular}{c}
$3.53\times10^{-5}$ \\
$3.54\times10^{-5}$
\end{tabular} & $\begin{cases} \partial_t u =  0.4505 \partial_{xx} v + 4.2458u^2 v +0.7964u^2 +\cdots \\ \partial_t v = -0.4086 \partial_{xx} u - 4.3048u v^2 +1.1144 v^2 +\cdots\end{cases}$\\ 
\addlinespace
& $r=25\%$, without NAS &\begin{tabular}{c}
$0.18$ \\
$0.18$
\end{tabular}&\begin{tabular}{c}
$2.6342$ \\
$2.4819$
\end{tabular}& \begin{tabular}{c}
$4.0021$ \\
$3.2305$
\end{tabular} & \begin{tabular}{c}
$7.69\times10^{-5}$ \\
$7.69\times10^{-5}$
\end{tabular} & $\begin{cases} \partial_t u = 3.6341u^2 v +0.4545u^3+0.1544u^3v +\cdots \\ \partial_t v = -0.4086 \partial_{xx} u - 4.3048u v^2 +1.1144 v^2 +\cdots\end{cases}$\\
\midrule
\multirow{3}{*}{Wave} 
& $r=5.0\%$, $\mathrm{FE}_{\phi}$+NAS & 1.00& 0.0080 & 0.0056 & $9.19\times10^{-4}$ & $\partial_{tt}u=0.9999\partial_{xx}u+0.9920\partial_{yy}u$\\
& $r=5.0\%$, without $\mathrm{FE}_{\phi}$& 1.00& 0.7775 & 0.7291 & $6.30\times10^{-3}$ & $\partial_{tt}u=0.2225\partial_{xx}u+0.3227\partial_{yy}u$\\
& $r=5.0\%$, without NAS  & 1.00 & 0.0073 & 0.0063 & $6.90\times10^{-4}$ & $\partial_{tt} u=0.9927\partial_{xx}u+0.9948\partial_{yy}u$\\
\midrule
\multirow{3}{*}{SG} 
& $r=10\%$, $\mathrm{FE}_{\phi}$+NAS & 1.00& 0.1341 & 0.1052 & $3.08\times10^{-3}$ & $\partial_{tt}u =  0.8775\partial_{xx}u+0.9855\partial_{yy}u-1.1341sin(u)$\\
& $r=10\%$, without $\mathrm{FE}_{\phi}$  & 0.33 & 4.0208 & 2.7365 & $1.62\times10^{-1}$ & $\partial_{tt}=1.7245\partial_{xx}u+1.0665\partial_{yy}u-5.0208\sin(u)+\cdots$ \\
& $r=10\%$, without NAS  & 0.50 & 0.5838 &0.4628 & $5.17\times10^{-1}$ & $\partial_{tt} u=1.5838\partial_{xx}u+0.5167\partial_{yy}u-1.2611u$ \\
\midrule
\multirow{3}{*}{NS($Re=100$)} &$r=10\%,\ \mathrm{FE}_{\phi}$+NAS & 1.00 & 0.0170 & 0.0052 & $1.23\times10^{-3}$ & $\partial_t \omega = -0.9927\partial_x(u\omega) -1.0004\partial_y(v\omega) + 0.0099\partial_{xx}\omega + 0.0102\partial_{yy}\omega$ \\
&$r=10\%$, without $\mathrm{FE}_{\phi}$& 0.75& 1.0000 & 0.0241 & $1.02\times10^{-2}$ & $\partial_t \omega = -1.001\partial_x(u\omega) -0.9676\partial_y(v\omega) + 0.0131\partial_{yy}\omega$ \\
&$r=10\%$, without NAS& 1.00 & 0.0433 & 0.0080 & $3.25\times10^{-3}$ & $\partial_t \omega = -0.9915\partial_x(u\omega) -0.9926\partial_y(v\omega) + 0.0104\partial_{xx}\omega + 0.0103\partial_{yy}\omega$ \\

\bottomrule
\end{tabular}}
\end{table}

\begin{table}[H]
\centering
\caption{Detailed results of our method in comparison experiments.}
\label{tab:results_comparison_ours}
\begin{tabular}{c c c ccc l}
\toprule
$\sigma_{\text{NR}}$ & $N_{\text{data}}$ & TPR & ${E_{\infty}}$ & ${E_2}$ & ${\mathcal{L}(\hat{U},U)}$ & Discovered Equation \\
\midrule
\multicolumn{7}{c}{\textbf{Burgers Equation}} \\ 
\midrule
25\% & 4000 & 1.00 & 0.0098 & 0.0097 & $2.67\times10^{-4}$ & $\partial_t u = -0.4950\partial_x u^2 + 0.0999\partial_{xx}u$ \\ 
50\% & 4000 & 1.00 & 0.0593 & 0.0117 & $7.58\times10^{-4}$ & $\partial_t u = -0.4992\partial_x u^2 + 0.0941\partial_{xx}u$ \\
75\% & 4000 & 1.00 & 0.1746 & 0.0343 & $2.35\times10^{-3}$ & $\partial_t u = -0.5010\partial_x u^2 + 0.1175\partial_{xx}u$ \\
100\% & 4000 & 1.00 & 0.1931 & 0.0552 & $4.24\times10^{-3}$ & $\partial_t u = -0.5011\partial_x u^2 + 0.1021\partial_{xx}u$ \\
\midrule
\multicolumn{7}{c}{\textbf{KdV Equation}} \\
\midrule
25\% & 4000 & 1.00 & 0.0040 & 0.0036 & $1.12\times10^{-2}$ & $\partial_t u = -0.5006\partial_x u^2 - 1.0040\partial_{xx}u$ \\ 
50\% & 4000 & 1.00 & 0.0136 & 0.0062 & $2.90\times10^{-2}$ & $\partial_t u = -0.5068\partial_x u^2 - 0.9988\partial_{xxx}u$ \\
75\% & 4000 & 1.00 & 0.0393 & 0.0352 & $6.94\times10^{-2}$ &
$\partial_t u =  -0.5016\partial_x u^2 - 0.9607\partial_{xxx}u$ \\
100\% & 4000 & 1.00 & 0.0732 & 0.0719 & $1.42\times10^{-1}$ & $\partial_t u = -0.4634\partial_x u^2 - 0.9285\partial_{xxx}u$ \\
\midrule
\multicolumn{7}{c}{\textbf{KS Equation}} \\
\midrule
25\% & 10000 & 1.00 & 0.0066 & 0.0049 & $1.01\times10^{-2}$ & $\partial_t u = -0.5024\partial_x(u^2) - 1.0066\partial_{xx} u - 0.9980\partial_{xxxx} u$ \\
50\% & 10000 & 1.00 & 0.0487 & 0.0342 & $1.09\times10^{-2}$ & $\partial_t u = -0.4862\partial_x(u^2) -1.0487\partial_{xx} u -0.9917\partial_{xxxx} u$ \\ 
75\% & 10000 & 1.00 & 0.0895 & 0.0331 & $8.61\times10^{-2}$ & $\partial_t u = -0.4552\partial_x(u^2) -0.9855\partial_{xx} u -0.9841\partial_{xxxx} u$ \\
100\% & 10000 & 1.00 & 0.1340 & 0.0973 & $9.85\times10^{-2}$ & $\partial_t u = -0.4330\partial_x(u^2) -0.9387\partial_{xx} u - 0.8856\partial_{xxxx} u$ \\ 
\bottomrule
\end{tabular}
\end{table}
\printcredits

\FloatBarrier
\clearpage
\begin{spacing}{0.91}   
\bibliographystyle{cas-model2-names}
\bibliography{cas-refs}
\end{spacing}

\end{document}